\documentclass{article}

\PassOptionsToPackage{sort&compress, numbers}{natbib}

\usepackage[final]{nips}
\usepackage[utf8]{inputenc} %
\usepackage[T1]{fontenc}    %
\usepackage{hyperref}       %
\usepackage{url}            %
\usepackage{algorithm}
\usepackage{graphicx}
\usepackage{color}
\usepackage{algorithmic}
\usepackage{eqparbox}
\usepackage{subcaption}
\usepackage{makecell}
\usepackage{bibentry}
\usepackage{float}
\usepackage{placeins}

\usepackage{amsmath,amssymb}

\usepackage{booktabs}       %
\usepackage{amsfonts}       %
\usepackage{nicefrac}       %
\usepackage{microtype}      %
\usepackage{xcolor}
\usepackage{xspace}
\usepackage{verbatim}
\usepackage{cleveref}
\usepackage{pifont}%
\usepackage[normalem]{ulem}
\usepackage{siunitx} %
\newcommand{\pivotci}[2]{(CI: #1 -- #2)}  %
\newcommand{\percci}[2]{}  %
\newcommand{\tpivotci}[2]{(#1 -- #2)}  %
\newcommand{\tpercci}[2]{}  %
\newcommand{\na}[0]{-}  %

\nobibliography*

\title{Go-Explore: a New Approach for Hard-Exploration Problems }

\author{
\textbf{Adrien Ecoffet} \hspace{5mm} \textbf{Joost Huizinga} \hspace{5mm} \textbf{Joel Lehman} \hspace{5mm}  \textbf{Kenneth O. Stanley*}  \hspace{5mm}  \textbf{Jeff Clune*} \\
  Uber AI Labs\\
  San Francisco, CA 94103 \\
  \texttt{adrienecoffet,joost.hui,jclune@gmail.com} \\
  *Co-senior authors
}
\begin{document}

\maketitle

\vspace{-6mm}
Authors' note: We recommend reading (and citing) our updated paper, ``First return, then explore'': 

Ecoffet, A., Huizinga, J., Lehman, J., Stanley, K.O. and Clune, J. First return, then explore. \emph{Nature} \textbf{590,} 580–586 (2021). \href{https://doi.org/10.1038/s41586-020-03157-9}{https://doi.org/10.1038/s41586-020-03157-9}

It can be found at \href{https://tinyurl.com/Go-Explore-Nature}{https://tinyurl.com/Go-Explore-Nature}.
\vspace{4mm}

\begin{abstract}
A grand challenge in reinforcement learning is intelligent exploration, especially when rewards are sparse or deceptive. Two Atari games serve as benchmarks for such hard-exploration domains: Montezuma's Revenge and Pitfall. On both games, current RL algorithms perform poorly, even those with intrinsic motivation, which is the dominant method to encourage exploration and improve performance on hard-exploration domains.  To address this shortfall, we introduce a new algorithm called Go-Explore. 
It exploits the following principles: (1)~remember states that have previously been visited, (2)~first return to a promising state (without exploration), then explore from it, and (3)~solve simulated environments through exploiting any available means (including by introducing determinism), then robustify (create a policy that can reliably perform the solution) via imitation learning. The combined effect of these principles generates dramatic performance improvements on hard-exploration problems.
On Montezuma's Revenge, without being provided any domain knowledge, Go-Explore scores over 43,000 points, almost 4 times the previous state of the art. Go-Explore can also easily harness human-provided domain knowledge, and when augmented with it Go-Explore scores a mean of over 650,000 points on Montezuma's Revenge. Its max performance of 18 million surpasses the human world record by an order of magnitude, thus meeting even the strictest definition of ``superhuman'' performance. On Pitfall, Go-Explore with domain knowledge is the first algorithm to score above zero. Its mean performance of almost 60,000 points also exceeds expert human performance.
Because Go-Explore can produce many high-performing demonstrations automatically and cheaply, it also outperforms previous imitation learning work in which the solution was provided in the form of a human demonstration. Go-Explore opens up many new research directions into improving it and weaving its insights into current RL algorithms. It may also enable progress on previously unsolvable hard-exploration problems in a variety of domains, especially the many that often harness a simulator during training (e.g. robotics).

\end{abstract}
\section{Introduction} \label{intro} %

Reinforcement learning (RL) has experienced significant progress in recent years, achieving
superhuman performance in board games such as Go~\cite{silver2016mastering,Silver2017MasteringTG} and in classic video games such as Atari~\cite{mnih:nature15}. However, this progress obscures some of the deep unmet challenges in scaling RL to complex real-world domains. In particular, many important tasks require \emph{effective exploration} to be solved, i.e.\ to explore and learn about the world even when rewards are \emph{sparse} or \emph{deceptive}. In sparse-reward problems, precise sequences of many (e.g.\ hundreds or more) actions must be taken between obtaining rewards. Deceptive-reward problems are even harder, because instead of feedback rarely being provided, the reward function actually provides misleading feedback for reaching the overall global objective, which can lead to getting stuck on local optima. Both sparse and deceptive reward problems constitute ``hard-exploration'' problems, and classic RL algorithms perform poorly on them~\cite{bellemare2016unifying}. Unfortunately, most challenging real-world problems are also hard-exploration problems. That is because we often desire to provide abstract goals (e.g.\ ``find survivors and tell us their location,'' or ``turn off the valve to the leaking pipe in the reactor''), and such reward functions do not provide detailed guidance on how to solve the problem (sparsity) while also often creating unintended local optima (deception)~\cite{Amodei2016ConcretePI,Lehman2018TheSC,lehman,Chrabaszcz2018BackTB}.

For example, in the case of finding survivors in a disaster area, survivors will be few and far between, thus introducing sparsity. Even worse, if we also instruct the robot to minimize damage to itself, this additional reward signal may actively teach the robot not to explore the environment, because exploration is initially much more likely to result in damage than it is to result in finding a survivor. This seemingly sensible additional objective thus introduces deception on top of the already sparse reward problem.

To address these challenges, this paper introduces \emph{Go-Explore}, a new algorithm for hard-exploration problems
that dramatically improves state-of-the-art performance in two classic hard-exploration
benchmarks: the Atari games Montezuma's Revenge and Pitfall.

Prior to Go-Explore, the typical approach to sparse reward problems has been \textit{intrinsic motivation} (IM)~\cite{bellemare2016unifying,schmidhuber1991possibility,oudeyer2009intrinsic,barto2013intrinsic}, which supplies the RL agent with intrinsic rewards (IRs) that encourage exploration (augmenting or replacing extrinsic reward that comes from the environment). 
IM is often motivated by psychological concepts such as curiosity~\cite{Schmidhuber2006DevelopmentalRO,schmidhuber1991curious} or novelty-seeking~\cite{lehman,conti:arxiv17}, which play a role in how humans explore and learn.
While IM has produced exciting progress in sparse reward problems, in many domains IM approaches are still far from fully solving the problem, including on Montezuma's Revenge and Pitfall.
We hypothesize that, amongst other issues, such failures stem from two root causes that we call \emph{detachment} and \emph{derailment}.

\begin{figure}
    \centering
    \includegraphics[width=\linewidth]{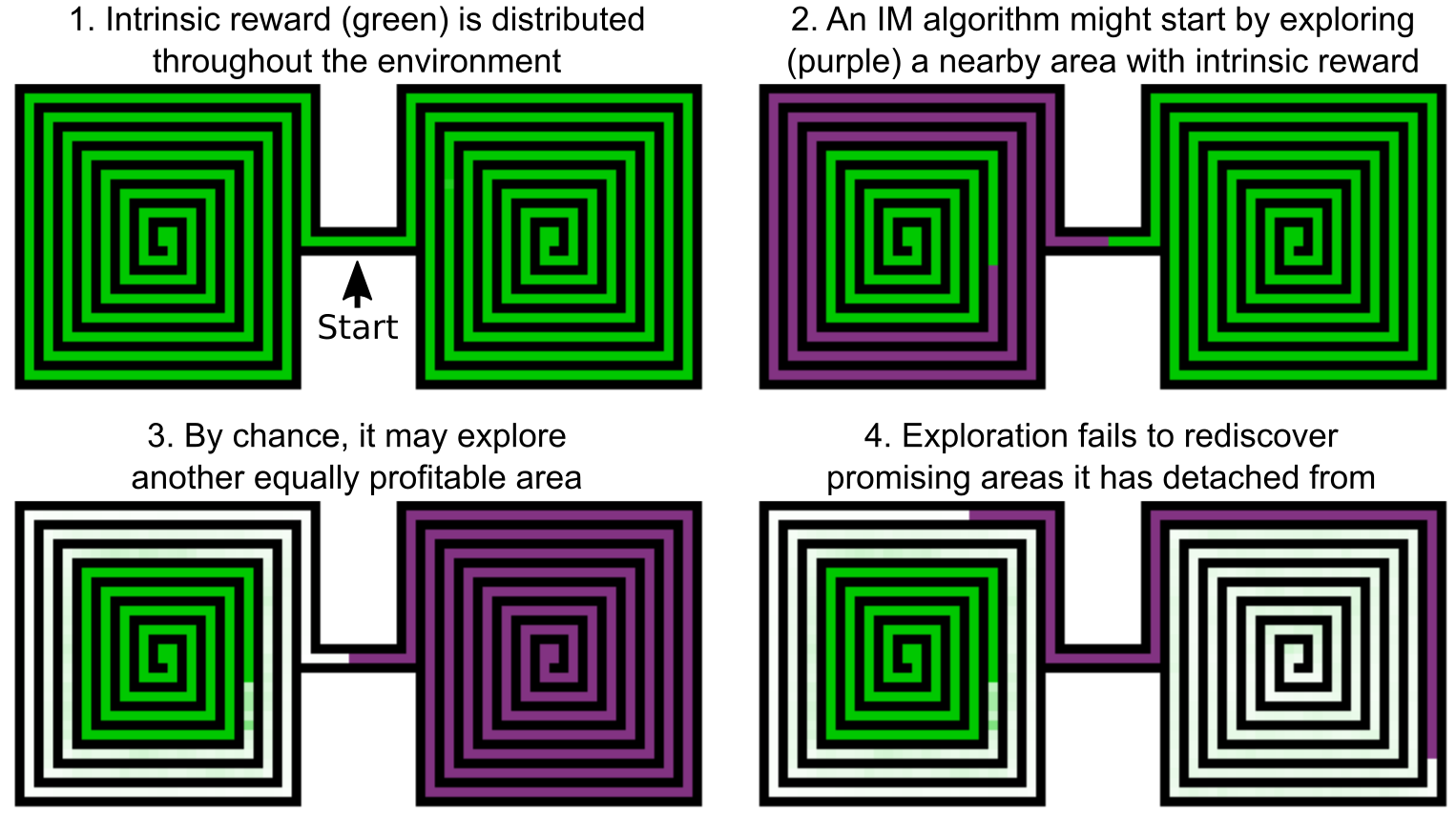}
    \caption{\textbf{A hypothetical example of detachment in intrinsic motivation (IM) algorithms.} Green areas indicate intrinsic reward, white indicates areas where no intrinsic reward remains, and purple areas indicate where the algorithm is currently exploring. (1)~The agent starts each episode between the two mazes. (2)~It may by chance start exploring the West maze and IM may drive it to learn to traverse, say, 50\% of it. (3)~Because current algorithms sprinkle in randomness (either in actions or parameters) to try to produce new behaviors to find explicit or intrinsic rewards, by chance the agent may at some point begin exploring the East maze, where it will also encounter a lot of intrinsic reward. After completely exploring the East maze, it has no explicit memory of the promising exploration frontier it abandoned in the West maze. It likely would also have no implicit memory of this frontier due to the problem of catastrophic forgetting~\cite{velez2017diffusion, ellefsen2015neural, kirkpatrick2017overcoming,french1999catastrophic}. (4)
    Worse, the path leading to the frontier in the West maze has already been explored, so no (or little) intrinsic motivation remains to rediscover it. We thus say the algorithm has \emph{detached} from a frontier of states that provide intrinsic motivation. As a result, exploration can stall when areas close to where the current agent visits have already been explored. This problem would be remedied if the agent remembered and returned to previously discovered promising areas for exploration, which Go-Explore does. }
    \label{fig:detachment}
\end{figure}

\emph{Detachment} is the idea that an agent driven by IM could become detached from the frontiers of high intrinsic reward (IR).
To understand detachment, we must first consider that intrinsic reward is nearly always a consumable resource: a curious agent is curious about states to the extent that it has not often visited them (similar arguments apply for surprise, novelty, or prediction-error seeking agents~\cite{bellemare2016unifying,Achiam2017SurpriseBasedIM,conti:arxiv17,burda:rnd2018}).
If an agent discovers multiple areas of the state space that produce high IR, its policy may in the short term focus on one such area. After exhausting some of the IR offered by that area, the policy may by chance begin consuming IR in another area. Once it has exhausted that IR, it is difficult for it to rediscover the frontier it detached from in the initial area, because it has already consumed the IR that led to that frontier (Fig.~\ref{fig:detachment}), and it likely will not remember how to return to that frontier due to catastrophic forgetting~\cite{velez2017diffusion, ellefsen2015neural, kirkpatrick2017overcoming,french1999catastrophic}. Each time this process occurs, a potential avenue of exploration can be lost, or at least be difficult to rediscover. In the worst case, there may be a dearth of remaining IR near the areas of state space visited by the current policy (even though much IR might remain elsewhere), and therefore no learning signal remains to guide the agent to further explore in an effective and informed way. One could slowly add intrinsic rewards back over time, but then the entire fruitless process could repeat indefinitely. In theory a replay buffer could prevent detachment, but in practice it would have to be large to prevent data about the abandoned frontier to not be purged before it becomes needed, and large replay buffers introduce their own optimization stability difficulties~\cite{Zhang2017ADL,Liu2017TheEO}. 
The Go-Explore algorithm addresses detachment by explicitly storing an archive of promising states visited so that they can then be revisited and explored from later. 

\emph{Derailment} can occur when an agent has discovered a promising state and it would be beneficial to return to that state and explore from it. Typical RL algorithms attempt to enact such desirable behavior by running the policy that led to the initial state again, but with some stochastic perturbations to the existing policy mixed in to encourage a slightly different behavior (e.g. exploring further). 
The stochastic perturbation is performed because IM agents have two layers of exploration mechanisms: (1)~the higher-level IR \emph{incentive} that rewards when new states are reached, and (2)~a more basic exploratory mechanism such as epsilon-greedy exploration, action-space noise, or parameter-space noise~\cite{sutton1998reinforcement,plappert2017parameter,Rckstie2008StateDependentEF}. Importantly, IM agents rely on the latter mechanism to \emph{discover} states containing high IR, and the former mechanism to \emph{return} to them. However, the longer, more complex, and more precise a sequence of actions needs to be in order to reach a previously-discovered high-IR state, the more likely it is that such stochastic perturbations will ``derail'' the agent from ever returning to that state. That is because the needed precise actions are naively perturbed by the basic exploration mechanism, causing the agent to only rarely succeed in reaching the known state to which it is drawn, and from which further exploration might be most effective. To address derailment, an insight in Go-Explore is that effective exploration can be decomposed into first \emph{returning} to a promising state (without intentionally adding any exploration) before then \emph{exploring} further.

\begin{figure}[t]
    \centering
    \includegraphics[width=\linewidth]{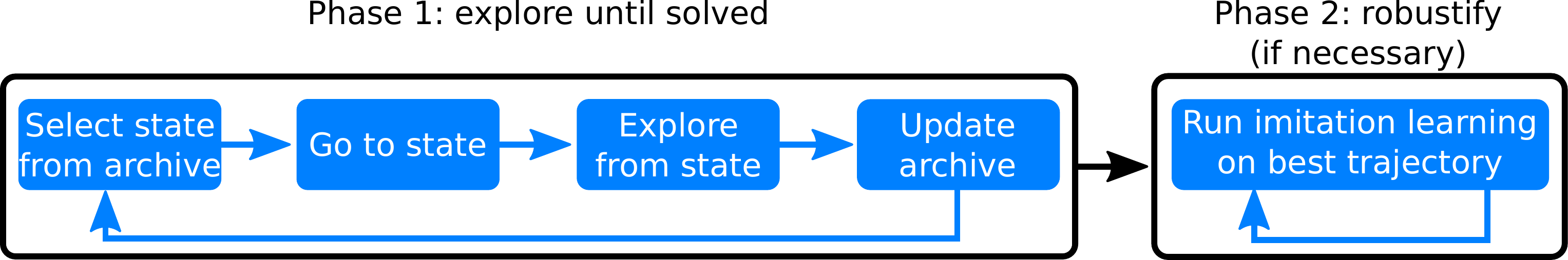}
    \caption{\textbf{A high-level overview of the Go-Explore algorithm.}}
    \label{fig:goexp_overview}
\end{figure}

Go-Explore is an explicit response to both detachment and derailment that is also designed to achieve robust solutions in stochastic environments. The version presented here works in two phases (Fig.~\ref{fig:goexp_overview}): (1)~first solve the problem in a way that may be brittle, such as solving a deterministic version of the problem (i.e. discover how to solve the problem at all), and (2)~then robustify (i.e. train to be able to reliably perform the solution in the presence of stochasticity).\footnote{Note that this second phase is in principle not necessary if Phase 1 itself produces a policy that can handle stochastic environments (Section~\ref{sec:determinism}).} Similar to IM algorithms, Phase 1 focuses on exploring infrequently visited states, which forms the basis for dealing with sparse-reward and deceptive problems. In contrast to IM algorithms, Phase 1 addresses detachment and derailment by accumulating an archive of states and ways to reach them through two strategies: (a) add all interestingly different states visited so far into the archive, and (b) each time a state from the archive is selected to explore from, first \textbf{Go} back to that state (without adding exploration), and then \textbf{Explore} further from that state in search of new states (hence the name ``Go-Explore''). 

An analogy of searching a house can help one contrast IM algorithms and Phase 1 of Go-Explore. IM algorithms are akin to searching through a house with a flashlight, which casts a narrow beam of exploration first in one area of the house, then another, and another, and so on, with the light being drawn towards areas of intrinsic motivation at the edge of its small visible region. It can get lost if at any point the beam fails to fall on any area with intrinsic motivation remaining. Go-Explore more resembles turning the lights on in one room of a house, then its adjacent rooms, then their adjacent rooms, etc., until the entire house is illuminated. Go-Explore thus gradually expands its sphere of knowledge in all directions simultaneously until a solution is discovered. 

If necessary, the second phase of Go-Explore \emph{robustifies} high-performing trajectories from the
archive such that they are robust to the stochastic dynamics of the true environment.
Go-Explore robustifies via imitation learning (aka learning from demonstrations or LfD~\cite{hester2017deep,pohlen2018observe,salimans2018learning,Ho2016GenerativeAI}), a technique that learns how to solve a task from human demonstrations. The only difference with Go-Explore is that the solution demonstrations are produced automatically by Phase 1 of Go-Explore instead of being provided by humans. 
The input to this phase is one or more high-performing trajectories, and the output is a robust policy able to consistently achieve similar performance.
The combination of both phases instantiates a powerful algorithm for hard-exploration problems, able to deeply explore sparse- and deceptive-reward environments and robustify high-performing trajectories into reliable solutions that perform well in the unmodified, stochastic test environment. 

Some of these ideas are similar to ideas proposed in related work.  Those connections are discussed in Section~\ref{sec:related_work}. That said, we believe we are the first to combine these ideas in this way and demonstrate that doing so provides substantial performance improvements on hard-exploration problems.
 
To explore its potential, we test Go-Explore on two hard-exploration benchmarks from the Arcade Learning Environment (ALE)~\cite{bellemare2013arcade,Machado2018RevisitingTA}: Montezuma's Revenge and Pitfall. Montezuma's Revenge has become an important  benchmark for exploration algorithms (including intrinsic motivation algorithms)~\cite{bellemare2016unifying,garriga2017solving,tang2017exploration,gruslys2017reactor,sasikumar2017exploration,ostrovski2017count,Stanton2018DeepCS,ODonoghue2018TheUB,burda:rnd2018,Choi2018ContingencyAwareEI} because precise sequences of hundreds of actions must be taken in between receiving rewards. 
Pitfall is even harder because its rewards are sparser (only 32 positive rewards are scattered over 255 rooms) and because many actions yield small negative rewards that dissuade RL algorithms from exploring the environment. 

Classic RL algorithms (i.e.\ those without intrinsic motivation) such as DQN~\cite{mnih:nature15}, A3C~\cite{a3c}, Ape-X~\cite{horgan:apexdqn2018} and IMPALA~\cite{espeholt:impala2018} perform poorly on these domains even with up to 22 billion game frames of experience, scoring 2,500 or lower on Montezuma's Revenge and failing to solve level one, and scoring $\leq0$ on Pitfall. Those results exclude experiments that are evaluated in a deterministic test environment~\cite{liu2019learning, Keramati2018FastEW} or were given human demonstrations~\cite{hester2017deep, pohlen2018observe, aytar2018playing}. On Pitfall, the lack of positive rewards and frequent negative rewards causes RL algorithms to learn a policy that effectively does nothing, either standing completely still or moving back and forth near the start of the game (\url{https://youtu.be/Z0lYamtgdqQ}~\cite{such2018atari}). 
 
These two games are also tremendously difficult for planning algorithms, even when allowed to plan directly within the game emulator. Classical planning algorithms such as UCT~\cite{Kocsis2006BanditBM,Kocsis2006ImprovedMS,Browne2012ASO} (a powerful form of Monte Carlo tree search~\cite{Browne2012ASO,Chaslot2008MonteCarloTS}) obtain 0 points on Montezuma's Revenge because the state space is too large to explore effectively, even with probabilistic methods~\cite{bellemare2013arcade,Lipovetzky2015ClassicalPW}. 

Despite being specifically designed to tackle sparse reward problems and being the dominant method for them, IM algorithms also struggle with Montezuma's Revenge and Pitfall, although they perform better than algorithms without IM. On Montezuma's Revenge, the best such algorithms thus far average around 11,500 with a maximum of 17,500~\cite{burda:rnd2018,Choi2018ContingencyAwareEI}. One solved level 1 of the game in 10\% of its runs~\cite{burda:rnd2018}. Even with IM, no algorithm scores greater than 0 on Pitfall (in a stochastic test environment, without a human demonstration). 
We hypothesize that detachment and derailment are major reasons why IM algorithms do not perform better.

When exploiting easy-to-provide domain knowledge, Go-Explore on Montezuma's Revenge scores a mean of 666,474, and its best run scores over 18 million and solves 1,441 levels. On Pitfall, Go-Explore scores a mean of 59,494 and a maximum of 107,363, which is close to the maximum of the game of 112,000 points. Without exploiting domain knowledge, Go-Explore still scores a mean of 43,763 on Montezuma's Revenge. All scores are dramatic improvements over the previous state of the art.
This and all other claims about solving the game and producing state-of-the-art scores assume that, while stochasticity is required during \emph{testing}, deterministic \emph{training} is allowable (discussed in Section~\ref{sec:determinism}).
We conclude that Go-Explore is a promising new algorithm for solving hard-exploration RL tasks with sparse and/or deceptive rewards.

\section{The Go-Explore Algorithm}

The insight that remembering and returning reliably to promising states is fundamental to effective exploration in sparse-reward problems is at the core of Go-Explore. Because this insight is so flexible and can be exploited in different ways, Go-Explore effectively encompasses a family of algorithms built around this key idea.  The variant implemented for the experiments in this paper and described in detail in this section relies on two distinct phases. While it provides a canonical demonstration of the possibilities opened up by Go-Explore, other variants are also discussed (e.g.\ in Section~\ref{sec:discussion_and_future_work}) to provide a broader compass for future applications. 

\subsection{Phase 1: Explore until solved}

In the two-phase variant of Go-Explore presented in this paper, the purpose of Phase 1 is to explore the state space and find one or more high-performing trajectories that can later be turned into a robust policy in Phase 2. To do so, Phase 1 builds up an archive of \emph{interestingly} different game states, which we call ``cells'' (Section~\ref{sec:cell_representations}), and trajectories that lead to them. It starts with an archive that only contains the starting state.
From there, it builds the archive by repeating the following procedures: choose a cell from the current archive (Section~\ref{sec:selecting_cells}), return to that cell without adding any stochastic exploration (Section~\ref{sec:determinism}), and then explore from that location stochastically (Section~\ref{sec:explore_from_cell}). During this process, any newly encountered cells (as well as how to reach them) or improved trajectories to existing cells are added to the archive (Section~\ref{sec:update_the_archive}).

\subsubsection{Cell representations}
\label{sec:cell_representations}
\label{sec:no_domain_knowledge_representation}
\label{sec:domain_knowledge_representation}

One could, in theory, run Go-Explore directly in a high-dimensional state space (wherein each cell contains exactly one state); however doing so would be intractable in practice. To be tractable in high-dimensional state spaces like Atari, Phase 1 of Go-Explore needs a lower-dimensional space within which to search (although the final policy will still play in the same original state space, in this case pixels). Thus, the cell representation should conflate similar states while not conflating states that are meaningfully different.

In this way, a good cell representation should reduce the dimensionality of the observations into a meaningful low-dimensional space. A rich literature investigates how to obtain good representations from pixels. One option is to take latent codes from the middle of neural networks trained with traditional RL algorithms maximizing extrinsic and/or intrinsic motivation, optionally adding auxiliary tasks such as predicting rewards~\cite{Silver2017ThePE}. 
Additional options include unsupervised techniques such as networks that autoencode~\cite{lange2010deep} or predict future states, and other auxiliary tasks such as pixel control~\cite{Jaderberg2016ReinforcementLW}. 

While it will be interesting to test any or all of these techniques with Go-Explore in future work, for these initial experiments with Go-Explore we test its performance with two different representations: a simple one that does not harness game-specific domain knowledge, and one that does exploit easy-to-provide domain knowledge. 

\subsubsubsection{\textbf{Cell representations without domain knowledge}}

\begin{figure}
    \centering
    \includegraphics[width=\linewidth]{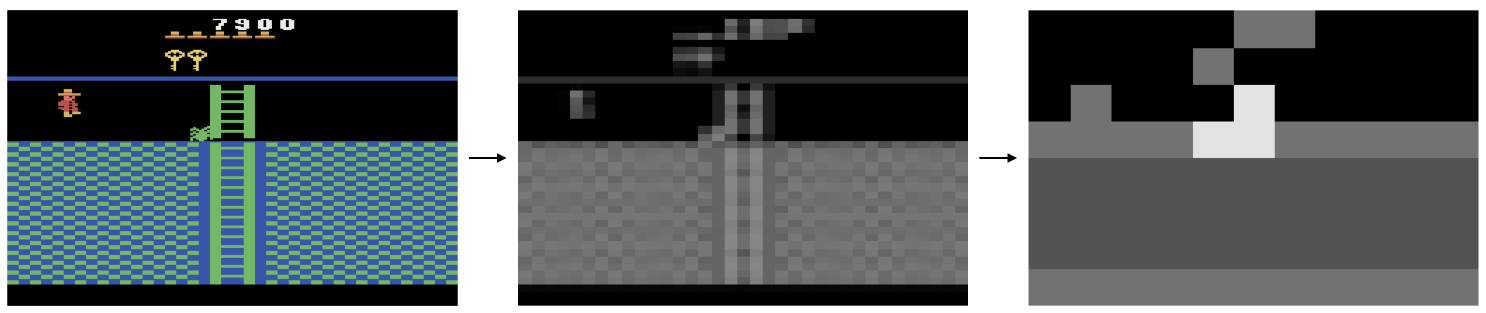}
    \caption{\textbf{Example cell representation without domain knowledge, which is simply to downsample each game frame.} The full observable state, a color image, is converted to grayscale and downscaled to an $11 \times 8$ image with 8 possible pixel intensities.}
    \label{fig:downscale_vis}
\end{figure}

We found that a very simple dimensionality reduction procedure produces surprisingly good results on Montezuma's Revenge.
The main idea is simply to downsample the current game frame. Specifically, we (1)~convert each game frame image to grayscale (2)~downscale it to an $11 \times 8$ image with area interpolation (i.e.\ using the average pixel value in the area of the downsampled pixel), (3)~rescale pixel intensities so that they are integers between 0 and 8, instead of the original 0 to 255 (Fig.~\ref{fig:downscale_vis}). The downscaling dimensions and pixel-intensity range were found by grid search. The aggressive downscaling used by this representation is reminiscent of the Basic feature set from \citet{bellemare2013arcade}. This cell representation requires no game-specific knowledge and is fast to compute. 

\subsubsubsection{\textbf{Cell representations with domain knowledge}}

The ability of an algorithm to integrate easy-to-provide domain knowledge can be an important asset. In Montezuma's Revenge, domain knowledge is provided as unique combinations of the $x,y$ position of the agent (discretized into a grid in which each cell is $16 \times 16$ pixels), room number, level number, and in which rooms the currently-held keys were found. In the case of Pitfall, only the $x,y$ position of the agent and the room number were used. All this information was extracted directly from pixels with simple hand-coded classifiers to detect objects such as the main character's location combined with our knowledge of the map structure in the two games (Appendix~\ref{sec:domain_from_pixels}). While Go-Explore provides the opportunity to leverage domain knowledge in the cell representation in Phase 1, the robustified neural network produced by Phase 2 still plays directly from pixels only.

\subsubsection{Selecting cells}
\label{sec:selecting_cells}

In each iteration of Phase 1, a cell is chosen from the archive to explore from. This choice could be made uniformly at random, but we can improve upon that baseline in many cases by creating (or learning) a heuristic for preferring some cells over others. In preliminary experiments, we found that such a heuristic can improve performance over uniform random sampling (data not shown).  The exact heuristic differs depending on the problem being solved, but at a high level, the heuristics in our work assign a positive weight to each cell that is higher for cells that are deemed more promising. For example, cells might be preferred because they have not been visited often, have recently contributed to discovering a new cell, or are expected to be near undiscovered cells. The weights of all cells are normalized to represent the probability of each cell being chosen next. No cell is ever given a weight equal to 0, so that all cells in principle remain available for further exploration. The exact heuristics from our experiments are described in Appendix~\ref{sec:selection_details}.

\subsubsection{Returning to cells and opportunities to exploit deterministic simulators}
\label{sec:determinism}

One of the main principles of Go-Explore is to return to a promising cell without added exploration before exploring from that cell. The Go-Explore philosophy is that we should make returning to that cell as easy as possible given the constraints of the problem. The easiest way to return to a cell is if the world is \emph{deterministic} and \emph{resettable}, such that one can reset the state of the simulator to a previous visit to that cell. Whether performing such resets is allowable for RL research is an interesting subject of debate that was motivated by the initial announcement of Go-Explore~\cite{ecoffet:bloggoexplore2018}. The ability to harness determinism and perform such resets forces us to recognize that there are two different types of problems we wish to solve with RL algorithms: those that require stochasticity at test time only, and those that require stochasticity during both testing and training. 

We start with the former. Because current RL algorithms can take unsafe actions~\cite{mcallister2018robustness,kahn2017uncertainty} and require tremendous amounts of experience to learn~\cite{horgan:apexdqn2018,espeholt:impala2018,Lake2017BuildingMT}, the majority of applications of RL in the foreseeable future will likely require training in a simulator before being transferred to (and optionally fine-tuned in) the real world. For example, most work with learning algorithms for robotics train in a simulator before transferring the solution to the real world; that is because learning directly on the robot is slow, sample-inefficient, can damage the robot, and can be unsafe~\cite{koos2013transferability,cully:nature15,andrychowicz2018learning}. Fortunately, for many domains, simulators are available (e.g.\ robotics simulators, traffic simulators, etc.). An insight of Go-Explore is that we can take advantage of the fact that such simulators can be made deterministic to improve performance, especially on hard-exploration problems. For many types of problems, we want a reliable final solution (e.g.\ a robot that reliably finds survivors after a natural disaster) and there is no principled reason to care whether we obtain this solution via initially deterministic training. If we can solve previously unsolvable problems, including ones that are stochastic at evaluation (test) time, via making simulators deterministic, we should take advantage of this opportunity.

There are also cases where a simulator is not available and where learning algorithms must confront stochasticity during training. To create and test algorithms for this second type of problem, we cannot exploit determinism and resettability. Examples of this class of problems include when we must learn directly in the real world (and an effective simulator is not available and cannot be learned), or when studying the learning of biological animals, including ourselves. We believe Go-Explore can handle such situations by training goal-conditioned policies~\cite{andrychowicz2017hindsight,schaul2015universal} that reliably return to cells in the archive during the exploration phase, which is an interesting area for future research. While computationally much more expensive, this strategy would result in a fully trained policy at the end of the exploration phase, meaning there would be no need for a robustification phase at the end. We note that there are some problems where the environment has forms of stochasticity that prevent the algorithm from reliably returning to a particular cell, regardless of which action the agent takes (e.g.\ in poker, there is no sequence of actions that reliably leads you to a state where you have two aces). We leave a discussion and study of whether Go-Explore helps in that problem setting for future work. 

With this distinction in mind, we can now ask whether Montezuma's Revenge and Pitfall represent the first type of domain (where all we care about is a solution that is robust to stochasticity at test time) or the second (situations where the algorithm must handle stochasticity while training). We believe few people in the community had considered this question before our initial blog post on Go-Explore~\cite{ecoffet:bloggoexplore2018} and that it created a healthy debate on this subject. Because Atari games are proxies for the problems we want to solve with RL, and because both types of problems exist, a natural conclusion is that we should have benchmarks for each. One version of a task can require stochasticity during testing only, and another can require stochasticity during both training and testing. All results and claims in this version of this paper are for the version of these domains that does not require stochasticity during training (i.e.\ stochasticity is required during evaluation only). Applying Go-Explore when training is stochastic remains an exciting avenue of research for the near future.

For problems in which all we care about is a reliable policy at test time, a key insight behind Go-Explore is that we can first solve the problem (Phase 1), and then (if necessary) deal with making the solution more robust later (Phase 2). In contrast with the usual view of determinism as a stumbling block to producing agents that are robust and high-performing, it can be made an ally during exploration and then the solution extended to nondeterminism afterwards via robustification. An important domain where such insights can help is robotics, where training is often done in simulation before policies are transferred to the real world~\cite{koos2013transferability,cully:nature15,andrychowicz2018learning}.

For the experiments in this paper, because we harness deterministic training, we could return to a cell by storing the sequence of actions that lead to it and subsequently replay those actions. However, simply saving the state of the emulator (in addition to this sequence of steps) and restoring that state when revisiting a cell gains additional efficiency. Doing so reduced the number of steps that needed to be simulated by at least one order of magnitude (Appendix~\ref{sec:performance}).

Due to the fact that the present version of Go-Explore operates in a deterministic setting during Phase~1, each cell is associated with an open-loop sequence of instructions that lead to it given the initial state, not a proper policy that maps any state to an action. A true policy is produced during robustification in Phase 2 (Section~\ref{sec:robustification}).

\subsubsection{Exploration from cells}
\label{sec:explore_from_cell}

Once a cell is reached, any exploration method can be applied to find new cells. In this work the agent explores by taking random actions for $k=100$ training frames, with a $95\%$ probability of repeating the previous action at each training frame (frames at which the agent is allowed to take an action, thus not including any frames skipped due to frame skip, see Appendix~\ref{sec:frame_definition}).
Besides reaching the $k = 100$ training frame limit for exploration, exploration is also aborted at the episode's end (defined in Appendix~\ref{sec:episode_end}), and the action that led to the episode ending is ignored because it does not produce a destination cell.

Interestingly, such exploration does not require a neural network or other controller, and indeed no neural network was used for the exploration phase (Phase 1) in any of the experiments in this paper (we do not train a neural network until Phase 2). The fact that entirely random exploration works so well highlights the surprising power of simply returning to promising cells before exploring further, though we believe exploring intelligently (e.g.\ via a trained policy) would likely improve our results and is an interesting avenue for future work.

\subsubsection{Updating the archive}
\label{sec:update_the_archive}

While an agent is exploring from a cell, the archive updates in two conditions. The first condition is when the agent visits a cell that was not yet in the archive (which can happen multiple times while exploring from a given cell). In this case, that cell is added to the archive with four associated pieces of metadata: (1)~how the agent got to that cell (here, a full trajectory from the starting state to that cell), (2)~the state of the environment at the time of discovering the cell (if the environment supports such an operation, which is true for the two Atari-game domains in this paper), (3)~the cumulative score of that trajectory, and (4)~the length of that trajectory. 

The second condition is when a newly-encountered trajectory is ``better'' than that belonging to a cell already in the archive. For the experiments below, we define a new trajectory as better than an existing trajectory when the new trajectory either has a higher cumulative score or when it is a shorter trajectory with the same score. In either case, the existing cell in the archive is updated with the new trajectory, the new trajectory length, the new environment state, and the new score. In addition, information affecting the likelihood of this cell being chosen (see Appendix~\ref{sec:selection_details}) is reset, including the total number of times  the cell has been chosen and the number of times the cell has been chosen since leading to the discovery of another cell. Resetting these values is beneficial when cells conflate many different states because a new way of reaching a cell may actually be a more promising stepping stone to explore from (so we want to encourage its selection).
We do not reset the counter that records the number of times the cell has been visited because that would make recently discovered cells indistinguishable from recently updated cells, and recently discovered cells (i.e. those with low visit counts) are more promising to explore because they are likely near the surface of our expanding sphere of knowledge.

Because cells conflate many states, we cannot assume that a trajectory from start state $A$ through cell $B$ to cell $C$ will still reach $C$ if we substitute a different, better way to get from $A$ to $B$; therefore, the better way of reaching a cell is not integrated into the trajectories of other cells that built upon the original trajectory. However, performing such substitutions might work with goal-conditioned or otherwise robust policies, and investigating that possibility is an interesting avenue for future work. 

\subsubsection{Batch implementation}

We implemented Phase 1 in parallel to take advantage of multiple CPUs (our experiments ran on a single machine with 22 CPU cores): at each step, a batch of $b$ cells is selected (with replacement) according to the rules described in Section~\ref{sec:selecting_cells} and Appendix~\ref{sec:selection_details}, and exploration from each of these cells proceeds in parallel for each. Besides using the multiple CPUs to run more instances of the environment, a high $b$ also saves time by recomputing cell selection probabilities less frequently, which is important as this computation accounts for a significant portion of run time as the archive gets large (though this latter factor could be mitigated in other ways in the future). Because the size of $b$ also has an indirect effect on the exploration behavior of Go-Explore (for instance, the initial state is guaranteed to be chosen $b$ times at the very first iteration), it is in effect a hyperparameter, whose values are given in Appendix~\ref{sec:hyperparameter_phase1}.

\subsection{Phase 2: Robustification}
\label{sec:robustification}

If successful, the result of Phase 1 is one or more high-performing trajectories. However, if Phase~1 of Go-Explore harnessed determinism in a simulator, such trajectories will not be robust to any stochasticity, which is present at test time. Phase 2 addresses this gap by creating a policy robust to noise via imitation learning, also called learning from demonstration (LfD). Importantly, stochasticity is added during Phase 2 so that the final policy is robust to the stochasticity it will face during its evaluation in the test environment. 
Thus the policy being trained has to learn how to mimic and/or perform as well as the trajectory obtained from the Go-Explore exploration phase while simultaneously dealing with circumstances that were not present in the original trajectory. Depending on the stochasticity of the environment, this adjustment can be highly challenging, but nevertheless is far easier than attempting to solve a sparse-reward problem from scratch.

While most imitation learning algorithms could be used for Phase 2, different types of imitation learning algorithms can qualitatively affect the resulting policy. LfD algorithms that try to closely mimic the behavior of the demonstration may struggle to improve upon it. For this reason, we chose an LfD algorithm that has been shown capable of improving upon its demonstrations: the Backward Algorithm from \citet{salimans2018learning}. It works by starting the agent near the last state in the trajectory, and then running an ordinary RL algorithm from there (in this case Proximal Policy Optimization (PPO)~\cite{Schulman2017ProximalPO}). Once the algorithm has learned to obtain the same or a higher reward than the example trajectory from that starting place near the end of the trajectory, the algorithm backs the agent's starting point up to a slightly earlier place along the trajectory, and repeats the process until eventually the agent has learned to obtain a score greater than or equal to the example trajectory all the way from the initial state. Note that a similar algorithm was discovered independently at around the same time by \citet{Resnick2018BackplayM}.

While this approach to robustification effectively treats the expert trajectory as a curriculum for the agent, the policy is only optimized to maximize its own score, and not actually forced to accurately mimic the trajectory. For this reason, this phase is able to further optimize the expert trajectories, as well as generalize beyond them, both of which we observed in practice in our experiments (Section~\ref{sec:results}). In addition to seeking a higher score than the original trajectory, because it is an RL algorithm with a discount factor that prizes near-term rewards more than those gathered later, it also has a pressure to improve the efficiency with which it collects rewards. Thus if the original trajectory contains unnecessary actions (like visiting a dead end and returning), such behavior could be eliminated during robustification (a phenomenon we also observed).

\subsection{Additional experimental and analysis details}
\label{sec:experimental_details}

Comparing sample complexity for RL algorithms trained on Atari games can be tricky due to the common usage of \emph{frame skipping}~\cite{such:arxiv17,Machado2018RevisitingTA}, wherein a policy only sees and acts every $n$th (here, 4) frame, and that action is repeated for intervening frames to save the computation of running the policy. Specifically, it can be ambiguous whether the frames that are skipped are counted (which we call ``game frames'') or ignored (which we call ``training frames'') when discussing sample complexity. In this work, we always qualify the word ``frame'' accordingly and all numbers we report are measured in \emph{game} frames. Appendix~\ref{sec:frame_definition} further details the subtleties of this issue.

Because the Atari games are deterministic by default, some form of stochasticity needs to be introduced to provide a stochastic test environment, which is desirable to make Atari an informative test bed for RL algorithms. Following previous work, we introduce stochasticity into the Atari environment with two previously employed techniques: \emph{random no-ops} and \emph{sticky actions}.

Random no-ops means that the agent is forced to take up to 30 no-ops (do nothing commands) at the start of the game. Because most Atari games run on a timer that affects whether hazards are present or not, or where different hazards, items, or enemies are located, taking a random number of no-ops puts the world into a slightly different state each time, meaning that fixed trajectories (such as the ones found by Go-Explore Phase 1) will no longer work. Random no-ops were first introduced by \citet{mnih:nature15}, and they were adopted as a primary source of stochasticity in most subsequent papers working in the Atari domain~\cite{mnih:nature15,van2016deep,wang2015dueling,schaul2015prioritized,van2016learning,salimans2017evolution,hester2017deep,gruslys2017reactor,Bellemare2017ADP,ODonoghue2018TheUB,Hessel2018RainbowCI,espeholt:impala2018,horgan:apexdqn2018,pohlen2018observe,aytar2018playing}.

While random no-ops prevent single, memorized trajectories from solving Atari games, the remainder of the game remains deterministic, meaning there is still much determinism that can be exploited. While several other forms of stochasticity have been proposed (e.g.\ humans restarts~\cite{nair2015massively}, random frame skips~\cite{brockman}, etc.), a particularly elegant form is sticky actions~\cite{Machado2018RevisitingTA}, where at each game frame there exists some probability of repeating the previous action instead of performing a newly chosen action. This way to introduce stochasticity is akin to how humans are not frame perfect, but may hold a button for slightly longer than they intended, or how they may be slightly late in pressing a button. Because Atari games have been designed for human play, the addition of sticky actions generally does not prevent a game from being solvable, and it adds some stochasticity to every state in the game, not just the start. Although our initial blog post~\cite{ecoffet:bloggoexplore2018} only included random no-ops, in this paper our robustification and all post-robustification test results are produced with a combination of both random no-ops and sticky actions. All algorithms we compare against in Section~\ref{sec:results} and in Appendix~\ref{sec:scores} likewise were tested with some form of stochasticity (in the form of no-ops, sticky actions, human starts, or some combination thereof), though it is worth noting that, unlike Go-Explore, most also had to handle stochasticity throughout training. Relevant algorithms that were tested in a deterministic environment are discussed in Section~\ref{sec:related_work}.

All hyperparameters were found by performing a separate grid-search for each experiment. The final, best performing hyperparameters are listed in Appendix~\ref{sec:hyperparameter_phase1}, tables~\ref{tab:hyper_no_domain} and~\ref{tab:hyper_domain}.
All confidence intervals given are 95\% bootstrapped confidence intervals computed using the pivotal (also known as empirical) method~\cite{Zoubir2007BootstrapMA}, obtained by resampling 10,000 times. Confidence intervals are reported with the following notation: $stat$ \pivotci{$lower$}{$upper$}\percci{$lower$}{$upper bound$} where $stat$ is the statistic (a mean unless otherwise specified). In graphs containing shaded areas, those areas indicate the 95\% percentile bootstrapped confidence interval of the mean, obtained by resampling 1,000 times. 
Graphs of the exploration phase (Phase 1) depict data at approximately every 4M game frames and graphs of the robustification phase (Phase 2) depict data at approximately every 130,000 game frames.

Because the robustification process can diverge even after finding a solution, the neural network at the end of training does not necessarily perform well, even if a high-performing solution was found at some point during this process. To retrieve a neural network that performs well regardless of when it was found, all robustification runs (Phase 2) produced a checkpoint of the neural network approximately every 13M game frames. Because the performance values recorded during robustification are noisy, we cannot select the best performing checkpoint from those performance values alone. As such, at the end of each robustification run, out of the checkpoints with the lowest \texttt{max\_starting\_point} (or close to it), a random subset of checkpoints (between 10 and 50) was tested to evaluate the performance of the neural network stored within that checkpoint. We test a random subset because robustification runs usually produce more successful checkpoints then we can realistically test. The highest-scoring checkpoint for each run was then re-tested to account for the selection bias inherent in selecting the best checkpoint. The scores from this final retest are the ones we report.

The neural network from each checkpoint is evaluated with random no-ops and sticky actions until at least 5 scores for each of the 31 possible starting no-ops (from 0 to 30 inclusive) are obtained. The mean score for each no-op is then calculated and the final score for the checkpoint is the grand mean of the individual no-op scores. Unless otherwise specified, the default time limit of 400,000 game frames imposed by OpenAI Gym~\cite{brockman} is enforced.

\section{Results}
\label{sec:results}

\subsection{Montezuma's Revenge}
\label{sec:mr_no_domain_knowledge}
\subsubsection{Without domain knowledge in the cell representation}

\FloatBarrier

\begin{figure}[htb]
    \begin{subfigure}[t]{.33\textwidth}
        \centering
        \includegraphics[width=\linewidth]{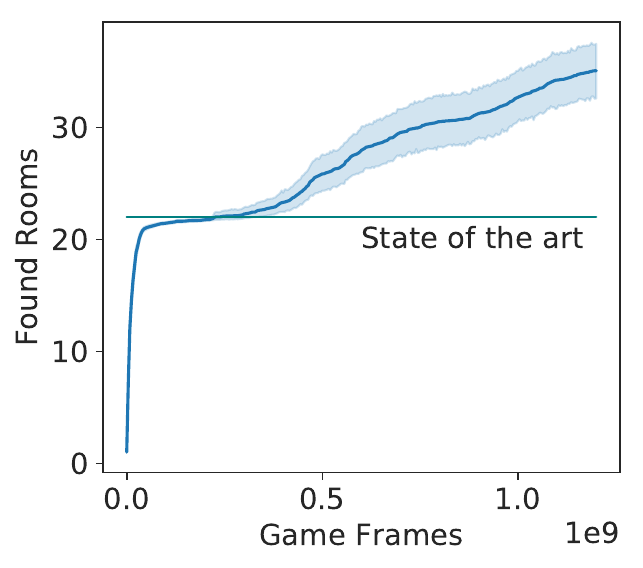}
        \caption{Number of rooms found}
        \label{fig:mont_no_domain_room}
    \end{subfigure}
    \begin{subfigure}[t]{.33\textwidth}
        \centering
        \includegraphics[width=\linewidth]{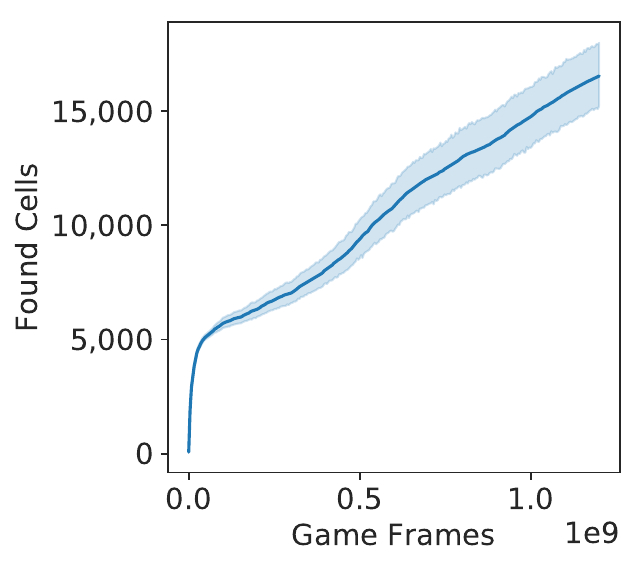}
        \caption{Number of cells found}
        \label{fig:mont_no_domain_cell}
    \end{subfigure}
    \begin{subfigure}[t]{.33\textwidth}
        \centering
        \includegraphics[width=\linewidth]{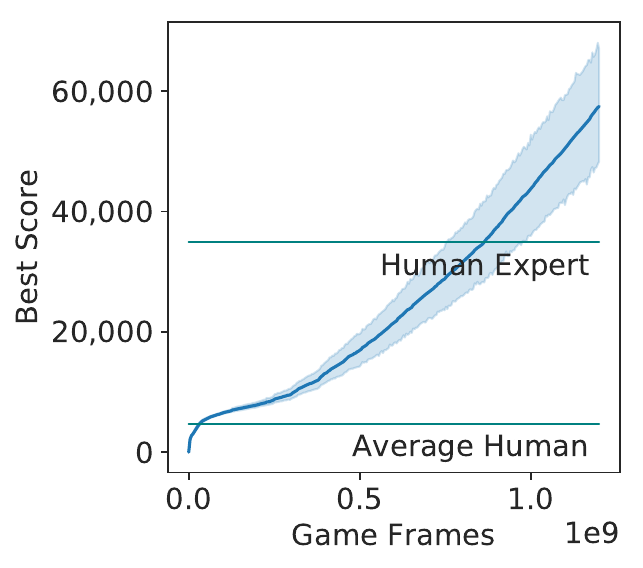}
        \caption{Maximum score in archive}
        \label{fig:mont_no_domain_score}
    \end{subfigure}
    \caption{\textbf{Performance of the exploration phase of Go-Explore with downscaled frames on Montezuma's Revenge.} Lines indicating human and the algorithmic state of the art are for comparison, but recall that the Go-Explore scores in this plot are on a deterministic version of the game (unlike the post-Phase 2 scores presented in this section).}
    \label{fig:mont_no_domain_overall}
\end{figure}

In this first experiment, we run Go-Explore on Montezuma's Revenge with the downsampled image cell representation, which does not require game-specific domain knowledge. Despite the simplicity of this cell representation, Phase 1 of Go-Explore solves level 1 in 57\% of runs after 1.2B game frames (a modest number by modern standards~\cite{horgan:apexdqn2018,espeholt:impala2018}), with one of the 100 runs also solving level 2, and visits a mean of 35 rooms \pivotci{33}{37}\percci{33}{38}
(Fig.~\ref{fig:mont_no_domain_room}). The number of new cells being discovered is still increasing linearly after 1.2B game frames, indicating that results would likely be even better were it run longer (Fig.~\ref{fig:mont_no_domain_cell}). Phase 1 of Go-Explore achieves a mean score of 57,439 \pivotci{47,843}{67,224}\percci{48,167}{67,364} (Fig.~\ref{fig:mont_no_domain_score}). Level 1 was solved after a mean of 640M \pivotci{567M}{711M}\percci{570M}{713M} game frames, which took a mean of 10.8 \pivotci{9.5}{12.0}\percci{9.5}{12.1} hours on a single, 22-CPU machine (note that these level 1 numbers exclude the runs that never solved level 1 after 1.2B game frames). See Appendix~\ref{sec:performance} for more details on performance.

Amusingly, Go-Explore discovered a little-known bug in Montezuma's Revenge called the ``treasure room curse''~\cite{atari_bugs}. If the agent performs a specific sequence of actions, it can remain in the treasure room (the final room before being sent to the next level) indefinitely, instead of being automatically moved to the next level after some time. Because gems giving 1,000 points keep appearing in the treasure room, it is possible to easily achieve very high scores once it has been triggered. Finding bugs in games and simulators, as Go-Explore did, is an interesting reminder of the power and creativity of optimization algorithms~\cite{Lehman2018TheSC}, and is commercially valuable as a debugging tool to identify and fix such bugs before shipping simulators and video games. A video of the treasure room curse  as triggered by Go-Explore is available at \url{https://youtu.be/civ6OOLoR-I}.

In 51 out of the 57 runs that solved level 1, the highest-scoring trajectory found by Go-Explore exploited the bug. To prevent scores from being inflated due to this bug, we filtered out trajectories that triggered the treasure room curse bug when extracting the highest scoring trajectory from each run of Go-Explore for robustification (Appendix~\ref{sec:filter_out_bug} provides details).

\begin{figure}
    \begin{subfigure}[t]{.495\textwidth}
        \centering
        \includegraphics[width=\textwidth]{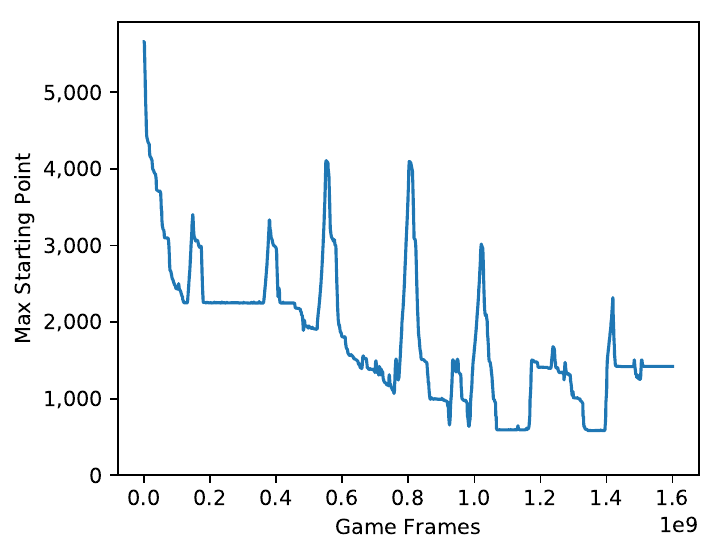}
        \caption{Failed robustification with 1 demonstration}
        \label{fig:robust_fail}
    \end{subfigure}
    \begin{subfigure}[t]{.495\textwidth}
        \centering
        \includegraphics[width=\textwidth]{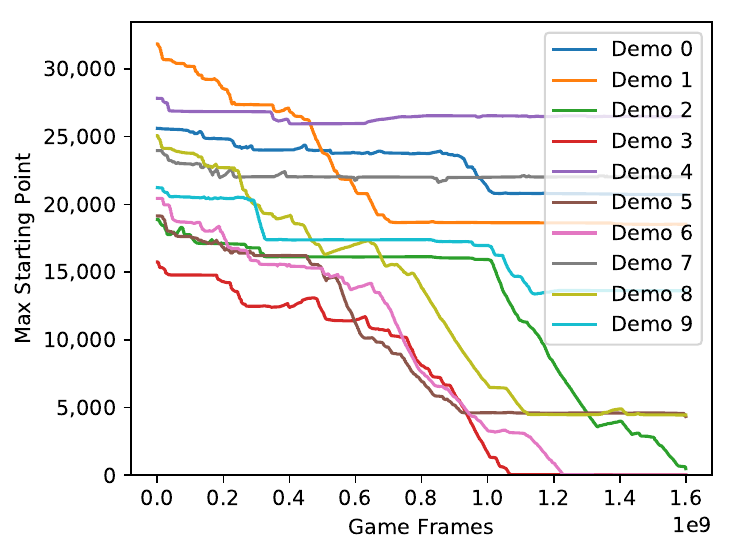}
        \caption{Successful robustification with 10 demonstrations}
        \label{fig:robust_multi}
    \end{subfigure}
    \caption{\textbf{Examples of maximum starting point over training for robustifying using different numbers of demonstrations.} Success is achieved as soon as \emph{any} of the curves gets sufficiently close (e.g. within 50 units) to 0, because that means the agent is able to perform as well as at least one of the demonstrations.}
    \label{fig:robust_overall}
\end{figure}

As mentioned in Section~\ref{sec:robustification}, we used Salimans \& Chen’s Backward Algorithm~\cite{salimans2018learning} for robustification. However, we found it somewhat unreliable in learning from a single demonstration (Fig.~\ref{fig:robust_fail}). Indeed, only $40\%$ of our attempts at robustifying trajectories that solved level 1 were successful when using a single demonstration.

However, because Go-Explore can produce many demonstrations, we modified the Backward Algorithm to simultaneously learn from multiple demonstrations (details in Appendix~\ref{sec:backward_mods}). To simulate the use case in which Phase 1 is run repeatedly until enough successful demonstrations (in this case 10) are found, we extracted the highest scoring non-bug demonstration from each of the 57 out of 100 Phase 1 runs that had solved level 1, and randomly assigned them to one of 5 non-overlapping groups of 10 demonstrations (7 demonstrations were left over and ignored), each of which was used for a robustification run. When training with 10 demonstration trajectories, all 5 robustification runs were successful. Fig.~\ref{fig:robust_multi} shows an example of successful robustification with 10 trajectories.

\begin{figure}
    \centering
    \includegraphics[width=0.8\linewidth]{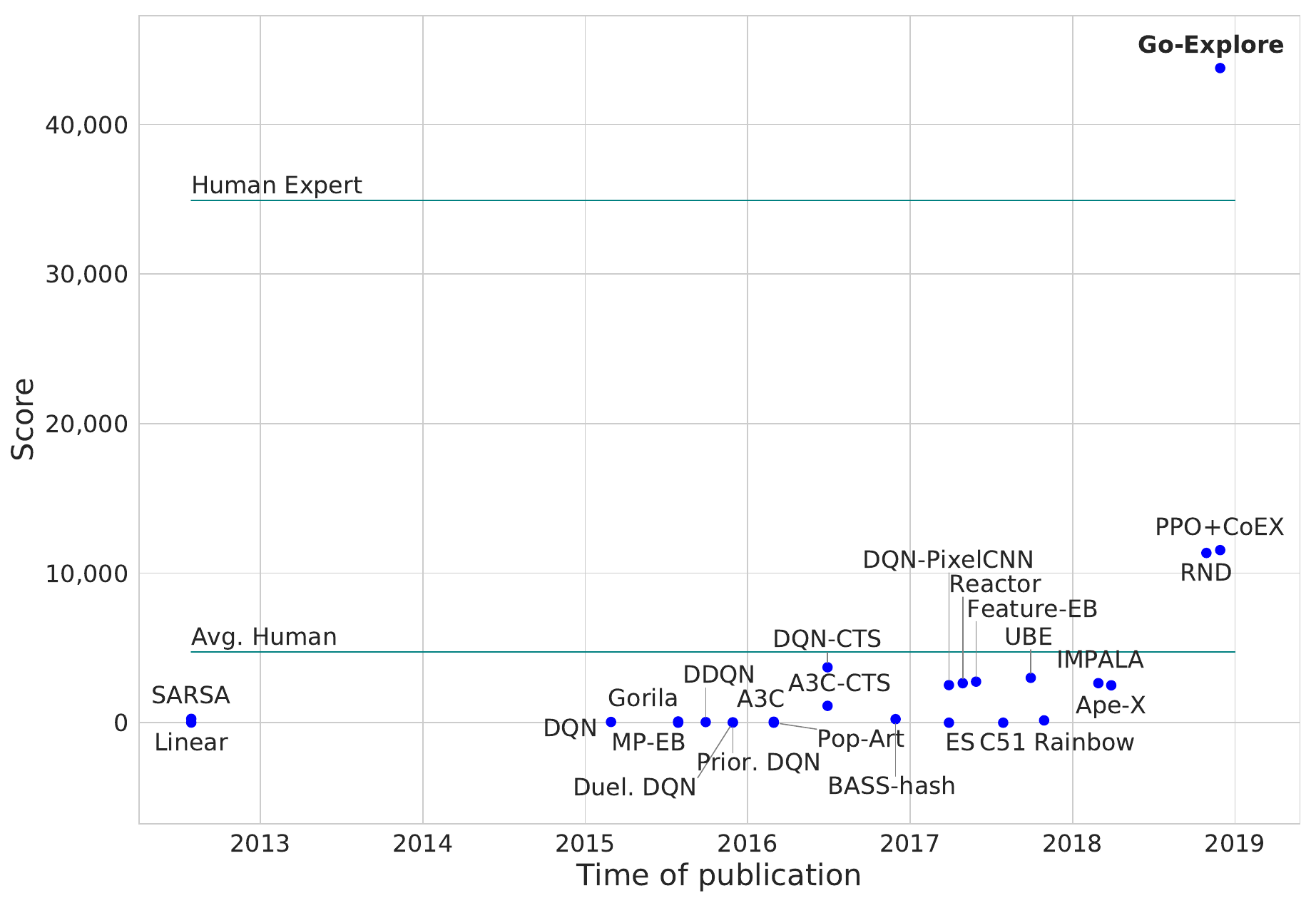}
    \caption{\textbf{History of progress on Montezuma's Revenge vs.\ the version of Go-Explore that does not harness domain knowledge.} Go-Explore significantly improves on the prior state of the art. These data are presented in tabular form in Appendix~\ref{sec:scores}.}
    \label{fig:mont_history_no_dom}
\end{figure}

In the end, our robustified policies achieve a mean score of 43,763 \pivotci{36,718}{50,196}\percci{37,379}{50,436}, substantially higher than the human expert mean of 34,900~\cite{pohlen2018observe}. All policies successfully solve level 1 (with a 99.8\% success rate over different stochastic evaluations of the policies), and one of our 5 policies also solves level 2 100\% of the time.
Fig.~\ref{fig:mont_history_no_dom} shows how these results compare with prior work.

Surprisingly, the computational cost of Phase 2 is greater than that of Phase 1. These Phase 2 results were achieved after a mean of 4.35B \pivotci{4.27B}{4.45B}\percci{4.25B}{4.43B} game frames of training, which took a mean of 2.4 \pivotci{2.4}{2.5}\percci{2.4}{2.5} days of training (details in Appendix~\ref{sec:performance}).

\subsubsection{With domain knowledge in the cell representation}
\label{sec:mr_domain_knowledge}

\begin{figure}
    \begin{subfigure}[t]{.33\textwidth}
        \centering
        \includegraphics[width=\linewidth]{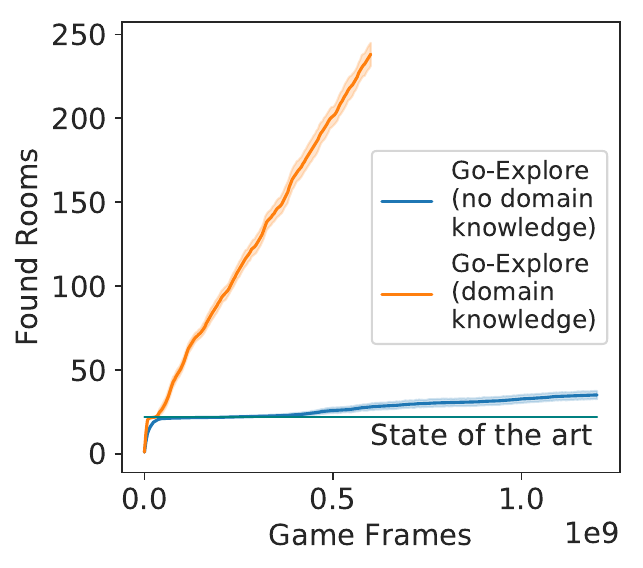}
        \caption{Number of rooms found}
        \label{fig:mont_domain_room}
    \end{subfigure}
    \begin{subfigure}[t]{.33\textwidth}
        \centering
        \includegraphics[width=\linewidth]{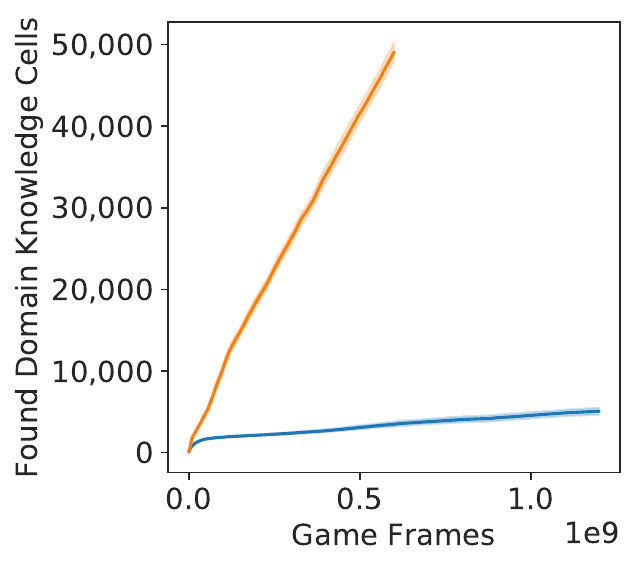}
        \caption{Number of cells found}
        \label{fig:mont_domain_cell}
    \end{subfigure}
    \begin{subfigure}[t]{.33\textwidth}
        \centering
        \includegraphics[width=\linewidth]{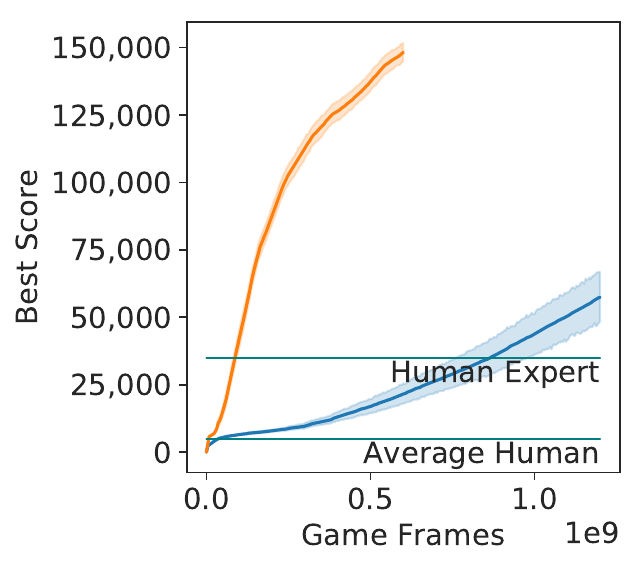}
        \caption{Maximum score in archive}
        \label{fig:mont_domain_score}
    \end{subfigure}
    \caption{\textbf{Performance on Montezuma's Revenge of Phase 1 of Go-Explore with and without domain knowledge.} The algorithm finds more rooms, cells, and higher scores with the easily provided domain knowledge, and does so with a better sample complexity. For (\subref{fig:mont_domain_cell}), we plot the number of cells found in the no-domain-knowledge runs according to the more intelligent cell representation from the domain-knowledge run to allow for an equal comparison.}
    \label{fig:mont_domain_overall}
\end{figure}

On Montezuma's Revenge, when harnessing domain knowledge in its cell representation (Section~\ref{sec:domain_knowledge_representation}), Phase 1 of Go-Explore finds a total of 238 \pivotci{231}{245}\percci{231}{245} rooms, solves a mean of 9.1 \pivotci{8.8}{9.4}\percci{8.8}{9.4} levels (with every run solving at least 7 levels), and does so in roughly half as many game frames as with the downscaled image cell representation (Fig.~\ref{fig:mont_domain_room}). Its scores are also extremely high, with a mean of 148,220 \pivotci{144,580}{151,730}\percci{144,756}{151,818} (Fig.~\ref{fig:mont_domain_score}). These results are averaged over 50 runs. 

As with the downscaled version, Phase 1 of Go-Explore with domain knowledge was still discovering additional rooms, cells, and ever-higher scores linearly when it was stopped (Fig.~\ref{fig:mont_domain_overall}). Indeed, because every level of Montezuma's Revenge past level 3 is nearly identical to level 3 (except for the scores on the screen and the stochastic timing of events) and because each run had already passed level 3, it would likely continue to find new rooms, cells, and higher scores forever.

Domain knowledge runs spend less time exploiting the treasure room bug because we preferentially select cells in the highest level reached so far (Appendix~\ref{sec:selection_details}). Doing so encourages exploring new levels instead of exploring the treasure rooms on previous levels to keep exploiting the treasure room bug. The highest final scores thus come from trajectories that solved many levels. Because knowing the level number constitutes domain knowledge, non-domain knowledge runs cannot take advantage of this information and are thus affected by the bug more.

In terms of computational performance, Phase 1 with domain knowledge solves the first level after a mean of only 57.6M \pivotci{52.7M}{62.3M}\percci{52.8M}{62.5M} game frames, corresponding to 0.9 \pivotci{0.8}{1.0}\percci{0.9}{1.0} hours on a single 22-CPU machine. Solving level 3, which effectively means solving the entire game as discussed above, is accomplished in a mean of 173M \pivotci{164M}{182M}\percci{165M}{182M} game frames, corresponding to 6.8 \pivotci{6.2}{7.3}\percci{6.2}{7.3} hours. Appendix~\ref{sec:performance} provides full performance details.

For robustification, we chose trajectories that solve level 3, truncated to the exact point at which level 3 is solved because, as mentioned earlier, all levels beyond level 3 are nearly identical aside from the pixels that display the score, which of course keep changing, and some global counters that change the timing of aspects of the game like when laser beams turn on and off.

We performed 5 robustification runs with demonstrations from the Phase 1 experiments above, each of which had a demonstration from each of 10 different Phase 1 runs. All 5 runs succeeded. The resulting mean score is 666,474 \pivotci{461,016}{915,557}\percci{414,902}{871,933}, far above both the prior state of the art and the non-domain knowledge version of Go-Explore. As with the downscaled frame version, Phase 2 was slower than Phase 1, taking a mean of 4.59B \pivotci{3.09B}{5.91B}\percci{3.26B}{6.05B} game frames, corresponding to a mean of 2.6 \pivotci{1.8}{3.3}\percci{1.9}{3.4} days of training.

The networks show substantial evidence of generalization to the minor changes in the game beyond level 3: although the trajectories they were trained on only solve level 3, these networks solved a mean of 49.7 levels \pivotci{32.6}{68.8}\percci{29.5}{66.7}. In many cases, the agents did not die, but were stopped by the maximum limit of 400,000 game frames imposed by default in OpenAI Gym~\cite{brockman}. Removing this limit altogether, our best single run from a robustified agent achieved a score of 18,003,200 and solved 1,441 levels during 6,198,985 game frames, corresponding to 28.7 hours of game play (at 60 game frames per second, Atari's original speed) before losing all its lives. This score is over an order of magnitude higher than the human world record of 1,219,200~\cite{atari_scoreboard}, thus achieving the strictest definition of ``superhuman'' performance. A video of the agent solving the first ten levels can be seen here: \url{https://youtu.be/gnGyUPd_4Eo}.

\begin{figure}
    \centering
    \includegraphics[width=0.8\linewidth]{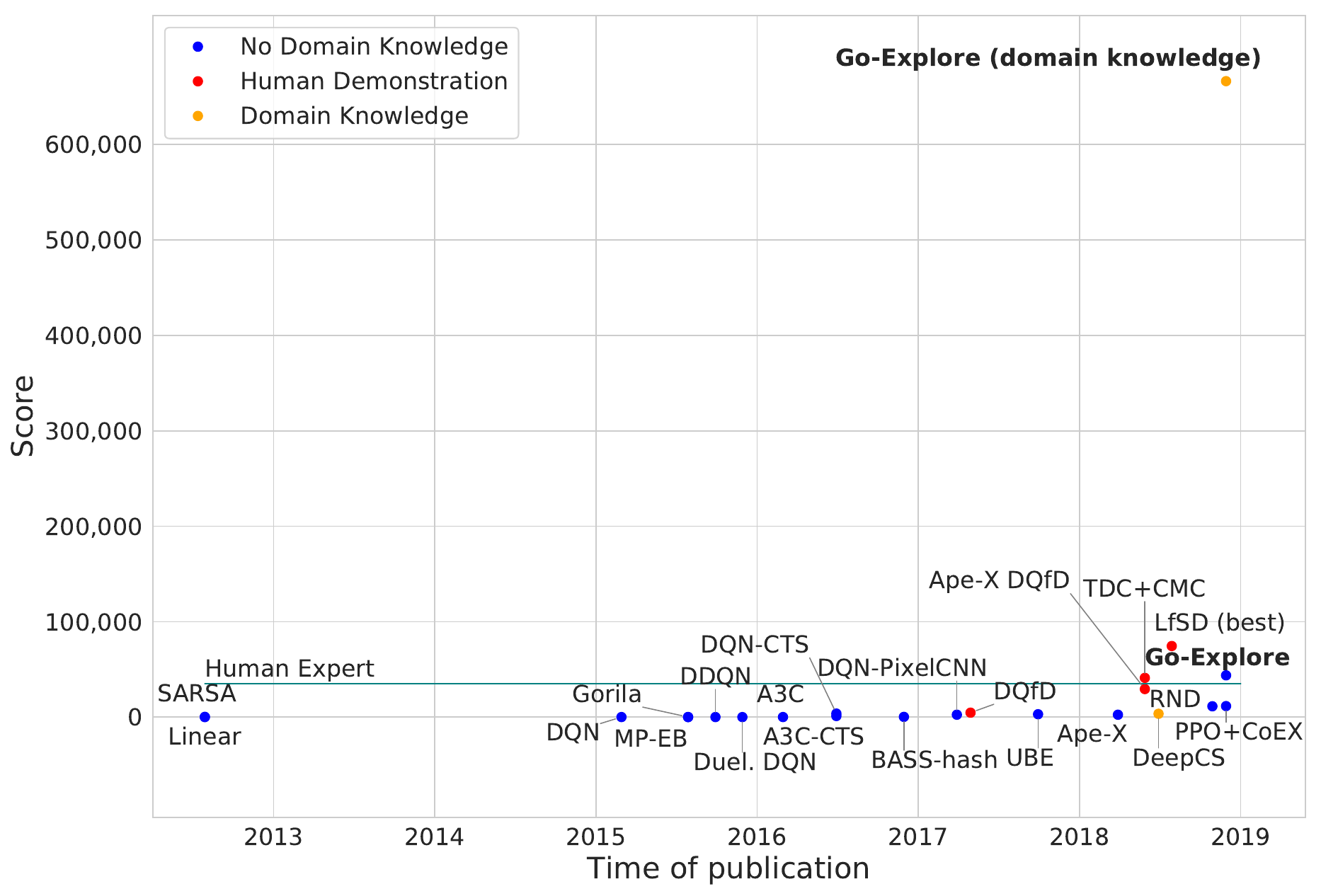}
    \caption{\textbf{Historical progress on Montezuma's Revenge vs. the version of Go-Explore that harnesses domain knowledge.} With domain knowledge, Go-Explore dramatically outperforms prior work, the no-domain-knowledge version of Go-Explore, and even prior work with imitation learning that was provided the solution in the form of human demonstrations. The data are presented in tabular form in Appendix~\ref{sec:scores}.}
    \label{fig:mont_history_dom}
\end{figure}

Fig.~\ref{fig:mont_history_dom} compares the performance of Go-Explore to historical results (including the previous state of the art), the no-domain-knowledge version of Go-Explore, and previous imitation learning work that relied on human demonstrations to solve the game. The version of Go-Explore that harnesses domain knowledge dramatically outperforms them all. Specifically, Go-Explore produces scores over 9 times greater than those reported for imitation learning from human demonstrations~\cite{salimans2018learning} and over 55 times the score reported for the prior state of the art 
without human demonstrations~\cite{Choi2018ContingencyAwareEI}.

That Go-Explore outperforms imitation learning plus human demonstrations is particularly noteworthy, as human-provided solutions are arguably a much stronger form of domain knowledge than that provided to Go-Explore. We believe that this result is due to the higher quality of demonstrations that Go-Explore was able to produce for Montezuma's Revenge vs.\ those provided by humans in the previous imitation learning work. The demonstrations used in our work range in score from 35,200 to 51,900 (lower than the final mean score of 148,220 for Phase 1 because these demonstrations are limited to only solving up to level 3) and most importantly, they all solve level 3. The demonstration originally used with the Backward Algorithm~\cite{salimans2018learning} reaches a score of 71,500 but doesn't solve level 3, thus preventing it from generalizing to further levels. The demonstrations used in DQfD and Ape-X DQfD~\cite{hester2017deep,pohlen2018observe} only range in score from 32,300 to 34,900. In this last case, it is not clear whether level 3 was solved in any of the demonstrations, but we believe this is unlikely given the reported scores because they are lower than the lowest level-3-solving scores found by Go-Explore and given the fact that the human demonstration used by the Backward Algorithm scored twice as high without solving level 3.

One interesting benefit of a robustification phase with an imitation learning algorithm that does not try to mimic the original demonstration is that it can improve upon that demonstration. Because of the discount on future rewards that exists in the base RL algorithm PPO, there is a pressure to remove inefficiencies in the demonstration. Videos of Go-Explore policies reveal efficient movements. In contrast, IM algorithms specifically reward reaching novel states, meaning that policies produced by them often do seemingly inefficient things like deviating to explore dead ends or jumping often to touch states only accessible by jumping, even though doing so is not necessary to gain real reward. An example of a Deep Curiosity Search agent~\cite{Stanton2018DeepCS} performing such inefficient jumps can be viewed at \url{https://youtu.be/-Fy2va3IbQU}, and a random network distillation~\cite{burda:rnd2018} IM agent can be viewed at \url{https://youtu.be/40VZeFppDEM}. These results suggest that IM algorithms could also benefit from a robustification phase in which they focus only on real-game reward once the IM phase has sufficiently explored the state space.

\FloatBarrier

\subsection{Pitfall}
\label{sec:pitfall_results}

\begin{figure}
    \begin{subfigure}[t]{.33\textwidth}
        \centering
        \includegraphics[width=\linewidth]{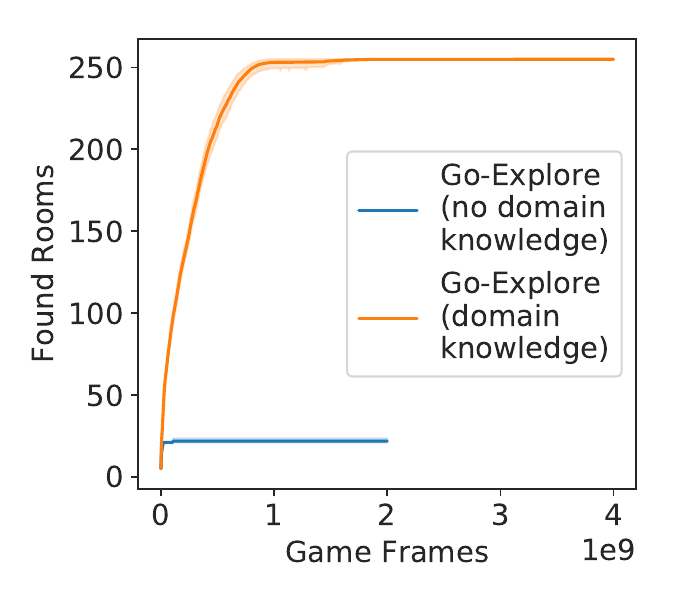}
        \caption{Number of rooms found}
        \label{fig:pitfall_domain_room}
    \end{subfigure}
    \begin{subfigure}[t]{.33\textwidth}
        \centering
        \includegraphics[width=\linewidth]{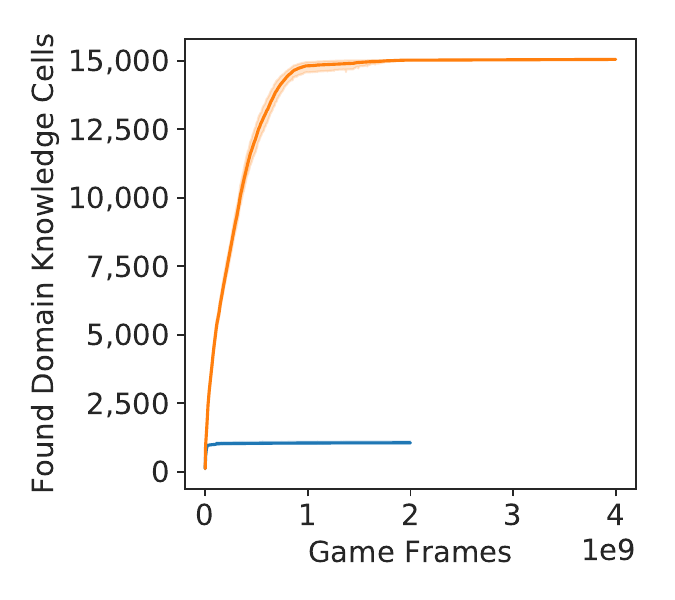}
        \caption{Number of cells found}
        \label{fig:pitfall_domain_cell}
    \end{subfigure}
    \begin{subfigure}[t]{.33\textwidth}
        \centering
        \includegraphics[width=\linewidth]{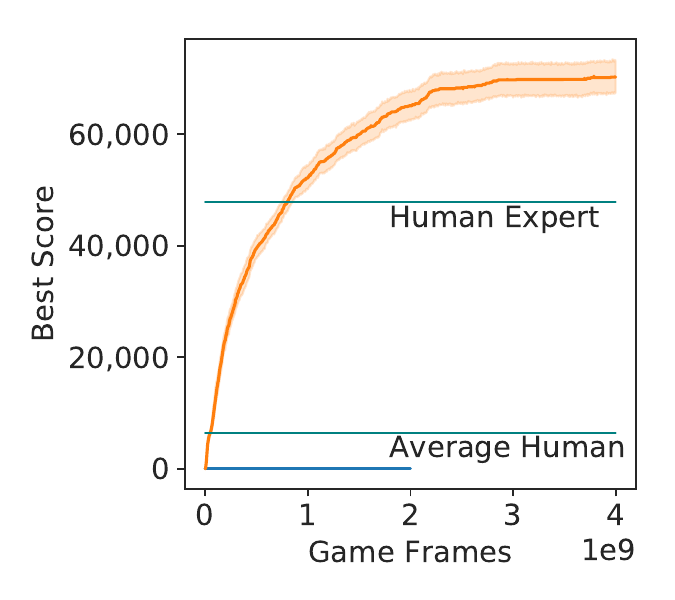}
        \caption{Maximum score in archive}
        \label{fig:pitfall_domain_score}
    \end{subfigure}
    \caption{\textbf{Performance on Pitfall of Phase 1 of Go-Explore with and without domain knowledge.}
    Without domain knowledge, the exploration phase finds about 22 rooms (a), but it then quickly stops finding new rooms (a) or cells (b) (here, we display discovery of domain-knowledge cells to enable a fair comparison, see Appendix~\ref{sec:pitfall_cell_representation} for progress on the domain-agnostic cell representation), and it doesn't find any rewards (c).
    With domain knowledge, the exploration phase of Go-Explore finds all 255 rooms (\subref{fig:pitfall_domain_room}) and trajectories scoring a mean 70,264 points (\subref{fig:pitfall_domain_score}).
    In addition, even though the number of rooms (a) and the number cells (b) found stagnates after about 2B game frames, score continues to go up for about another billion game frames. This is possible because, in Pitfall, there can exist many different trajectories to the same cell that vary in score. As such, once all reachable cells have been discovered, Go-Explore relies on replacing lower-scoring trajectories with higher-scoring trajectories to increase its score. The final score is not the maximum score that can be reached in Pitfall (the maximum score in Pitfall is 112,000), but Go-Explore finds itself in a local optima where higher scoring trajectories cannot be found starting from any of the trajectories currently in the archive.
    Lines represent the mean over 10 (without domain knowledge) and 40 (with domain knowledge) independent runs.}
    \label{fig:pitfall_explore}
\end{figure}

We next test Go-Explore on the harder, more deceptive game of Pitfall, for which all previous RL algorithms scored $\leq0$ points, except those that were evaluated on the fully deterministic version of the game~\cite{liu2019learning, Keramati2018FastEW} or relied on human demonstrations~\cite{hester2017deep, pohlen2018observe, aytar2018playing}. 
As with Montezuma's Revenge, we first run Go-Explore with the simple, domain-general, downscaled representation described in Section~\ref{sec:no_domain_knowledge_representation}, with the same hyperparameters. With these settings, Go-Explore is able to find 22 rooms, but it is unable to find any rewards (Fig.~\ref{fig:pitfall_explore}). We believe that this number of rooms visited is greater than the previous state of the art, but the number of rooms visited is infrequently reported so we are unsure.
In preliminary experiments, Go-Explore with a more fine-grained downscaling procedure (assigning 16 different pixel values to the screen, rather than just 8) is able to find up to 30 rooms, but it then runs out of memory (Appendix~\ref{sec:pitfall_cell_representation}). Perhaps with a more efficient or distributed computational setup this representation could perform well on the domain, a subject we leave to future work.
We did not attempt to robustify any of the trajectories because no positive reward was found.

We believe the downscaled-image cell representation underperforms on Pitfall because the game is partially observable, and frequently contains many importantly different states that appear almost identical (even in the unaltered observation space of the game itself), but require different actions (Appendix~\ref{sec:pitfall_nearly_identical_states}).
One potential solution to this problem would be to change to a cell representation that takes previous states into account to disambiguate such situations. Doing so is an interesting direction for future work.

Next, we tested Go-Explore with domain knowledge (Section~\ref{sec:domain_knowledge_representation}). The cell representation with domain knowledge is not affected by the partial observability of Pitfall because it maintains the room number, which is information that disambiguates the visually identical states (note that we can keep track of the room number from pixel information only by keeping track of all screen transitions that happened along the trajectory). With it, the exploration phase of Go-Explore (Phase 1) is able to visit all 255 rooms and its best trajectories collect a mean of 70,264 \percci{67,390}{73,180}\pivotci{67,287}{73,150} points (Fig.~\ref{fig:pitfall_explore}).

We attempted to robustify the best trajectories, but the full-length trajectories found in the exploration phase did not robustify successfully (Appendix~\ref{sec:pitfall_long_robustification_failure}), possibly because different behaviors may be required for states that are visually hard to distinguish (Appendix~\ref{sec:pitfall_nearly_identical_states}). Note that the domain-knowledge cell representation does not help in this situation, because the network trained in the robustification phase (Phase 2) is not presented with the cell representation from the exploration phase (Phase 1). The network thus has to learn to keep track of past information by itself. Remembering the past is possible, as the network of the agent does include a fully recurrent layer, but it is unclear to what degree this layer stores information from previous rooms, especially because the Backward Algorithm loads the agent at various points in the game without providing the agent with the history of rooms that came before. This can make it difficult for the agent to learn to store information from previous states. As such, robustifying these long trajectories remains a topic for future research.

\begin{figure}
    \centering
    \includegraphics[width=0.8\linewidth]{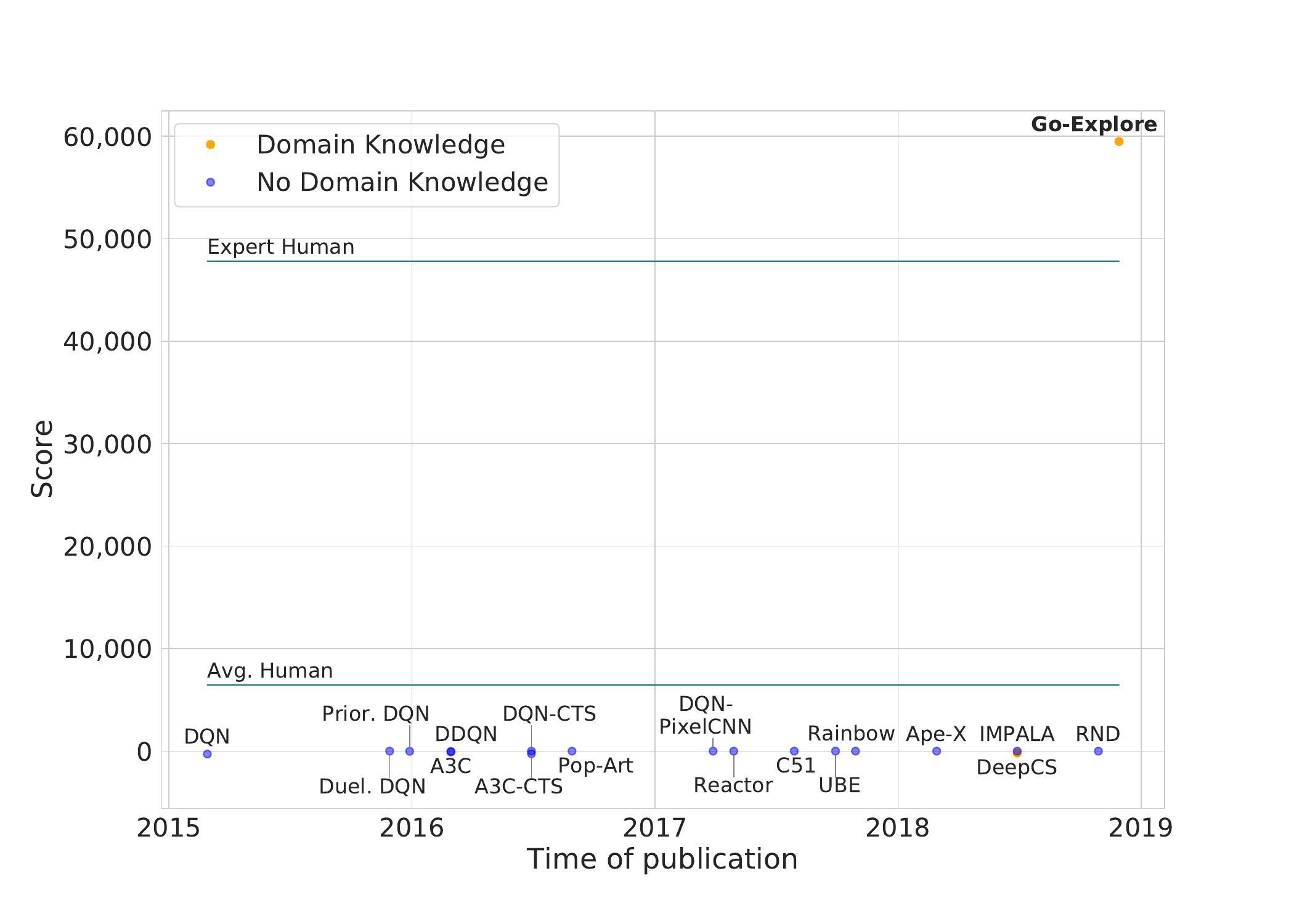}
    \caption{\textbf{Historical progress on Pitfall vs.\ the version of Go-Explore that harnesses domain knowledge.} Go-Explore achieves a mean of over 59,000 points, greatly outperforming the prior state of the art. The data are presented in tabular form in Appendix~\ref{sec:scores}.}
    \label{fig:pitfall_history}
\end{figure}

We found that shorter trajectories scoring roughly 35,824 \pivotci{34,225}{37,437}\percci{34,196}{37,429} points could be successfully robustified. To obtain these shorter trajectories, we truncated all trajectories in the archive produced in Phase 1 to 9,000 training frames (down from the total of 18,000 training frames), and then selected the highest scoring trajectory out of these truncated trajectories. We then further truncated this highest scoring trajectory such that it would end right after the collection of the last obtained reward, to ensure that the Backward Algorithm would always start right before obtaining a reward, resulting in trajectories with a mean length of 8,304 \pivotci{8,118}{8,507}\percci{8,100}{8,491} training frames.

From the truncated trajectories, the robustification phase (Phase 2) of Go-Explore is able to produce agents that collect 59,494 \pivotci{49,042}{72,721}\percci{46,044}{69,987} points (mean over 10 independent runs), substantially outperforming both the prior state of the art and human experts (Fig.~\ref{fig:pitfall_history}). These trajectories required a mean of 8.20B \pivotci{6.73B}{9.74B}\percci{6.63B}{9.70B} game frames to robustify, which took a mean of 4.5 \pivotci{3.7}{5.3}\percci{3.6}{5.3} days. The best rollout of the best robustified policy obtained a score of 107,363 points, and a video of this rollout is available at: \url{https://youtu.be/IJMdYOnsDpA}.

Interestingly, the mean performance of the robustified networks of 59,494 is higher than the maximum performance among the demonstration trajectories of 45,643. This score difference is too large to be the result of small optimizations along the example trajectories (e.g.\ by avoiding more of the negative rewards in the environment), thus suggesting that, as with Montezuma's Revenge, these policies are able to generalize well beyond the example trajectories they were provided.

\section{Discussion and Future Work}
\label{sec:discussion_and_future_work}

Three key principles enable Go-Explore to perform so well on hard-exploration problems: (1)~remember good exploration stepping stones, (2)~first return to a state, then explore and, (3)~first solve a problem, then robustify (if necessary). 

These principles do not exist in most RL algorithms, but it would be interesting to weave them in. As discussed in Section~\ref{intro}, contemporary RL algorithms do not do follow principle 1, leading to \emph{detachment}. Number 2 is important because current RL algorithms explore by randomly perturbing the parameters or actions of the current policy in the hope of exploring new areas of the environment, which is ineffective when most changes break or substantially change a policy such that it cannot first return to hard-to-reach states before further exploring from them (an issue we call \emph{derailment}). 
Go-Explore solves this problem by first returning to a state and then exploring from there. Doing so enables deep exploration that can find a solution to the problem, which can then be robustified to produce a reliable policy (principle number 3). 

The idea of preserving and exploring from stepping stones in an archive comes from the quality diversity (QD) family of algorithms (like MAP-elites~\cite{cully:nature15,mouret:arxiv15} and novelty search with local competition~\cite{lehman:gecco11}), and Go-Explore is an enhanced QD algorithm based on MAP-Elites. However, previous QD algorithms focus on exploring the space of behaviors by randomly perturbing the current archive of policies (in effect departing from a stepping stone in policy space rather than in state space), as opposed to explicitly exploring state space by departing to explore anew from precisely where in state space a previous exploration left off. In effect, Go-Explore offers significantly more controlled exploration of state space than other QD methods by ensuring that the scope of exploration is cumulative through state space as each new exploratory trajectory departs from the endpoint of a previous one.

It is remarkable that the current version of Go-Explore works by taking entirely random actions during exploration (without any neural network) and that it is effective even when applied on a very simple discretization of the state space. Its success despite such surprisingly simplistic exploration strongly suggests that remembering and exploring from good stepping stones is a key to effective exploration, and that doing so even with otherwise naive exploration helps the search more than contemporary deep RL methods for finding new states and representing those states. Go-Explore might be made even more powerful by combining it with effective, learned representations. It could further benefit from replacing the current random exploration with more intelligent exploration policies, which would allow the efficient reuse of skills required for exploration (e.g.\ walking). Both of these possible improvements are promising avenues for future work.

Go-Explore also demonstrates how exploration and dealing with environmental stochasticity are problems that can be solved separately by first performing exploration in a deterministic environment and then robustifying relevant solutions. The reliance on having access to a deterministic environment may initially seem like a drawback of Go-Explore, but we emphasize that deterministic environments are available in many popular RL domains, including videos games, robotic simulators, or even learned world models. Once a brittle solution is found, or especially a diverse set of brittle solutions, a robust solution can then be produced in simulation. If the ultimate goal is a policy for the real world (e.g.\ in robotics), one can then use any of the many available techniques for transferring the robust policy from simulation to the real world~\cite{cully:nature15,koos2013transferability,tobin2017domain}. In addition, we expect that future work will demonstrate that it is possible to substitute exploiting determinism to return to states with a goal-conditioned policy~\cite{andrychowicz2017hindsight,schaul2015universal} that learns to deal with stochastic environments from the start (during training). Such an algorithm would still benefit from the first two principles of Go-Explore, and possibly the third too, as even a goal-conditioned policy could benefit from additional optimization once the desired goal is known.

A possible objection is that, while this method already works in the high-dimensional domain of Atari-from-pixels, it might not scale to truly high-dimensional domains like simulations of the real world. We believe Go-Explore can be adapted to such high-dimensional domains, but it will likely have to marry a more intelligent cell representation of interestingly different states (e.g.\ learned, compressed representations of the world) with intelligent (instead of random) exploration. Indeed, the more conflation (mapping more states to the same cell) one does, the more probable it is that one will need intelligent exploration to reach such qualitatively different cells.

Though our current implementation of Go-Explore can handle the deceptive reward structure found in Pitfall, its exploitation of determinism makes it vulnerable to a new form of deception we call the ``busy-highway problem.'' Consider an environment in which the agent needs to cross a busy highway. One option is to traverse the highway directly on foot, but that creates so much risk of being hit by a car that no policy could reliably cross this way. A safer alternative would be to take a bridge that goes over the highway, which would constitute a detour, but be guaranteed to succeed. By making the environment deterministic for Phase 1, the current version of Go-Explore would eventually succeed in traversing the highway directly, leading to a much shorter trajectory than by taking the bridge. Thus all the solutions chosen for robustification will be ones that involve crossing the highway directly instead of taking the bridge, making robustification impossible.

One solution to this issue would be to provide robustification with more demonstrations from Phase 1 of Go-Explore (which could include some that take the bridge instead of crossing the highway), or even all of the trajectories it gathers during Phase 1. With this approach, robustification would be able to fall back on the bridge trajectories when the highway trajectories fail to robustify. While this approach should help, it may still be the case that so much of the experience gathered by Go-Explore Phase 1 is dependent on trajectories that are impossible to reproduce reliably that learning from these Go-Explore trajectories is less efficient than learning from scratch. How common this class of problem is in practice is an empirical question and an interesting subject for future work. However, we hypothesize that versions of Go-Explore that deal with stochasticity throughout training (e.g.\ by training goal-conditioned policies to return to states) would not be affected by this issue, as they would not succeed in crossing the highway reliably except by taking the bridge.

One promising area for future work is robotics. Many problems in robotics, such as figuring out the right way to grasp an object, how to open doors, or how to locomote, are hard-exploration problems. Even harder are tasks that require long sequences of actions, such as asking a robot to find survivors, clean a house, or get a drink from the refrigerator. Go-Explore could enable a robot to learn how to do these things in simulation. 
Because conducting learning in the real world is slow and may damage the robot, most robotic work already involves first optimizing in a simulator and then transferring the policy to the real world~\cite{cully:nature15, lee2013evolving, andrychowicz2018learning, koos2013transferability}. Go-Explore's ability to exploit determinism can then be helpful because robotic simulators could be made deterministic for Phase 1 of Go-Explore. 
The full pipeline could look like the following: (1)~Solve the problem in a deterministic simulator via Phase 1 of Go-Explore. (2)~Robustify the policy in simulation by adding stochasticity to the simulation via Phase 2 of Go-Explore. (3)~Transfer the policies to the real world, optionally adding techniques to help cross the simulation-reality gap~\cite{cully:nature15, andrychowicz2018learning, koos2013transferability}, including optionally further learning via these techniques or any learning algorithm. Of course, this pipeline could also be changed to using a goal-conditioned version of Go-Explore if appropriate. Overall, we are optimistic that Go-Explore may make many previously unsolvable robotics problems solvable, and we are excited to see future research in this area from our group and others. 

Interestingly, the Go-Explore algorithm has implications and applications beyond solving sparse- or deceptive-reward problems. The algorithm's ability to broadly explore the state space can unearth important facets of the domain that go beyond reward, e.g.\ the distribution of states that contain a particular agent (e.g.\ a game character or robot) or are near to catastrophic outcomes. For example, within AI safety~\cite{Amodei2016ConcretePI} one open problem is that of safe exploration~\cite{garcia2015comprehensive}, wherein the process of training an effective real-world policy is constrained by avoiding catastrophe-causing actions during that training. In the robotics setting where Go-Explore is applied in simulation (before attempting transfer to the real world), the algorithm could be driven explicitly to search for diverse simulated catastrophes (in addition to or instead of reward). Such a catastrophe collection could then be leveraged to train agents that act more carefully in the real world, especially while learning~\cite{lipton2016combating,saunders2018trial}. Beyond this example, there are likely many other possibilities for how the data produced by Go-Explore could be productively put to use (e.g.\ as a source of data for generative models, to create auxiliary objectives for policy training, or for understanding other agents in the environment by inverse reinforcement learning).

\section{Related Work}
\label{sec:related_work}

Go-Explore is reminiscent of earlier work that separates exploration and exploitation (e.g.\ \citet{colas:geppg2018}), in which exploration follows a reward-agnostic Goal Exploration Process~\cite{forestier2017intrinsically} (an algorithm similar to novelty search~\cite{lehman}), from which experience is collected to prefill the replay buffer of an off-policy RL algorithm, in this case DDPG~\cite{Lillicrap2015ContinuousCW}. This algorithm then extracts the highest-rewarding policy from the experience gathered.
In contrast, Go-Explore further decomposes exploration into three elements: Accumulate stepping stones (interestingly different states), return to promising stepping stones, and explore from them in search of additional stepping stones (i.e.\ principles 1 and 2 above). The impressive results Go-Explore achieves by slotting in very simple algorithms for each element shows the value of this decomposition. 

The aspect of Go-Explore of first finding a solution and then robustifying around it has precedent in Guided Policy Search~\cite{levine2016end}. However, this method requires a non-deceptive, non-sparse, differentiable loss function to find solutions, meaning it cannot be applied directly to problems where rewards are discrete, sparse, or deceptive, as both Atari and many real-world problems are. Further, Guided Policy Search requires having a differentiable model of the world or learning a set of local models, which to be tractable requires the full state of the system to be observable during training time. 

More recently, \citet{Oh2018SelfImitationL} combined A2C with a ``Self-Imitation Learning'' loss on the best trajectories found during training. This is reminiscent of Go-Explore's robustification phase, except for the fact that Self-Imitation Learning's imitation loss is used throughout learning, while imitation learning is a separate phase in Go-Explore. Self-Imitation Learning's 2,500 point score on Montezuma's Revenge was close to the state of the art at the time of its publication.

Another algorithm that is related to the idea of first returning before exploring is Bootstrapped DQN~\cite{Osband2016DeepEV}. It trains an ensemble of networks that approximate the Q function, but with bootstrapping the data so each network is trained on a different random subset of the data. Each training episode, it picks one of the networks and acts according to the policy it implies. In frequently visited areas of the search space, all of the networks will have lots of data and are likely to converge to the same policy (thus, exploration will be low). However, in rarely visited areas of the state space, the networks would ideally have different Q-value predictions, meaning that in different episodes different choices will be made, yielding exploration. At a high level, the dynamics can thus allow an agent to first return to an area of the search space with little exploration before exploring from it. That said, this algorithm will still try to focus on returning to one narrow area of the search space (the one it is currently exploring, see the flashlight metaphor of IM algorithms in Section \ref{intro}) before exploring, and thus is still likely to suffer from the issue of detachment described in Section~\ref{intro}. Indeed, empirically Bootstrapped DQN scores only 100 on Montezuma's Revenge, and detachment may be a large reason why.

Recall Traces~\cite{Goyal2018RecallTB} also implement the idea of returning to previously discovered states. They do so by running a backtracking model to create virtual trajectories towards states heuristically considered valuable and they include those virtual trajectories during training with the help of an imitation learning loss, thereby increasing the likelihood that these states will be revisited and explored from. Contrary to Go-Explore, Recall Traces do not separate returning to states and exploring from those states, thus the algorithm helps ameliorate detachment, but not derailment. The method improved sample efficiency in several sparse reward domains, but was not tested on Montezuma's Revenge or Pitfall.

Closely related to the first two principles of Go-Explore is the work by \citet{liu2019learning}, which takes a hierarchical reinforcement learning approach in which an abstract MDP is created through the conflation of multiple states into abstract states, which are similar to the cells in Go-Explore.
This abstract MDP stores all abstract states (i.e.~cells) that it encounters, thus keeping track of promising states to explore from, and it navigates the MDP in a reliable way before exploring from a selected abstract-MDP state, thus implementing the idea of returning before exploring. One difference with Go-Explore is that this algorithm does not use a trajectory of actions to return to a cell, but instead relies on a set of sub-policies, called skills, which are executed in sequence to navigate the abstract MDP. While this set of skills is flexible, in that it allows the same skill to be reused for different transitions, it takes time to train a new skill, potentially making it computationally expensive to explore as deep into the game as Go-Explore does. Another difference is that the algorithm by \citet{liu2019learning} does not implement a robustification phase, but instead relies on the abstract MDP, even at evaluation time. While this means the algorithm does not require any additional training, it also means the algorithm can never improve upon the limits of the constructed MDP.
The algorithm from \citet{liu2019learning}, which harnesses domain knowledge, scores 12,500 on Montezuma's Revenge and 15,000 on Pitfall, though these scores come from evaluation in the deterministic version of the environment (they do provide results on stochastic test environments for a different game: Private Eye).
Go-Explore scores substantially more in both Montezuma's Revenge and Pitfall despite being tested in a stochastic environment and, in the case of Montezuma's Revenge, even when not relying on domain knowledge. 

In a similar vein, \citet{dong2019explicit} maintains an explicit memory of novel states and explores after returning to them via a goal-conditioned policy, though their algorithm only reaches scores of around 1,000 on Montezuma's Revenge, substantially less than Go-Explore. We speculate that this is due to (1) Its use of a fixed-capacity pool of potential next states to visit, which might not be able to keep up with the large number of possible interestingly different states present in Montezuma's Revenge, and (2) By determining whether a goal is reached based on a pixel based measure, their goal-conditioned policy could have a hard time learning to return to a previously visited state, as the pixel-based match requires all moving objects, such as enemies, to be in very similar locations before a goal is considered reached.
The insights of keeping an archive of known states and exploring to discover new states to add to the archive dates back at least to the $E^3$ algorithm~\cite{kearns2002near}, although the $E^3$ authors note that it does not work in high-dimensional problems for which tabular methods are intractable and function approximation (or some form of conflation) is required. Go-Explore can be seen as an $E^3$-like algorithm that adapts some of its principles to high-dimensional domains.

The idea of planning (searching in a deterministic model of the world to find a good strategy) and then training a policy to mimic what was learned is reminiscent of \citet{Guo2014DeepLF}. It plans (in the Atari emulator) with UCT~\cite{Kocsis2006BanditBM,Kocsis2006ImprovedMS,Browne2012ASO}, which is slow, and then trains a much faster policy with supervised learning to imitate the planning algorithm. At first glance it seems that in \citet{Guo2014DeepLF} UCT serves a similar role to the exploration phase in Go-Explore, but UCT is quite different in several ways that make it inferior for domains that are either high-dimensional or hard-exploration. That is true even though UCT does have a form of exploration bonus.

UCT plans in a model of the world so as to decide on the next action to take in the real environment. An exploration bonus is used during the planning phase, but only extrinsic rewards are considered when choosing the next action to take. This approach can improve performance in domains with relatively dense rewards, but fails in sparse rewards domains as rewards are likely to be beyond the planning horizon of the algorithm. Once planning what to do from one state is done, an action is taken and the planning process is run again from the next state. UCT does not try to explore all states, and each run of UCT is independent of which states were visited in previous planning steps. As such, UCT (either within an episode, or across episodes) does not try to discover new terrain: instead its exploration bonus only helps it within the current short-horizon planning phase. As mentioned in Section~\ref{intro}, UCT scores 0 on Montezuma's Revenge and Pitfall~\cite{bellemare2013arcade,Lipovetzky2015ClassicalPW}.

Another approach to planning is Fractal Monte Carlo (FMC)~\cite{Cerezo2018FractalAA}. When choosing the next action, it takes into account both the expected reward and novelty of that action, and in that way is more similar to Go-Explore. In FMC, a planning process is initiated from each state the agent visits. Planning is done within a deterministic version of the game emulator. A fixed number of workers are started in the state from which planning is occurring, and they perform random walks in state space. Periodically, workers that have accumulated lower reward and/or are in less novel states are replaced by ``clones'' of more successful workers. Novelty is approximated as the Euclidean distance of the worker's state (in the original, raw, observation space) to that of a randomly selected other worker. 

FMC reaches a score of 5,600 on Montezuma's Revenge, substantially higher than UCT. We believe this increased performance is due to at least three factors: (1) its planning process puts more emphasis on depth than breadth due to its finite amount of workers as opposed to the exponential branching factor that UCT needs to handle; (2) it favors novel states within a planning iteration, so actions that lead to hard-to-reach states such as jumping an enemy are more likely to be chosen; (3) having an exploration bonus based on Euclidean distance is more informative than UCT's exact-match state bonus, because more distant states are recognized as being more novel than states that differ by, say, one pixel. One major reason we believe FMC performs worse than Go-Explore is because, like UCT, it restarts its planning process from scratch each time an action is taken. That means it can cycle indefinitely between the same few states, because it does not have a means over time of remembering which states it has visited in order to attempt to explore all states, and instead must rely on random chance to break out of cycling. This phenomenon is apparent when watching its agent play: \url{https://youtu.be/FgaXa0uCBR4}. 
Although its greater focus on depth rather than breadth versus UCT extends its planning horizon enough to reach the first few rewards available in Montezuma's Revenge, that seemingly was insufficient for it to reach the even sparser rewards found later in the game that are easily found by Go-Explore.

On Pitfall, SOORL~\cite{Keramati2018FastEW} was the first planning algorithm to achieve a non-zero score, but did so in a deterministic test environment. It does so through a combination of learning a model of the environment, domain knowledge, and a value function that is optimistic about the value of unseen states, thus effectively providing an exploration bonus. At the end of 50 episodes of training, which was the maximum reported number of episodes, SOORL achieves an average of about 200 points across runs, and its best run scored an average of 606.6 with a maximum of 4,000.

Another way to view Phase 1 of Go-Explore is as being similar to a graph-search algorithm over nodes that are made up of the conflated states, and with unknown edges between the different nodes, meaning that nodes can never fully be marked as ``closed''. Specifically, the algorithm has to empirically discover the existence of an edge between two nodes, for example by executing a sequence of random actions that leads from one node to another node, and, as a result, it is never clear whether a node is closed because it is always possible that additional edges from this node exist, but that they have not been discovered yet. Prioritizing which nodes to explore by assigning a weight to them is reminiscent of graph-search algorithms such as Dijkstra's algorithm~\cite{Dijkstra:1959:NTP:2722880.2722945} and A*~\cite{Hart:astar}. Graph-search algorithms as a means of exploration in planning have been investigated in algorithms such as Rapidly-exploring Random Trees (RRTs)~\cite{LaValle1998RapidlyExploringRT}, which were recently used to explore Atari games by~\citet{zhan2018taking}. Indeed, Go-Explore exhibits important similarities with RRTs as they both keep track of an archive of states and trajectories to those states. However, there are some crucial differences, including: (1) RRTs proceed by first sampling a goal to attempt to reach, which can be impractical in environments where reachable states are not known a priori (and which is particularly pernicious in high-dimensional state spaces, such as pixels or even learned encodings, where most randomly selected goals are unreachable), such as Atari, and (2) RRTs do not have the concept of ``cells'' present in Go-Explore and thus RRTs can add many very similar states to their archive that do little to help the algorithm reach meaningfully different unexplored areas of the search space.
In general, we believe that Go-Explore points to an interesting future research direction in adapting the principles behind graph-search algorithms to high dimensional state spaces.

Even more distantly related are the many variants of intrinsically motivated model-free reinforcement learning algorithms. The relation between Go-Explore and these algorithms is discussed in Section~\ref{intro} and many specific algorithms are included in our comparison in Appendix~\ref{sec:scores}, as they account for most of the high-scoring work on Montezuma's Revenge prior to Go-Explore.

\section{Conclusion}

Go-Explore represents an exciting new family of algorithms for solving hard-exploration reinforcement learning problems, meaning those with sparse and/or deceptive rewards. It opens up a large number of new research directions beyond the simple version described in this paper, including experimenting with different archives, different methods for choosing which cells to return to, different cell representations, different exploration methods, and different robustification methods. We expect Go-Explore will accelerate progress in a variety of challenging domains such as robotics. It will also be interesting to see not only the domains in which it excels, but also those in which it fails. Go-Explore thus opens a new playground of possibilities for future research, and we hope the community will join us in investigating this new terrain.

\subsubsection*{Acknowledgments}
We thank the following for helpful discussions on the Go-Explore algorithm and the ideas behind it: Peter Dayan, Zoubin Ghahramani, Shimon Whiteson, Juergen Schmidhuber, Ian Osband, and Kevin Clune. We also appreciate input from all of the members of Uber AI Labs, especially Vashisht Madhavan, Felipe Petroski Such, John Sears, and Thomas Miconi. We are also deeply appreciative of the machine learning community at large for providing feedback that refined our thinking and exposition of Go-Explore, including all of those that provided commentary on Reddit, Twitter, and via other online mediums such as blog posts about our work. Finally, we are grateful to Leon Rosenshein, Joel Snow, Thaxton Beesley, the Colorado Data Center team and the entire OpusStack Team at Uber for providing our computing platform and for technical support.

\bibliographystyle{unsrtnat}{\small\bibliography{nips,ucf,nnstrings,nn,goexplore}}

\begin{thebibliography}{105}
\providecommand{\natexlab}[1]{#1}
\providecommand{\url}[1]{\texttt{#1}}
\expandafter\ifx\csname urlstyle\endcsname\relax
  \providecommand{\doi}[1]{doi: #1}\else
  \providecommand{\doi}{doi: \begingroup \urlstyle{rm}\Url}\fi

\bibitem[Silver et~al.(2016)Silver, Huang, Maddison, Guez, Sifre, Van
  Den~Driessche, Schrittwieser, Antonoglou, Panneershelvam, Lanctot,
  et~al.]{silver2016mastering}
David Silver, Aja Huang, Chris~J Maddison, Arthur Guez, Laurent Sifre, George
  Van Den~Driessche, Julian Schrittwieser, Ioannis Antonoglou, Veda
  Panneershelvam, Marc Lanctot, et~al.
\newblock Mastering the game of go with deep neural networks and tree search.
\newblock \emph{Nature}, 529\penalty0 (7587):\penalty0 484--489, 2016.

\bibitem[Silver et~al.(2017{\natexlab{a}})Silver, Schrittwieser, Simonyan,
  Antonoglou, Huang, Guez, Hubert, Baker, Lai, Bolton, Chen, Lillicrap, Hui,
  Sifre, van~den Driessche, Graepel, and Hassabis]{Silver2017MasteringTG}
David Silver, Julian Schrittwieser, Karen Simonyan, Ioannis Antonoglou, Aja
  Huang, Arthur Guez, Thomas Hubert, L~Robert Baker, Matthew Lai, Adrian
  Bolton, Yutian Chen, Timothy~P. Lillicrap, Fan Hui, Laurent Sifre, George
  van~den Driessche, Thore Graepel, and Demis Hassabis.
\newblock Mastering the game of go without human knowledge.
\newblock \emph{Nature}, 550:\penalty0 354--359, 2017{\natexlab{a}}.

\bibitem[Mnih et~al.(2015)Mnih, Kavukcuoglu, Silver, Rusu, Veness, Bellemare,
  Graves, Riedmiller, Fidjeland, Ostrovski, Petersen, Beattie, Sadik,
  Antonoglou, King, Kumaran, Wierstra, Legg, and Hassabis]{mnih:nature15}
Volodymyr Mnih, Koray Kavukcuoglu, David Silver, Andrei~A Rusu, Joel Veness,
  Marc~G Bellemare, Alex Graves, Martin Riedmiller, Andreas~K Fidjeland, Georg
  Ostrovski, Stig Petersen, Charles Beattie, Amir Sadik, Ioannis Antonoglou,
  Helen King, Dharshan Kumaran, Daan Wierstra, Shane Legg, and Demis Hassabis.
\newblock Human-level control through deep reinforcement learning.
\newblock \emph{Nature}, 518\penalty0 (7540):\penalty0 529--533, 2015.

\bibitem[Bellemare et~al.(2016)Bellemare, Srinivasan, Ostrovski, Schaul,
  Saxton, and Munos]{bellemare2016unifying}
Marc Bellemare, Sriram Srinivasan, Georg Ostrovski, Tom Schaul, David Saxton,
  and Remi Munos.
\newblock Unifying count-based exploration and intrinsic motivation.
\newblock In \emph{NIPS}, pages 1471--1479, 2016.

\bibitem[Amodei et~al.(2016)Amodei, Olah, Steinhardt, Christiano, Schulman, and
  Man{\'e}]{Amodei2016ConcretePI}
Dario Amodei, Chris Olah, Jacob Steinhardt, Paul~F. Christiano, John Schulman,
  and Dan Man{\'e}.
\newblock Concrete problems in ai safety.
\newblock \emph{CoRR}, abs/1606.06565, 2016.

\bibitem[Lehman et~al.(2018)Lehman, Clune, Misevic, Adami, Beaulieu, Bentley,
  Bernard, Beslon, Bryson, Chrabaszcz, Cheney, Cully, Doncieux, Dyer, Ellefsen,
  Feldt, Fischer, Forrest, Fr{\'e}noy, Gagn{\'e}, Goff, Grabowski, Hodjat,
  Hutter, Keller, Knibbe, Krcah, Lenski, Lipson, MacCurdy, Maestre,
  Miikkulainen, Mitri, Moriarty, Mouret, Nguyen, Ofria, Parizeau, Parsons,
  Pennock, Punch, Ray, Schoenauer, Shulte, Sims, Stanley, Taddei, Tarapore,
  Thibault, Weimer, Watson, and Yosinksi]{Lehman2018TheSC}
Joel Lehman, Jeff Clune, Dusan Misevic, Christoph Adami, Julie Beaulieu,
  Peter~J. Bentley, Samuel Bernard, Guillaume Beslon, David~M. Bryson, Patryk
  Chrabaszcz, Nick Cheney, Antoine Cully, St{\'e}phane Doncieux, Fred~C. Dyer,
  Kai~Olav Ellefsen, Robert Feldt, Stephan Fischer, Stephanie Forrest, Antoine
  Fr{\'e}noy, Christian Gagn{\'e}, Leni K.~Le Goff, Laura~M. Grabowski, Babak
  Hodjat, Frank Hutter, Laurent Keller, Carole Knibbe, Peter Krcah, Richard~E.
  Lenski, Hod Lipson, Robert~B MacCurdy, Carlos Maestre, Risto Miikkulainen,
  Sara Mitri, David~E. Moriarty, Jean-Baptiste Mouret, Anh Tuan~Le Nguyen,
  Charles Ofria, Marc Parizeau, David~P. Parsons, Robert~T. Pennock, William~F.
  Punch, Thomas~S. Ray, Marc Schoenauer, Eric Shulte, Karl Sims, Kenneth~O.
  Stanley, François Taddei, Danesh Tarapore, Simon Thibault, Westley Weimer,
  Richard Watson, and Jason Yosinksi.
\newblock The surprising creativity of digital evolution: A collection of
  anecdotes from the evolutionary computation and artificial life research
  communities.
\newblock \emph{CoRR}, abs/1803.03453, 2018.

\bibitem[Lehman and Stanley(2011{\natexlab{a}})]{lehman}
Joel Lehman and Kenneth~O. Stanley.
\newblock Novelty search and the problem with objectives.
\newblock In \emph{Genetic Programming Theory and Practice IX (GPTP 2011)},
  2011{\natexlab{a}}.

\bibitem[Chrabaszcz et~al.(2018)Chrabaszcz, Loshchilov, and
  Hutter]{Chrabaszcz2018BackTB}
Patryk Chrabaszcz, Ilya Loshchilov, and Frank Hutter.
\newblock Back to basics: Benchmarking canonical evolution strategies for
  playing atari.
\newblock In \emph{IJCAI}, 2018.

\bibitem[Schmidhuber(1991{\natexlab{a}})]{schmidhuber1991possibility}
J{\"u}rgen Schmidhuber.
\newblock A possibility for implementing curiosity and boredom in
  model-building neural controllers.
\newblock In \emph{Proc. of the international conference on simulation of
  adaptive behavior: From animals to animats}, pages 222--227,
  1991{\natexlab{a}}.

\bibitem[Oudeyer and Kaplan(2009)]{oudeyer2009intrinsic}
Pierre-Yves Oudeyer and Frederic Kaplan.
\newblock What is intrinsic motivation? a typology of computational approaches.
\newblock \emph{Frontiers in Neurorobotics}, 1:\penalty0 6, 2009.

\bibitem[Barto(2013)]{barto2013intrinsic}
Andrew~G Barto.
\newblock Intrinsic motivation and reinforcement learning.
\newblock In \emph{Intrinsically motivated learning in natural and artificial
  systems}, pages 17--47. Springer, 2013.

\bibitem[Schmidhuber(2006)]{Schmidhuber2006DevelopmentalRO}
J{\"u}rgen Schmidhuber.
\newblock Developmental robotics, optimal artificial curiosity, creativity,
  music, and the fine arts.
\newblock \emph{Connect. Sci.}, 18:\penalty0 173--187, 2006.

\bibitem[Schmidhuber(1991{\natexlab{b}})]{schmidhuber1991curious}
J{\"u}rgen Schmidhuber.
\newblock Curious model-building control systems.
\newblock In \emph{Neural Networks, 1991. 1991 IEEE International Joint
  Conference on}, pages 1458--1463. IEEE, 1991{\natexlab{b}}.

\bibitem[Conti et~al.(2017)Conti, Madhavan, {Petroski Such}, Lehman, Stanley,
  and Clune]{conti:arxiv17}
Edoardo Conti, Vashisht Madhavan, Felipe {Petroski Such}, Joel Lehman,
  Kenneth~O. Stanley, and Jeff Clune.
\newblock Improving exploration in evolution strategies for deep reinforcement
  learning via a population of novelty-seeking agents.
\newblock \emph{arXiv preprint arXiv:1712.06560}, 2017.

\bibitem[Achiam and Sastry(2017)]{Achiam2017SurpriseBasedIM}
Joshua Achiam and S.~Shankar Sastry.
\newblock Surprise-based intrinsic motivation for deep reinforcement learning.
\newblock \emph{CoRR}, abs/1703.01732, 2017.

\bibitem[Burda et~al.(2018)Burda, Edwards, Storkey, and Klimov]{burda:rnd2018}
Yuri Burda, Harrison Edwards, Amos Storkey, and Oleg Klimov.
\newblock Exploration by random network distillation.
\newblock \emph{arXiv preprint arXiv:1810.12894}, 2018.

\bibitem[Velez and Clune(2017)]{velez2017diffusion}
Roby Velez and Jeff Clune.
\newblock Diffusion-based neuromodulation can eliminate catastrophic forgetting
  in simple neural networks.
\newblock \emph{PloS one}, 12\penalty0 (11):\penalty0 e0187736, 2017.

\bibitem[Ellefsen et~al.(2015)Ellefsen, Mouret, Clune, and
  Bongard]{ellefsen2015neural}
Kai~Olav Ellefsen, Jean-Baptiste Mouret, Jeff Clune, and Josh~C Bongard.
\newblock Neural modularity helps organisms evolve to learn new skills without
  forgetting old skills.
\newblock \emph{PLoS Comput Biol}, 11\penalty0 (4):\penalty0 e1004128, 2015.

\bibitem[Kirkpatrick et~al.(2017)Kirkpatrick, Pascanu, Rabinowitz, Veness,
  Desjardins, Rusu, Milan, Quan, Ramalho, Grabska-Barwinska,
  et~al.]{kirkpatrick2017overcoming}
James Kirkpatrick, Razvan Pascanu, Neil Rabinowitz, Joel Veness, Guillaume
  Desjardins, Andrei~A Rusu, Kieran Milan, John Quan, Tiago Ramalho, Agnieszka
  Grabska-Barwinska, et~al.
\newblock Overcoming catastrophic forgetting in neural networks.
\newblock \emph{Proceedings of the national academy of sciences}, page
  201611835, 2017.

\bibitem[French(1999)]{french1999catastrophic}
Robert~M French.
\newblock Catastrophic forgetting in connectionist networks.
\newblock \emph{Trends in cognitive sciences}, 3\penalty0 (4):\penalty0
  128--135, 1999.

\bibitem[Zhang and Sutton(2017)]{Zhang2017ADL}
Shangtong Zhang and Richard~S. Sutton.
\newblock A deeper look at experience replay.
\newblock \emph{CoRR}, abs/1712.01275, 2017.

\bibitem[Liu and Zou(2017)]{Liu2017TheEO}
Ruishan Liu and James Zou.
\newblock The effects of memory replay in reinforcement learning.
\newblock \emph{CoRR}, abs/1710.06574, 2017.

\bibitem[Sutton and Barto(1998)]{sutton1998reinforcement}
Richard~S Sutton and Andrew~G Barto.
\newblock \emph{Reinforcement learning: An introduction}, volume~1.
\newblock Bradford, 1998.

\bibitem[Plappert et~al.(2017)Plappert, Houthooft, Dhariwal, Sidor, Chen, Chen,
  Asfour, Abbeel, and Andrychowicz]{plappert2017parameter}
Matthias Plappert, Rein Houthooft, Prafulla Dhariwal, Szymon Sidor, Richard~Y
  Chen, Xi~Chen, Tamim Asfour, Pieter Abbeel, and Marcin Andrychowicz.
\newblock Parameter space noise for exploration.
\newblock \emph{arXiv preprint arXiv:1706.01905}, 2017.

\bibitem[R{\"u}ckstie\ss et~al.(2008)R{\"u}ckstie\ss, Felder, and
  Schmidhuber]{Rckstie2008StateDependentEF}
Thomas R{\"u}ckstie\ss, Martin Felder, and J{\"u}rgen Schmidhuber.
\newblock State-dependent exploration for policy gradient methods.
\newblock In \emph{ECML/PKDD}, 2008.

\bibitem[Hester et~al.(2018)Hester, Vecerik, Pietquin, Lanctot, Schaul, Piot,
  Horgan, Quan, Sendonaris, Osband, Dulac-Arnold, Agapiou, Leibo, and
  Gruslys]{hester2017deep}
Todd Hester, Matej Vecerik, Olivier Pietquin, Marc Lanctot, Tom Schaul, Bilal
  Piot, Dan Horgan, John Quan, Andrew Sendonaris, Ian Osband, Gabriel
  Dulac-Arnold, John Agapiou, Joel~Z. Leibo, and Audrunas Gruslys.
\newblock Deep q-learning from demonstrations.
\newblock In \emph{AAAI}, 2018.

\bibitem[Pohlen et~al.(2018)Pohlen, Piot, Hester, Azar, Horgan, Budden,
  Barth-Maron, van Hasselt, Quan, Ve{\v{c}}er{\'\i}k,
  et~al.]{pohlen2018observe}
Tobias Pohlen, Bilal Piot, Todd Hester, Mohammad~Gheshlaghi Azar, Dan Horgan,
  David Budden, Gabriel Barth-Maron, Hado van Hasselt, John Quan, Mel
  Ve{\v{c}}er{\'\i}k, et~al.
\newblock Observe and look further: Achieving consistent performance on atari.
\newblock \emph{arXiv preprint arXiv:1805.11593}, 2018.

\bibitem[Salimans and Chen(2018)]{salimans2018learning}
Tim Salimans and Richard Chen.
\newblock Learning montezuma's revenge from a single demonstration.
\newblock \emph{arXiv preprint arXiv:1812.03381}, 2018.

\bibitem[Ho and Ermon(2016)]{Ho2016GenerativeAI}
Jonathan Ho and Stefano Ermon.
\newblock Generative adversarial imitation learning.
\newblock In \emph{NIPS}, 2016.

\bibitem[Bellemare et~al.(2013)Bellemare, Naddaf, Veness, and
  Bowling]{bellemare2013arcade}
Marc~G Bellemare, Yavar Naddaf, Joel Veness, and Michael Bowling.
\newblock The arcade learning environment: An evaluation platform for general
  agents.
\newblock \emph{J. Artif. Intell. Res.(JAIR)}, 47:\penalty0 253--279, 2013.

\bibitem[Machado et~al.(2018)Machado, Bellemare, Talvitie, Veness, Hausknecht,
  and Bowling]{Machado2018RevisitingTA}
Marlos~C. Machado, Marc~G. Bellemare, Erik Talvitie, Joel Veness, Matthew~J.
  Hausknecht, and Michael~H. Bowling.
\newblock Revisiting the arcade learning environment: Evaluation protocols and
  open problems for general agents.
\newblock \emph{J. Artif. Intell. Res.}, 61:\penalty0 523--562, 2018.

\bibitem[Garriga~Alonso(2017)]{garriga2017solving}
Adri{\`a} Garriga~Alonso.
\newblock Solving montezuma's revenge with planning and reinforcement learning,
  2017.

\bibitem[Tang et~al.(2017)Tang, Houthooft, Foote, Stooke, Chen, Duan, Schulman,
  DeTurck, and Abbeel]{tang2017exploration}
Haoran Tang, Rein Houthooft, Davis Foote, Adam Stooke, OpenAI~Xi Chen, Yan
  Duan, John Schulman, Filip DeTurck, and Pieter Abbeel.
\newblock \# exploration: A study of count-based exploration for deep
  reinforcement learning.
\newblock In \emph{NIPS}, pages 2750--2759, 2017.

\bibitem[Gruslys et~al.(2017)Gruslys, Azar, Bellemare, and
  Munos]{gruslys2017reactor}
Audrunas Gruslys, Mohammad~Gheshlaghi Azar, Marc~G Bellemare, and Remi Munos.
\newblock The reactor: A sample-efficient actor-critic architecture.
\newblock \emph{arXiv preprint arXiv:1704.04651}, 2017.

\bibitem[Martin et~al.(2017)Martin, Sasikumar, Everitt, and
  Hutter]{sasikumar2017exploration}
Jarryd Martin, Suraj~Narayanan Sasikumar, Tom Everitt, and Marcus Hutter.
\newblock Count-based exploration in feature space for reinforcement learning.
\newblock In \emph{IJCAI}, 2017.

\bibitem[Ostrovski et~al.(2017)Ostrovski, Bellemare, van~den Oord, and
  Munos]{ostrovski2017count}
Georg Ostrovski, Marc~G. Bellemare, A{\"a}ron van~den Oord, and R{\'e}mi Munos.
\newblock Count-based exploration with neural density models.
\newblock In \emph{ICML}, 2017.

\bibitem[Stanton and Clune(2018)]{Stanton2018DeepCS}
Christopher Stanton and Jeff Clune.
\newblock Deep curiosity search: Intra-life exploration improves performance on
  challenging deep reinforcement learning problems.
\newblock \emph{CoRR}, abs/1806.00553, 2018.

\bibitem[O'Donoghue et~al.(2018)O'Donoghue, Osband, Munos, and
  Mnih]{ODonoghue2018TheUB}
Brendan O'Donoghue, Ian Osband, R{\'e}mi Munos, and Volodymyr Mnih.
\newblock The uncertainty bellman equation and exploration.
\newblock In \emph{ICML}, 2018.

\bibitem[Choi et~al.(2018)Choi, Guo, Moczulski, Oh, Wu, Norouzi, and
  Lee]{Choi2018ContingencyAwareEI}
Jongwook Choi, Yijie Guo, Marcin Moczulski, Junhyuk Oh, Neal Wu, Mohammad
  Norouzi, and Honglak Lee.
\newblock Contingency-aware exploration in reinforcement learning.
\newblock \emph{CoRR}, abs/1811.01483, 2018.

\bibitem[Mnih et~al.(2016{\natexlab{a}})Mnih, Badia, Mirza, Graves, Lillicrap,
  Harley, Silver, and Kavukcuoglu]{a3c}
Volodymyr Mnih, Adria~Puigdomenech Badia, Mehdi Mirza, Alex Graves, Timothy
  Lillicrap, Tim Harley, David Silver, and Koray Kavukcuoglu.
\newblock Asynchronous methods for deep reinforcement learning.
\newblock In \emph{ICML}, pages 1928--1937, 2016{\natexlab{a}}.

\bibitem[Horgan et~al.(2018)Horgan, Quan, Budden, Barth-Maron, Hessel, van
  Hasselt, and Silver]{horgan:apexdqn2018}
Dan Horgan, John Quan, David Budden, Gabriel Barth-Maron, Matteo Hessel, Hado
  van Hasselt, and David Silver.
\newblock Distributed prioritized experience replay.
\newblock \emph{CoRR}, abs/1803.00933, 2018.

\bibitem[Espeholt et~al.(2018)Espeholt, Soyer, Munos, Simonyan, Mnih, Ward,
  Doron, Firoiu, Harley, Dunning, Legg, and Kavukcuoglu]{espeholt:impala2018}
Lasse Espeholt, Hubert Soyer, R{\'e}mi Munos, Karen Simonyan, Volodymyr Mnih,
  Tom Ward, Yotam Doron, Vlad Firoiu, Tim Harley, Iain Dunning, Shane Legg, and
  Koray Kavukcuoglu.
\newblock Impala: Scalable distributed deep-rl with importance weighted
  actor-learner architectures.
\newblock In \emph{ICML}, 2018.

\bibitem[Liu et~al.(2019)Liu, Keramati, Seshadri, Guu, Pasupat, Brunskill, and
  Liang]{liu2019learning}
Evan~Zheran Liu, Ramtin Keramati, Sudarshan Seshadri, Kelvin Guu, Panupong
  Pasupat, Emma Brunskill, and Percy Liang.
\newblock Learning abstract models for long-horizon exploration, 2019.
\newblock URL \url{https://openreview.net/forum?id=ryxLG2RcYX}.

\bibitem[Keramati et~al.(2019)Keramati, Whang, Cho, and
  Brunskill]{Keramati2018FastEW}
Ramtin Keramati, Jay Whang, Patrick Cho, and Emma Brunskill.
\newblock Fast exploration with simplified models and approximately optimistic
  planning in model based reinforcement learning, 2019.
\newblock URL \url{https://openreview.net/forum?id=HygS7n0cFQ}.

\bibitem[Aytar et~al.(2018)Aytar, Pfaff, Budden, Paine, Wang, and
  de~Freitas]{aytar2018playing}
Yusuf Aytar, Tobias Pfaff, David Budden, Tom~Le Paine, Ziyu Wang, and Nando
  de~Freitas.
\newblock Playing hard exploration games by watching youtube.
\newblock \emph{arXiv preprint arXiv:1805.11592}, 2018.

\bibitem[Such et~al.(2018)Such, Madhavan, Liu, Wang, Castro, Li, Schubert,
  Bellemare, Clune, and Lehman]{such2018atari}
Felipe~Petroski Such, Vashisht Madhavan, Rosanne Liu, Rui Wang, Pablo~Samuel
  Castro, Yulun Li, Ludwig Schubert, Marc Bellemare, Jeff Clune, and Joel
  Lehman.
\newblock An atari model zoo for analyzing, visualizing, and comparing deep
  reinforcement learning agents.
\newblock \emph{arXiv preprint arXiv:1812.07069}, 2018.

\bibitem[Kocsis and Szepesv{\'a}ri(2006)]{Kocsis2006BanditBM}
Levente Kocsis and Csaba Szepesv{\'a}ri.
\newblock Bandit based monte-carlo planning.
\newblock In \emph{ECML}, 2006.

\bibitem[Kocsis et~al.(2006)Kocsis, Szepesv{\'a}ri, and
  Willemson]{Kocsis2006ImprovedMS}
Levente Kocsis, Csaba Szepesv{\'a}ri, and Jan Willemson.
\newblock Improved monte-carlo search.
\newblock \emph{Univ. Tartu, Estonia, Tech. Rep}, 1, 2006.

\bibitem[Browne et~al.(2012)Browne, Powley, Whitehouse, Lucas, Cowling,
  Rohlfshagen, Tavener, Liebana, Samothrakis, and Colton]{Browne2012ASO}
Cameron Browne, Edward~Jack Powley, Daniel Whitehouse, Simon~M. Lucas, Peter~I.
  Cowling, Philipp Rohlfshagen, Stephen Tavener, Diego~Perez Liebana, Spyridon
  Samothrakis, and Simon Colton.
\newblock A survey of monte carlo tree search methods.
\newblock \emph{IEEE Transactions on Computational Intelligence and AI in
  Games}, 4:\penalty0 1--43, 2012.

\bibitem[Chaslot et~al.(2008)Chaslot, Bakkes, Szita, and
  Spronck]{Chaslot2008MonteCarloTS}
Guillaume Chaslot, Sander Bakkes, Istv{\'a}n Szita, and Pieter Spronck.
\newblock Monte-carlo tree search: A new framework for game ai.
\newblock In \emph{AIIDE}, 2008.

\bibitem[Lipovetzky et~al.(2015)Lipovetzky, Ram{\'i}rez, and
  Geffner]{Lipovetzky2015ClassicalPW}
Nir Lipovetzky, Miquel Ram{\'i}rez, and Hector Geffner.
\newblock Classical planning with simulators: Results on the atari video games.
\newblock In \emph{IJCAI}, 2015.

\bibitem[Silver et~al.(2017{\natexlab{b}})Silver, van Hasselt, Hessel, Schaul,
  Guez, Harley, Dulac-Arnold, Reichert, Rabinowitz, Barreto, and
  Degris]{Silver2017ThePE}
David Silver, Hado van Hasselt, Matteo Hessel, Tom Schaul, Arthur Guez, Tim
  Harley, Gabriel Dulac-Arnold, David~P. Reichert, Neil~C. Rabinowitz,
  Andr{\'e} Barreto, and Thomas Degris.
\newblock The predictron: End-to-end learning and planning.
\newblock In \emph{ICML}, 2017{\natexlab{b}}.

\bibitem[Lange and Riedmiller(2010)]{lange2010deep}
Sascha Lange and Martin Riedmiller.
\newblock Deep auto-encoder neural networks in reinforcement learning.
\newblock In \emph{The International Joint Conference on Neural Networks},
  pages 1--8. IEEE, 2010.

\bibitem[Jaderberg et~al.(2016)Jaderberg, Mnih, Czarnecki, Schaul, Leibo,
  Silver, and Kavukcuoglu]{Jaderberg2016ReinforcementLW}
Max Jaderberg, Volodymyr Mnih, Wojciech Czarnecki, Tom Schaul, Joel~Z. Leibo,
  David Silver, and Koray Kavukcuoglu.
\newblock Reinforcement learning with unsupervised auxiliary tasks.
\newblock \emph{CoRR}, abs/1611.05397, 2016.

\bibitem[Ecoffet et~al.(2018)Ecoffet, Huizinga, Lehman, Stanley, and
  Clune]{ecoffet:bloggoexplore2018}
Adrien Ecoffet, Joost Huizinga, Joel Lehman, Kenneth~O Stanley, and Jeff Clune.
\newblock Montezuma’s revenge solved by go-explore, a new algorithm for
  hard-exploration problems (sets records on pitfall, too).
\newblock \emph{Uber Engineering Blog}, Nov 2018.
\newblock URL \url{http://eng.uber.com/go-explore}.

\bibitem[McAllister et~al.(2018)McAllister, Kahn, Clune, and
  Levine]{mcallister2018robustness}
Rowan McAllister, Gregory Kahn, Jeff Clune, and Sergey Levine.
\newblock Robustness to out-of-distribution inputs via task-aware generative
  uncertainty.
\newblock \emph{arXiv preprint arXiv:1812.10687}, 2018.

\bibitem[Kahn et~al.(2017)Kahn, Villaflor, Pong, Abbeel, and
  Levine]{kahn2017uncertainty}
Gregory Kahn, Adam Villaflor, Vitchyr Pong, Pieter Abbeel, and Sergey Levine.
\newblock Uncertainty-aware reinforcement learning for collision avoidance.
\newblock \emph{arXiv preprint arXiv:1702.01182}, 2017.

\bibitem[Lake et~al.(2017)Lake, Ullman, Tenenbaum, and
  Gershman]{Lake2017BuildingMT}
Brenden~M. Lake, Tomer~D. Ullman, Joshua~B. Tenenbaum, and Samuel~J Gershman.
\newblock Building machines that learn and think like people.
\newblock \emph{The Behavioral and brain sciences}, 40:\penalty0 e253, 2017.

\bibitem[Koos et~al.(2013)Koos, Mouret, and Doncieux]{koos2013transferability}
Sylvain Koos, Jean-Baptiste Mouret, and St{\'e}phane Doncieux.
\newblock The transferability approach: Crossing the reality gap in
  evolutionary robotics.
\newblock \emph{IEEE Transactions on Evolutionary Computation}, 17\penalty0
  (1):\penalty0 122--145, 2013.

\bibitem[Cully et~al.(2015)Cully, Clune, Tarapore, and Mouret]{cully:nature15}
A.~Cully, J.~Clune, D.~Tarapore, and J.-B. Mouret.
\newblock Robots that can adapt like animals.
\newblock \emph{Nature}, 521:\penalty0 503--507, 2015.
\newblock \doi{10.1038/nature14422}.

\bibitem[Andrychowicz et~al.(2018)Andrychowicz, Baker, Chociej, Jozefowicz,
  McGrew, Pachocki, Petron, Plappert, Powell, Ray,
  et~al.]{andrychowicz2018learning}
Marcin Andrychowicz, Bowen Baker, Maciek Chociej, Rafal Jozefowicz, Bob McGrew,
  Jakub Pachocki, Arthur Petron, Matthias Plappert, Glenn Powell, Alex Ray,
  et~al.
\newblock Learning dexterous in-hand manipulation.
\newblock \emph{arXiv preprint arXiv:1808.00177}, 2018.

\bibitem[Andrychowicz et~al.(2017)Andrychowicz, Wolski, Ray, Schneider, Fong,
  Welinder, McGrew, Tobin, Abbeel, and Zaremba]{andrychowicz2017hindsight}
Marcin Andrychowicz, Filip Wolski, Alex Ray, Jonas Schneider, Rachel Fong,
  Peter Welinder, Bob McGrew, Josh Tobin, OpenAI~Pieter Abbeel, and Wojciech
  Zaremba.
\newblock Hindsight experience replay.
\newblock In \emph{Advances in Neural Information Processing Systems}, pages
  5048--5058, 2017.

\bibitem[Schaul et~al.(2015{\natexlab{a}})Schaul, Horgan, Gregor, and
  Silver]{schaul2015universal}
Tom Schaul, Daniel Horgan, Karol Gregor, and David Silver.
\newblock Universal value function approximators.
\newblock In \emph{International Conference on Machine Learning}, pages
  1312--1320, 2015{\natexlab{a}}.

\bibitem[Schulman et~al.(2017)Schulman, Wolski, Dhariwal, Radford, and
  Klimov]{Schulman2017ProximalPO}
John Schulman, Filip Wolski, Prafulla Dhariwal, Alec Radford, and Oleg Klimov.
\newblock Proximal policy optimization algorithms.
\newblock \emph{CoRR}, abs/1707.06347, 2017.

\bibitem[Resnick et~al.(2018)Resnick, Raileanu, Kapoor, Peysakhovich, Cho, and
  Bruna]{Resnick2018BackplayM}
Cinjon Resnick, Roberta Raileanu, Sanyam Kapoor, Alex Peysakhovich, Kyunghyun
  Cho, and Joan Bruna.
\newblock Backplay: "man muss immer umkehren".
\newblock \emph{CoRR}, abs/1807.06919, 2018.

\bibitem[{Petroski Such} et~al.(2017){Petroski Such}, Madhavan, Conti, Lehman,
  Stanley, and Clune]{such:arxiv17}
Felipe {Petroski Such}, Vashisht Madhavan, Edoardo Conti, Joel Lehman,
  Kenneth~O. Stanley, and Jeff Clune.
\newblock Deep neuroevolution: Genetic algorithms are a competitive alternative
  for training deep neural networks for reinforcement learning.
\newblock \emph{arXiv preprint arXiv:1712.06567}, 2017.

\bibitem[Van~Hasselt et~al.(2016)Van~Hasselt, Guez, and Silver]{van2016deep}
Hado Van~Hasselt, Arthur Guez, and David Silver.
\newblock Deep reinforcement learning with double q-learning.
\newblock In \emph{AAAI}, volume~2, page~5. Phoenix, AZ, 2016.

\bibitem[Wang et~al.(2016)Wang, de~Freitas, and Lanctot]{wang2015dueling}
Ziyu Wang, Nando de~Freitas, and Marc Lanctot.
\newblock Dueling network architectures for deep reinforcement learning.
\newblock In \emph{ICML}, 2016.

\bibitem[Schaul et~al.(2015{\natexlab{b}})Schaul, Quan, Antonoglou, and
  Silver]{schaul2015prioritized}
Tom Schaul, John Quan, Ioannis Antonoglou, and David Silver.
\newblock Prioritized experience replay.
\newblock \emph{arXiv preprint arXiv:1511.05952}, 2015{\natexlab{b}}.

\bibitem[van Hasselt et~al.(2016)van Hasselt, Guez, Hessel, Mnih, and
  Silver]{van2016learning}
Hado~P van Hasselt, Arthur Guez, Matteo Hessel, Volodymyr Mnih, and David
  Silver.
\newblock Learning values across many orders of magnitude.
\newblock In \emph{Advances in Neural Information Processing Systems}, pages
  4287--4295, 2016.

\bibitem[Salimans et~al.(2017)Salimans, Ho, Chen, Sidor, and
  Sutskever]{salimans2017evolution}
Tim Salimans, Jonathan Ho, Xi~Chen, Szymon Sidor, and Ilya Sutskever.
\newblock Evolution strategies as a scalable alternative to reinforcement
  learning.
\newblock \emph{arXiv preprint arXiv:1703.03864}, 2017.

\bibitem[Bellemare et~al.(2017)Bellemare, Dabney, and Munos]{Bellemare2017ADP}
Marc~G. Bellemare, Will Dabney, and R{\'e}mi Munos.
\newblock A distributional perspective on reinforcement learning.
\newblock In \emph{ICML}, 2017.

\bibitem[Hessel et~al.(2018)Hessel, Modayil, van Hasselt, Schaul, Ostrovski,
  Dabney, Horgan, Piot, Azar, and Silver]{Hessel2018RainbowCI}
Matteo Hessel, Joseph Modayil, Hado van Hasselt, Tom Schaul, Georg Ostrovski,
  Will Dabney, Dan Horgan, Bilal Piot, Mohammad~Gheshlaghi Azar, and David
  Silver.
\newblock Rainbow: Combining improvements in deep reinforcement learning.
\newblock In \emph{AAAI}, 2018.

\bibitem[Nair et~al.(2015)Nair, Srinivasan, Blackwell, Alcicek, Fearon,
  De~Maria, Panneershelvam, Suleyman, Beattie, Petersen,
  et~al.]{nair2015massively}
Arun Nair, Praveen Srinivasan, Sam Blackwell, Cagdas Alcicek, Rory Fearon,
  Alessandro De~Maria, Vedavyas Panneershelvam, Mustafa Suleyman, Charles
  Beattie, Stig Petersen, et~al.
\newblock Massively parallel methods for deep reinforcement learning.
\newblock \emph{arXiv preprint arXiv:1507.04296}, 2015.

\bibitem[Brockman et~al.(2016)Brockman, Cheung, Pettersson, Schneider,
  Schulman, Tang, and Zaremba]{brockman}
Greg Brockman, Vicki Cheung, Ludwig Pettersson, Jonas Schneider, John Schulman,
  Jie Tang, and Wojciech Zaremba.
\newblock Openai gym, 2016.

\bibitem[Zoubir and Iskander(2007)]{Zoubir2007BootstrapMA}
A.~M. Zoubir and D.~Robert Iskander.
\newblock Bootstrap methods and applications.
\newblock \emph{IEEE Signal Processing Magazine}, 24:\penalty0 10--19, 2007.

\bibitem[ata(2018{\natexlab{a}})]{atari_bugs}
Atari vcs/2600 easter egg list, 2018{\natexlab{a}}.
\newblock URL
  \url{http://www.ataricompendium.com/game_library/easter_eggs/vcs/easter_eggs.html}.

\bibitem[ata(2018{\natexlab{b}})]{atari_scoreboard}
Atari vcs/2600 scoreboard, Dec 2018{\natexlab{b}}.
\newblock URL
  \url{http://www.ataricompendium.com/game_library/high_scores/high_scores.html}.

\bibitem[Mouret and Clune(2015)]{mouret:arxiv15}
Jean{-}Baptiste Mouret and Jeff Clune.
\newblock Illuminating search spaces by mapping elites.
\newblock \emph{ArXiv e-prints}, abs/1504.04909, 2015.
\newblock URL \url{http://arxiv.org/abs/1504.04909}.

\bibitem[Lehman and Stanley(2011{\natexlab{b}})]{lehman:gecco11}
Joel Lehman and Kenneth~O. Stanley.
\newblock Evolving a diversity of virtual creatures through novelty search and
  local competition.
\newblock In \emph{GECCO '11: Proceedings of the 13th annual conference on
  Genetic and evolutionary computation}, pages 211--218, 2011{\natexlab{b}}.

\bibitem[Tobin et~al.(2017)Tobin, Fong, Ray, Schneider, Zaremba, and
  Abbeel]{tobin2017domain}
Josh Tobin, Rachel Fong, Alex Ray, Jonas Schneider, Wojciech Zaremba, and
  Pieter Abbeel.
\newblock Domain randomization for transferring deep neural networks from
  simulation to the real world.
\newblock In \emph{Intelligent Robots and Systems (IROS), 2017 IEEE/RSJ
  International Conference on}, pages 23--30. IEEE, 2017.

\bibitem[Lee et~al.(2013)Lee, Yosinski, Glette, Lipson, and
  Clune]{lee2013evolving}
S.~Lee, J.~Yosinski, K.~Glette, H.~Lipson, and J.~Clune.
\newblock Evolving gaits for physical robots with the hyperneat generative
  encoding: the benefits of simulation.
\newblock In \emph{Applications of Evolutionary Computing}. Springer, 2013.

\bibitem[Garc{\i}a and Fern{\'a}ndez(2015)]{garcia2015comprehensive}
Javier Garc{\i}a and Fernando Fern{\'a}ndez.
\newblock A comprehensive survey on safe reinforcement learning.
\newblock \emph{Journal of Machine Learning Research}, 16\penalty0
  (1):\penalty0 1437--1480, 2015.

\bibitem[Lipton et~al.(2016)Lipton, Azizzadenesheli, Kumar, Li, Gao, and
  Deng]{lipton2016combating}
Zachary~C Lipton, Kamyar Azizzadenesheli, Abhishek Kumar, Lihong Li, Jianfeng
  Gao, and Li~Deng.
\newblock Combating reinforcement learning's sisyphean curse with intrinsic
  fear.
\newblock \emph{arXiv preprint arXiv:1611.01211}, 2016.

\bibitem[Saunders et~al.(2018)Saunders, Sastry, Stuhlmueller, and
  Evans]{saunders2018trial}
William Saunders, Girish Sastry, Andreas Stuhlmueller, and Owain Evans.
\newblock Trial without error: Towards safe reinforcement learning via human
  intervention.
\newblock In \emph{Proceedings of the 17th International Conference on
  Autonomous Agents and MultiAgent Systems}, pages 2067--2069. International
  Foundation for Autonomous Agents and Multiagent Systems, 2018.

\bibitem[Colas et~al.(2018)Colas, Sigaud, and Oudeyer]{colas:geppg2018}
C{\'e}dric Colas, Olivier Sigaud, and Pierre-Yves Oudeyer.
\newblock {GEP-PG: Decoupling Exploration and Exploitation in Deep
  Reinforcement Learning Algorithms}.
\newblock In \emph{{International Conference on Machine Learning (ICML)}},
  Stockholm, Sweden, July 2018.
\newblock URL \url{https://hal.inria.fr/hal-01890151}.

\bibitem[Forestier et~al.(2017)Forestier, Mollard, and
  Oudeyer]{forestier2017intrinsically}
S{\'e}bastien Forestier, Yoan Mollard, and Pierre-Yves Oudeyer.
\newblock Intrinsically motivated goal exploration processes with automatic
  curriculum learning.
\newblock \emph{arXiv preprint arXiv:1708.02190}, 2017.

\bibitem[Lillicrap et~al.(2015)Lillicrap, Hunt, Pritzel, Heess, Erez, Tassa,
  Silver, and Wierstra]{Lillicrap2015ContinuousCW}
Timothy~P. Lillicrap, Jonathan~J. Hunt, Alexander Pritzel, Nicolas Heess, Tom
  Erez, Yuval Tassa, David Silver, and Daan Wierstra.
\newblock Continuous control with deep reinforcement learning.
\newblock \emph{CoRR}, abs/1509.02971, 2015.

\bibitem[Levine et~al.(2016)Levine, Finn, Darrell, and Abbeel]{levine2016end}
Sergey Levine, Chelsea Finn, Trevor Darrell, and Pieter Abbeel.
\newblock End-to-end training of deep visuomotor policies.
\newblock \emph{The Journal of Machine Learning Research}, 17\penalty0
  (1):\penalty0 1334--1373, 2016.

\bibitem[Oh et~al.(2018)Oh, Guo, Singh, and Lee]{Oh2018SelfImitationL}
Junhyuk Oh, Yijie Guo, Satinder Singh, and Honglak Lee.
\newblock Self-imitation learning.
\newblock In \emph{ICML}, 2018.

\bibitem[Osband et~al.(2016)Osband, Blundell, Pritzel, and
  Roy]{Osband2016DeepEV}
Ian Osband, Charles Blundell, Alexander Pritzel, and Benjamin~Van Roy.
\newblock Deep exploration via bootstrapped dqn.
\newblock In \emph{NIPS}, 2016.

\bibitem[Goyal et~al.(2018)Goyal, Brakel, Fedus, Lillicrap, Levine, Larochelle,
  and Bengio]{Goyal2018RecallTB}
Anirudh Goyal, Philemon Brakel, William Fedus, Timothy~P. Lillicrap, Sergey
  Levine, Hugo Larochelle, and Yoshua Bengio.
\newblock Recall traces: Backtracking models for efficient reinforcement
  learning.
\newblock \emph{CoRR}, abs/1804.00379, 2018.

\bibitem[Dong et~al.(2019)Dong, Mao, Cui, and Li]{dong2019explicit}
Honghua Dong, Jiayuan Mao, Xinyue Cui, and Lihong Li.
\newblock Explicit recall for efficient exploration, 2019.
\newblock URL \url{https://openreview.net/forum?id=B1GIB3A9YX}.

\bibitem[Kearns and Singh(2002)]{kearns2002near}
Michael Kearns and Satinder Singh.
\newblock Near-optimal reinforcement learning in polynomial time.
\newblock \emph{Machine learning}, 49\penalty0 (2-3):\penalty0 209--232, 2002.

\bibitem[Guo et~al.(2014)Guo, Singh, Lee, Lewis, and Wang]{Guo2014DeepLF}
Xiaoxiao Guo, Satinder~P. Singh, Honglak Lee, Richard~L. Lewis, and Xiaoshi
  Wang.
\newblock Deep learning for real-time atari game play using offline monte-carlo
  tree search planning.
\newblock In \emph{NIPS}, 2014.

\bibitem[Cerezo and Ballester(2018)]{Cerezo2018FractalAA}
Sergio~Hernandez Cerezo and Guillem~Duran Ballester.
\newblock Fractal ai: A fragile theory of intelligence.
\newblock \emph{CoRR}, abs/1803.05049, 2018.

\bibitem[Dijkstra(1959)]{Dijkstra:1959:NTP:2722880.2722945}
E.~W. Dijkstra.
\newblock A note on two problems in connexion with graphs.
\newblock \emph{Numer. Math.}, 1\penalty0 (1):\penalty0 269--271, December
  1959.
\newblock ISSN 0029-599X.
\newblock \doi{10.1007/BF01386390}.
\newblock URL \url{http://dx.doi.org/10.1007/BF01386390}.

\bibitem[Hart et~al.(1968)Hart, Nilsson, and Raphael]{Hart:astar}
P.~E. Hart, N.~J. Nilsson, and B.~Raphael.
\newblock A formal basis for the heuristic determination of minimum cost paths.
\newblock \emph{IEEE Transactions on Systems Science and Cybernetics},
  4\penalty0 (2):\penalty0 100--107, July 1968.
\newblock ISSN 0536-1567.
\newblock \doi{10.1109/TSSC.1968.300136}.

\bibitem[Lavalle(1998)]{LaValle1998RapidlyExploringRT}
Steven~M. Lavalle.
\newblock Rapidly-exploring random trees: A new tool for path planning.
\newblock Technical report, Iowa State University, 1998.

\bibitem[Zhan et~al.(2018)Zhan, Aytemiz, and Smith]{zhan2018taking}
Zeping Zhan, Batu Aytemiz, and Adam~M Smith.
\newblock Taking the scenic route: Automatic exploration for videogames.
\newblock \emph{arXiv preprint arXiv:1812.03125}, 2018.

\bibitem[Goldberg(1989)]{goldberg:gabook89}
David~E. Goldberg.
\newblock \emph{Genetic Algorithms in Search, Optimization and Machine
  Learning}.
\newblock Addison-Wesley, Reading, MA, 1989.

\bibitem[Lipowski and Lipowska(2011)]{Lipowski2011RoulettewheelSV}
Adam Lipowski and Dorota Lipowska.
\newblock Roulette-wheel selection via stochastic acceptance.
\newblock \emph{CoRR}, abs/1109.3627, 2011.

\bibitem[Bellemare et~al.(2012)Bellemare, Veness, and
  Bowling]{Bellemare2012InvestigatingCA}
Marc~G. Bellemare, Joel Veness, and Michael~H. Bowling.
\newblock Investigating contingency awareness using atari 2600 games.
\newblock In \emph{AAAI}, 2012.

\bibitem[Stadie et~al.(2015)Stadie, Levine, and
  Abbeel]{stadie2015incentivizing}
Bradly~C Stadie, Sergey Levine, and Pieter Abbeel.
\newblock Incentivizing exploration in reinforcement learning with deep
  predictive models.
\newblock \emph{arXiv preprint arXiv:1507.00814}, 2015.

\bibitem[Mnih et~al.(2016{\natexlab{b}})Mnih, Badia, Mirza, Graves, Lillicrap,
  Harley, Silver, and Kavukcuoglu]{mnih2016asynchronous}
Volodymyr Mnih, Adria~Puigdomenech Badia, Mehdi Mirza, Alex Graves, Timothy
  Lillicrap, Tim Harley, David Silver, and Koray Kavukcuoglu.
\newblock Asynchronous methods for deep reinforcement learning.
\newblock In \emph{International conference on machine learning}, pages
  1928--1937, 2016{\natexlab{b}}.

\end{thebibliography}

\appendix

\section{Appendix}

\subsection{The meaning of ``frames''}
\label{sec:frame_definition}

It is common practice to introduce ``frame skipping'' during training in the Atari domain, so that the agent only selects actions every $k$ frames instead of every single frame (the action then persists across $k$ frames). The most common value of $k$ is 4, and both our exploration and robustification phases were implemented with frame skipping with $k = 4$.

Following the recommendations in \citet{such:arxiv17}, we call the total number of frames produced by the underlying emulator ``game frames'' and the number of frames seen and acted on by the agent during training ``training frames.'' It can sometimes be difficult to know whether a reported number of frames corresponds to training frames or game frames, and the difference can be significant because the number of game frames is usually 4 times the number of training frames. In this work, frame counts are always reported as \emph{game} frames, as recommended by \citet{such:arxiv17} and \citet{Machado2018RevisitingTA}. Further, we always qualify the word ``frame'' with either ``training'' or ``game.'' This clarification is particularly important for the rare cases in which we are indeed referring to \emph{training} frames and not game frames, such as in Section~\ref{sec:explore_from_cell}, where we mention that in the exploration phase, actions are repeated with 95\% probability each \emph{training} frame.

\subsection{Episode end}
\label{sec:episode_end}

In the case of Montezuma's Revenge, the end of an episode is defined as a loss of life, while in the case of Pitfall it is the game-over signal. Both definitions of the end of an episode appear in the literature~\cite{Machado2018RevisitingTA}, and our use of differing approaches in Montezuma's Revenge and Pitfall was due to the greater difficulty of tracking room location based on pixels in Montezuma's Revenge if the character is allowed to lose lives (a difficulty which does not exist in Pitfall). Additionally, death in Pitfall grants the agent additional affordances, which is not the case in Montezuma's Revenge.
These factors are further explained in Appendix~\ref{sec:domain_from_pixels} below.

\subsection{Extraction of domain knowledge features from pixels}
\label{sec:domain_from_pixels}

Phase 1 of Go-Explore used the following domain knowledge features: the $x,y$ position of the agent, the current room, the current level and the rooms in which the currently held keys were found (these last two only apply to Montezuma's Revenge). Although these features can be found in RAM, they were extracted from pixels in our implementation for two reasons: (1)~extracting information from pixels is more similar to how a real world environment would be tackled and ensures we do not exploit any non-visible information that might be stored in the RAM, and (2)~we found that extracting values from the RAM could be unreliable at times: in Montezuma's Revenge, when the character moves into a new room, a black transition image is shown for a few frames. The current room and current $x,y$ position are updated at different times during these transition frames, so that reading these values from RAM would give a room number and $x,y$ position that are inconsistent.

The location of the agent could be extracted by training a simple classifier, or in an unsupervised way through contingency-awareness~\cite{Choi2018ContingencyAwareEI}, but it turns out that, in both Montezuma's Revenge and Pitfall, some pixel values only occur in the character sprite, making it trivial to identify the character location by searching for these values in the current frame. Coincidentally, searching for pixels with a red channel value of 228 is enough to find the character in both Montezuma's Revenge and Pitfall.

Room changes are identified by detecting sudden changes in $x,y$ position: if the character was located at the far right of the screen and is now located at the far left, it likely moved to a room on the right of the current room. In the case of Pitfall, additional domain knowledge is required: underground transitions move the players 3 rooms at a time instead of just 1, and the map wraps around so that the last room is situated to the left of the first room. In Montezuma's Revenge, knowledge of the map is not strictly necessary for room tracking, as the room transition rules are simple, but it is necessary for level tracking: any transition away from the treasure room is an increase in level.

Loss of life needs to be taken into account when tracking room changes: in Montezuma's Revenge, losing a life causes the character to be brought back to the exact location where it entered the room, so that if the character entered the room from the left and dies on the right of the room, the sudden change of $x$ value due to the character reviving on the left side of the room could be mistaken for a room change. Handling this behavior is possible, but we believe unnecessarily complicated for our purposes. For this reason, we end episodes on loss of life in Montezuma's Revenge. By contrast, in Pitfall, the character is brought to a fixed location on the left side of the screen that cannot be confused with a room change, so that there is no need to end the episode on life loss to simplify room tracking. Further, while losing a life is a strict waste of time in Montezuma's Revenge since it brings the agent back to a previously seen location, in Pitfall it can be used as a form of teleportation: if an agent enters a room from the right and loses a life soon after, it will be teleported all the way to the left of the room, thus skipping the hazards that may be in the middle. For this reason, we did not choose to end episodes on life loss in Pitfall.

Finally, key tracking in Montezuma's Revenge is done simply by pattern-matching for keys in the section of the screen that shows the current inventory, and tracking the room number associated to any increase in the current number of keys.

\subsection{Filtering out bug trajectories}
\label{sec:filter_out_bug}

As mentioned in Section~\ref{sec:mr_no_domain_knowledge}, we filtered out trajectories that triggered the treasure room bug when robustifying Montezuma's Revenge without domain knowledge. Such filtering was not necessary when using domain knowledge because none of the highest scoring trajectories triggered the bug, as explained in Section~\ref{sec:mr_domain_knowledge}. 

The filtering of bug trajectories was done by excluding all trajectories whose level was lower than the maximum level in the archive. That works because the bug makes it impossible to leave the treasure room and advance to the next level, so any trajectory that makes it to a new level did not trigger the bug in the previous level.

\subsection{Cell selection details}
\label{sec:selection_details}

As mentioned in Section~\ref{sec:selecting_cells}, cells are selected at each iteration by first assigning them a score, which is then normalized across all cells in the archive, yielding the probability of each cell being selected. The score of a cell is the sum of separate \emph{subscores}, which we now describe.

One important set of such subscores, called the \emph{count subscores}, are computed from attributes that represent the number of times a cell was interacted with in different ways. Specifically: the number of times a cell has already been chosen (i.e. selected as a cell to explore from), the number of times a cell was visited at any point during the exploration phase, and the number of times a cell has been chosen since exploration from it last produced the discovery of a new or better cell. In the case of each of these attributes, a lower count likely indicates a more promising cell to explore from (e.g.\ a cell that has been chosen more times already is less likely to lead to new cells than a cell that has been chosen fewer times).
The count subscore for each of these attributes is given by:

\begin{equation}
    CntScore(c, a) = w_a \cdot \left(\frac{1}{v(c, a) + \varepsilon_1}\right)^{p_a} + \varepsilon_2
\end{equation}

Here $c$ is the cell for which we are calculating the score, $v(c, a)$ is a function that returns the value of attribute $a$ for cell $c$, $w_a$ is the weight hyperparameter for attribute $a$, and $p_a$ is the power hyperparameter for attribute $a$. $\varepsilon_1$ helps prevent division by 0 and determines the relative weight of cells for which a given value is 0. $\varepsilon_2$ helps guarantee that no cell ever has a 0 probability of being chosen. In our implementation, $\varepsilon_1 = 0.001$ and $\varepsilon_2 = 0.00001$, which we chose after preliminary experiments showed that they worked well.

When cell representations are informed by domain knowledge (Section~\ref{sec:mr_domain_knowledge}), giving us the $x,y$ position of the agent, it is possible to determine the possible neighbors of given cells, and whether these neighbors are already present in the archive.
For those cases, we define a set of \emph{neighbor subscores}. Each neighbor subscore is defined as $w_n$ if neighbor $n$ does \emph{not} exist in the archive, and is $0$ otherwise.
The motivation behind these neighbor subscores is that cells that are lacking neighbors are likely at the edge of the current frontier of knowledge and are thus more likely to yield new cells. We consider 3 types of neighbors: vertical (2 neighbors), horizontal (2 neighbors), and (in the case of Montezuma's Revenge) cells that are in the same level, room and $x,y$ position, but are holding a larger number of keys (the intuition is that if a cell lacks a ``more keys'' neighbor, then it is the cell that is most capable of opening doors from its location). Neighbors of the same type share the same value for $w_n$ (Table~\ref{tab:hyper_domain}). These definitions result in the following neighbor subscore, assuming a function called $HasNeighbor(c, n)$ which returns $1$ if neighbor $n$ of cell $c$ is present in the archive, and which returns $0$ otherwise:

\begin{equation}
    NeighScore(c, n) = w_n \cdot (1 - HasNeighbor(c, n))
\end{equation}

In cases without domain knowledge, it is unclear what exactly would constitute a cell's neighbor, and so $NeighScore$ is defined as 0 in this case in our experiments.

Finally, in the case of Montezuma's Revenge with domain knowledge, cells are exponentially downweighted based on the distance to the maximum level currently reached, thereby favoring progress in the furthest level reached, while still keeping open the possibility of improving previous levels' trajectories:
\begin{equation}
    LevelWeight(c) = 0.1^{MaxLevel - Level(c)}
\end{equation}

In the case of Pitfall (where no notion of level exists) and Montezuma's Revenge without domain knowledge (where we do not know what the level of a given cell is), $LevelWeight$ is always 1.

The final cell score is then computed as follows:
\begin{equation}
    CellScore(c) = LevelWeight(c) \cdot \left[\left(\sum_{n} NeighScore(c, n)\right) + \left(\sum_a CntScore(c, a)\right) + 1\right]
\end{equation}

Note that $CellScore(c) > 0$ for all cells $c$. The cell selection probability is given by:
\begin{equation}
    CellProb(c) = \frac{CellScore(c)}{\sum_{c'} CellScore(c')}
\end{equation}

Hyperparameters (the different values of $w_a$, $p_a$ and $w_n$) were found through separate grid searches on each game (Montezuma's Revenge and Pitfall) and for each treatment (with or without domain knowledge). Detailed hyperparameter tables are found in Appendix~\ref{sec:hyperparameter_phase1} below.

\subsection{Phase 1 hyperparameters}
\label{sec:hyperparameter_phase1}

\FloatBarrier

Hyperparameter values were found through grid search. Here, the power hyperparameter $p_a$ (see Section~\ref{sec:selection_details}) found by grid search turned out to be 0.5 for all attributes $a$ in every experiment, so these are excluded from the tables for conciseness.

The ``count-based'' attributes are as follows: ``Times chosen'' is the number of times a cell was selected from the archive so far, ``Times chosen since new'' is the number of times the cell was selected from the archive since last time it led to a new cell being found or to a cell being improved, and ``Times seen'' is the number of times the cell was seen during exploration, regardless of whether it was chosen.

Table~\ref{tab:hyper_no_domain} shows the hyperparameters for the runs with downsampled frames, i.e. the ones without domain knowledge, and Table~\ref{tab:hyper_domain} shows the hyperparameters for the runs with domain knowledge.

\begin{table}[!htbp]
    \begin{center}
        \begin{tabular}{ c | c c } 
             & \makecell{Montezuma's Revenge} & Pitfall \\
            \hline
            Batch size & 100  & 1,000 \\
            Downsampled width & 11 & 11 \\ 
            Downsampled height & 8 & 8 \\
            Downsampled pixel range & 0-8 & 0-8 \\
            Times chosen weight ($w_{a=Chosen}$) & 0.1 & 1 \\
            Times chosen since new weight ($w_{a=ChosenSinceNew}$) & 0 & 1 \\
            Times seen weight ($w_{a=Seen}$) & 0.3 &  0 \\
        \end{tabular}
        \vspace{4mm}
        \caption{\textbf{Hyperparameter values for Montezuma's Revenge and Pitfall \emph{without} domain knowledge.}}
        \label{tab:hyper_no_domain} 
    \end{center}
\end{table}

\begin{table}[!htbp]
    \begin{center}
        \begin{tabular}{  c | c c } 
             & \makecell{Montezuma's Revenge} & Pitfall \\
            \hline
            Batch size & 1,000 & 1,000 \\
            Cell size & $16 \times 16$ & $16 \times 16$ \\
            Times chosen weight ($w_{a=Chosen}$) & 0 & 1 \\
            Times chosen since new weight ($w_{a=ChosenSinceNew}$) & 0 & 0.5 \\
            Times seen weight ($w_{a=Seen}$) & 0 & 0 \\
            Missing neighbor: horizontal ($w_{n \in Horizontal}$) & 0.3 & 1 \\
            Missing neighbor: vertical ($w_{n \in Vertical}$) & 0.1 & 0 \\
            Missing neighbor: more keys ($w_{n \in MoreKeys}$) & 10 & N/A \\
        \end{tabular}
        \vspace{4mm}
        \caption{\textbf{Hyperparameter values for Montezuma's Revenge and Pitfall \emph{with} domain knowledge.} Note: the Atari emulator resolution is $160 \times 210$, which results in "tall" frames. However, Atari games were meant to be displayed with wide pixels, resulting in frames wider than they are tall. The common way to achieve this effect is to duplicate pixels horizontally, resulting in a $320 \times 210$ frame. We divide the frame into a $16 \times 16$ grid \emph{after} the frame is adjusted to $320 \times 210$, so that in the original frame space our cells would be $8 \times 16$.}
        \label{tab:hyper_domain} 
    \end{center}
\end{table}

\FloatBarrier

\subsection{Modifications to the Backward Algorithm}
\label{sec:backward_mods}

\subsubsection{Multiple demonstrations}

As mentioned, we modified the Backward Algorithm to robustify with multiple demonstrations, 10 in the case of Montezuma's Revenge. For Pitfall with domain knowledge (we did not robustify any trajectories without domain knowledge) and with the truncated trajectories (Section~\ref{sec:pitfall_results}), we robustified with 4 demonstrations. We did not robustify the long Pitfall trajectories with multiple demonstrations. While doing so is expected to improve robustification performance, it is unclear whether multiple demonstrations would enable successful robustification of the full-length Pitfall runs, and we leave this question for future work.

Handling multiple demonstrations in the Backwards Algorithm was implemented by choosing a demonstration uniformly at random each time the Backward Algorithm selects a demonstration state from which to start a rollout. Demonstration-specific information such as the current \texttt{max\_starting\_point} (the latest frame in the demonstration that the Backward Algorithm will start from) and success rates (the proportion of runs starting from a given starting point that performed at least as well as the demonstration) were tracked separately for each demonstration (see~\citet{salimans2018learning} for details on the various attributes used by the Backward Algorithm).

\subsubsection{Modified hyperparameters}

For robustification, we kept the default hyperparameters given by~\citet{salimans2018learning}, with the following exceptions: we added random no-ops at the beginning of the trajectory when the starting point was equal to 0 and we also added sticky actions throughout learning (unless otherwise specified). In addition, to improve performance when robustifying from multiple demonstrations, we set the success rate parameter to 0.1 instead of 0.2, and we changed the parameter that determines how frequently the starting point can be updated to $200 \cdot nDemos$ steps instead of a fixed 1024 steps. To avoid cases where the reward would be hard to find from the first checkpoint (i.e. the checkpoint closest to the end of the game), we also changed an internal starting-frame parameter (i.e. the number of frames before the end that the backward process would start robustifying from) from 256 to 0.
We found that these parameters seemed to work better empirically, though we did not experiment with them extensively.

\subsubsection{Pitfall-specific changes}

The small negative rewards in combination with the large positive rewards encountered on Pitfall required two additional changes in this particular game. The first change is to replace reward clipping with reward scaling: instead of rewards being clipped to the range [-1, 1], rewards are multiplied by 0.001. This change was necessary because, in Pitfall, negative rewards can have values as small as -1 while positive rewards have values between 2,000 and 5,000. Because negative rewards are so common and positive rewards so rare, clipping rewards gives a huge relative boost to avoiding negative rewards relative to obtaining positive rewards, which makes learning nearly impossible. With reward scaling, the relative importance of the two types of rewards is preserved, and learning succeeds. The scaling factor of 0.001 for Pitfall's rewards creates a reward range similar to that of clipped rewards, facilitating the use of the same hyperparameters (learning rate, entropy coefficient etc.) across Montezuma's Revenge and Pitfall. We chose to make a special case of Pitfall instead of using reward scaling in general for our method because reward clipping is more amenable to sharing hyperparameters across many different games~\cite{mnih:nature15}. An alternative to these domain-specific adjustments would be to implement automated reward scaling methods such as Pop-Art~\cite{van2016learning}.

Another change to the canonical Backward Algorithm relates to fixing an issue with \emph{early termination} and negative rewards. To quickly eliminate rollouts that are slow in collecting the rewards from the demonstration, the original Backward Algorithm implements \emph{early termination}, where it terminates all rollouts that do not get the same (or a higher) cumulative reward as the demonstration within a certain number of steps (50 in our case). The early termination is implemented in the form of a sliding window, where the cumulative reward of the current rollout is compared with the cumulative reward of the demonstration from 50 time steps ago, and if the cumulative reward of the current rollout is lower, the rollout is terminated. 
For example, if the demonstration collected a reward of 100 points at time step 20 (counting from the starting point of the rollout, and assuming no other rewards were collected), then a rollout will have to collect at least 100 points before time step 70, otherwise the rollout will be terminated at time step 70.

The sliding window method for early termination works fine when only positive rewards exist, as the only reason the rollout can have a lower score than the demonstration is because it failed to collect a particular positive reward within the given time frame. However, if \emph{negative} rewards exist, a rollout can also be terminated by collecting a negative reward, even if the demonstration collected the same negative reward. For example, if the demonstration collected a negative reward of -1 at time step 20 (once again, counting from the starting point of the rollout and assuming no other rewards were collected), the rollout needs to avoid this negative reward at all costs; otherwise it will be terminated at time step 50, even though it followed the same behavior as the demonstration. The reason for such early termination is that, at time step 50, the rollout will be compared with the performance of the demonstration at time step 0, and at that time step, the demonstration has not collected the negative reward yet.

To avoid this termination criteria, we give the agent an allowed score deficit of 250 points, meaning a rollout will only be terminated if its score is more than 250 points lower than that of the demonstration from 50 time steps earlier. This convention means that, as long as the demonstration did not collect more 250 points of negative reward within the given 50 time steps, the rollout will not be terminated if it follows the demonstration. The value of 250 points was found empirically on Pitfall, though future work could look for a more general method of implementing early termination in domains with negative reward.

\subsection{Performance}
\label{sec:performance}

\FloatBarrier

All Phase 1 runs were done on single virtual machines with 22 CPU cores. Each virtual machine had 50GB of RAM.

Table~\ref{tab:mont_level_perf} shows various performance metrics as a function of the level reached during Phase 1 the ``domain knowledge'' experiment on Montezuma's Revenge. The full experiment ran for 600M game frames, which took a mean of 74.9 \pivotci{72.6}{77.2}\percci{72.5}{77.2} hours.

\begin{table}[!htbp]
    \begin{center}
        \begin{tabular}{ c | c c c | c } 
            Level  & Solved \% & \makecell{Game Frames\\(excl. replay)}  & Time (hours) & \makecell{Hypothetical\\Game Frames\\(incl. replay)} \\
        \hline
        1 & 100\% & 58M \tpivotci{53M}{62M}\tpercci{53M}{62M}  & 0.9 \tpivotci{0.9}{1.0}\tpercci{0.9}{1.0} & 1.5B \tpivotci{1.3B}{1.6B}\tpercci{1.3B}{1.7B} \\
        2 & 100\% & 104M \tpivotci{97M}{111M}\tpercci{98M}{112M}  & 2.5 \tpivotci{2.3}{2.7}\tpercci{2.3}{2.7} & 4.3B \tpivotci{3.9B}{4.6B}\tpercci{3.9B}{4.7B} \\
        3 & 100\% & 173M \tpivotci{164M}{182M}\tpercci{165M}{182M}  & 6.8 \tpivotci{6.2}{7.3}\tpercci{6.2}{7.3} & 12B \tpivotci{12B}{13B}\tpercci{12B}{13B} \\
        4 & 100\% & 242M \tpivotci{230M}{253M}\tpercci{230M}{253M}  & 12.7 \tpivotci{11.7}{13.6}\tpercci{11.7}{13.7} & 24B \tpivotci{23B}{26B}\tpercci{23B}{26B} \\
        5 & 100\% & 305M \tpivotci{292M}{318M}\tpercci{293M}{319M}  & 19.9 \tpivotci{18.6}{21.3}\tpercci{18.6}{21.3} & 38B \tpivotci{36B}{41B}\tpercci{36B}{41B} \\
        6 & 100\% & 373M \tpivotci{358M}{388M}\tpercci{358M}{388M}  & 29.4 \tpivotci{27.8}{31.2}\tpercci{27.7}{31.2} & 57B \tpivotci{53B}{60B}\tpercci{53B}{60B} \\
        7 & 100\% & 432M \tpivotci{416M}{448M}\tpercci{417M}{448M}  & 39.1 \tpivotci{37.0}{41.2}\tpercci{37.0}{41.3} & 75B \tpivotci{71B}{79B}\tpercci{71B}{79B} \\
        8 & 94\% & 487M \tpivotci{471M}{503M}\tpercci{471M}{503M}  & 49.8 \tpivotci{47.4}{52.1}\tpercci{47.3}{52.1} & 96B \tpivotci{92B}{101B}\tpercci{92B}{101B} \\
        9 & 70\% & 533M \tpivotci{518M}{548M}\tpercci{518M}{548M}  & 61.4 \tpivotci{58.5}{64.3}\tpercci{58.5}{64.3} & 117B \tpivotci{113B}{121B}\tpercci{113B}{122B} \\
        10 & 38\% & 561M \tpivotci{550M}{572M}\tpercci{549M}{572M}  & 70.4 \tpivotci{67.7}{73.2}\tpercci{67.7}{73.0} & 139B \tpivotci{134B}{144B}\tpercci{134B}{144B} \\
        11 & 12\% & 582M \tpivotci{570M}{595M}\tpercci{569M}{593M}  & 77.6 \tpivotci{72.9}{82.0}\tpercci{73.1}{82.2} & 151B \tpivotci{147B}{155B}\tpercci{147B}{156B} \\
        
        \end{tabular}
        \vspace{4mm}
        \caption{\textbf{Mean computational complexity to reach different levels for Montezuma's Revenge with domain knowledge.} The ``Game frame (incl. replay)'' metric shows the number of game frames that would have been played if we replayed trajectories instead of resetting the emulator state. It is a hypothetical metric, since we did not replay the trajectories, but instead reset the environment. The ``Solved \%'' column shows the proportion of runs that solved a given level. All other metrics are computed only for the subset of runs that did solve the level.}
        \label{tab:mont_level_perf}
    \end{center}
\end{table}

It is worth noting that time scales superlinearly with game frames primarily due to: (1)~Cell selection, which happens at a fixed interval, but takes an amount of time proportional to the number of cells, which is constantly growing. We note that our cell selection is a form of Roulette-Wheel Selection (RWS)~\cite{goldberg:gabook89}, which we implement naively with an $O(n)$ algorithm. $O(\log n)$ and even $O(1)$ implementations for RWS are possible~\cite{Lipowski2011RoulettewheelSV}, so that cell selection could be sped up substantially in the future. (2)~Trajectory concatenation, which is implemented in a naive way where each cell contains an array that represents the entire trajectory needed to reach it, such that if cell B was reached from cell A, cell B's trajectory will contain a copy of the trajectory that leads to cell A, plus the actions that can lead from cell A to cell B. The copying of trajectories ever increasing in length is negligible at the start of the algorithm, but takes up more and more time as the algorithm goes on. An alternative representation with better memory and computational efficiency would be to represent trajectories as linked lists of actions, and in reverse order, so that each action links to its predecessor. With this representation, if cell B is reached from cell A, only the actions leading from cell A to cell B need to be stored in cell B, with the first of these actions linking to the last action needed to reach cell A, which means that adding cell B would take constant time, instead of a time proportional to the length of the longest trajectories in memory. Further, the amount of memory would also grow linearly, and the number of actions stored in memory would be bounded by the number of actions ever taken during exploration.

For Montezuma's Revenge without domain knowledge, performance metrics are shown in Table~\ref{tab:mont_no_dom_level_perf}. The full experiment ran for 1.2B game frames, which took 26.9 \pivotci{25.6}{28.2}\percci{25.7}{28.3} hours. It is notable that this is faster than the experiment with domain knowledge in spite of processing twice as many frames. This is likely due to the same reasons that domain knowledge runs get slower over time: runs without domain knowledge find fewer cells and shorter trajectories, and are thus less affected by the slowdown.

\begin{table}[!htbp]
    \begin{center}
        \begin{tabular}{ c | c c c | c } 
            Level  & Solved \% & \makecell{Game Frames\\(excl. replay)}  & Time (hours) & \makecell{Hypothetical\\Game Frames\\(incl. replay)} \\
        \hline
        1 & 57\% & 640M \tpivotci{567M}{711M}\tpercci{570M}{713M}  & 10.8 \tpivotci{9.5}{12.0}\tpercci{9.5}{12.1} & 33B \tpivotci{28B}{37B}\tpercci{28B}{38B} \\
        2 & 1\% & 592M \tpivotci{592M}{592M}\tpercci{592M}{592M}  & 11.4 \tpivotci{11.4}{11.4}\tpercci{11.4}{11.4} & 32B \tpivotci{32B}{32B}\tpercci{32B}{32B} \\

        \end{tabular}
        \vspace{4mm}
        \caption{\textbf{Mean computational complexity to reach different levels for Montezuma's Revenge without domain knowledge.} The ``Game frame (incl. replay)'' metric shows the number of game frames that would have been played if we replayed trajectories instead of resetting the emulator state. It is a hypothetical metric, since we did not replay the trajectories, but instead reset the environment. The ``Solved \%'' column shows the proportion of runs that solved a given level. All other metrics are computed only for the subset of runs that did solve the level.}
        \label{tab:mont_no_dom_level_perf}
    \end{center}
\end{table}

For Pitfall with domain knowledge, the threshold at which to compare game frames is not as clear as it is for Montezuma's Revenge. In order to include data from all of our 40 runs, we report the required game frames for reaching the lowest score achieved out of those runs, which is $47,534$. Reaching this threshold required a mean of 794.0M \pivotci{715.9M}{869.8M}\percci{718.1M}{871.1M} game frames, which takes 25.0 \pivotci{21.4}{28.3}\percci{21.7}{28.4} hours, and it would have required a mean of 100.8B \pivotci{84.1B}{116.0B}\percci{85.9B}{116.9B} game frames if trajectories had to be replayed from the start of the game. The full experiment lasted for 4.0B game frames, which took a mean of 186.3 \pivotci{184.9}{187.8}\percci{184.8}{187.7} hours. The full experiment would have required 1,060.4B \pivotci{1,048.5B}{1,071.7B}\percci{1,049.2B}{1,072.1B} game frames if trajectories had to be replayed from the start of the game.

Because Pitfall without domain knowledge did not obtain any rewards, it is hard to define good thresholds at which to compare game frames. In addition, regardless of the threshold we choose, the resulting data would not be representative of the resources Go-Explore would need to make progress on Pitfall (instead, it would represent the resource usage when Go-Explore fails to make progress). For those two reasons, we do not include Pitfall without domain knowledge in the remainder of this section.

We did not monitor the precise memory usage of Phase 1, beyond the fact that all our runs succeeded on machines with 50GB of RAM. Another indicator is the size of the serialized checkpoints produced at the end of each run, as these checkpoints contain all the necessary data to run Go-Explore, including the complete set of all cells, the metadata used in cell selection (see Appendix~\ref{sec:selection_details}), and the trajectories needed to reach the cells. Uncompressed, these files serialized using \texttt{pickle} have a mean size of 341.2MB \pivotci{292.3MB}{389.3MB}\percci{294.2MB}{390.5MB} in the case of Montezuma's Revenge without domain knowledge, and 2.8GB \pivotci{2.7GB}{2.9GB}\percci{2.7GB}{2.9GB} with domain knowledge. For Pitfall with domain knowledge, the mean uncompressed checkpoint size was 1.30GB \pivotci{1.29GB}{1.31GB}\percci{1.29GB}{1.31GB}.

For robustification, each run used 16 workers, each equipped with a single GPU, for a total of 16 GPUs per run. For Montezuma's Revenge without domain knowledge, runs lasted up to 5B game frames though the selected checkpoints were produced after a mean of 4.35B \pivotci{4.27B}{4.45B}\percci{4.25B}{4.43B} game frames (which took a mean of 2.4 \pivotci{2.4}{2.5}\percci{2.4}{2.5} days). For Montezuma's Revenge with domain knowledge, runs lasted up to 10B game frames but selected checkpoints were produced after a mean of 4.59B \pivotci{3.09B}{5.91B}\percci{3.26B}{6.05B} game frames, which took a mean of 2.6 \pivotci{1.8}{3.3}\percci{1.9}{3.4} days. For Pitfall with domain knowledge, runs lasted for about 12B game frames and selected checkpoints were produced after a mean of 8.20B \pivotci{6.73B}{9.74B}\percci{6.63B}{9.70B} game frames, which took a mean of 4.5 \pivotci{3.7}{5.3}\percci{3.6}{5.3} days.

\FloatBarrier

\subsection{Scores}
\label{sec:scores}

\FloatBarrier

Table~\ref{tab:scores} compares the results of Go-Explore with many other algorithms. The scores for the other algorithms are with stochastic \emph{testing} in the form of random no-ops, sticky actions, human restarts, or a combination thereof.
In the case of Go-Explore, both random no-ops and sticky actions were present in testing. As mentioned in Section~\ref{sec:determinism}, Go-Explore was \emph{trained} partially without sticky actions or random no-ops, whereas many of the algorithms in this table also handled stochasticity throughout training.

\begin{table}[!htbp]
    \begin{center}
        \begin{tabular}{ c | c c } 
            Algorithm  & \makecell{Montezuma's Revenge} & Pitfall \\
        \hline
        SARSA~\cite{Bellemare2012InvestigatingCA,mnih:nature15} & 259 & \na \\
        Linear~\cite{Bellemare2012InvestigatingCA,mnih:nature15} & 10.7 & \na \\
        DQN~\cite{mnih:nature15,wang2015dueling} & 0 & -286.1 \\
        Gorila~\cite{nair2015massively} & 84 & \na \\
        MP-EB~\cite{stadie2015incentivizing} & 0 & \na \\
        DDQN~\cite{van2016deep,wang2015dueling} & 42 & -30 \\
        Duel. DQN~\cite{wang2015dueling} & 22 & 0 \\
        Prior. DQN~\cite{schaul2015prioritized} & 13 & -15 \\
        A3C~\cite{mnih2016asynchronous} & 67 & -78 \\
        Pop-Art~\cite{van2016learning} & 0 & -3 \\
        DQN-CTS~\cite{bellemare2016unifying} & 3,705 & 0 \\
        A3C-CTS~\cite{bellemare2016unifying} & 1,127 & -259 \\
        BASS-hash~\cite{tang2017exploration} & 238 & \na \\
        DQN-PixelCNN~\cite{ostrovski2017count} & 2,514 & 0 \\
        ES~\cite{salimans2017evolution} & 0 & \na \\
        Reactor~\cite{gruslys2017reactor} & 2,643.5 & -3.5 \\
        Feature-EB~\cite{sasikumar2017exploration} & 2,745 &  \na\\
        C51~\cite{Bellemare2017ADP} & 0 & 0 \\
        UBE~\cite{ODonoghue2018TheUB} & 3,000 & 0 \\
        Rainbow~\cite{Hessel2018RainbowCI} & 154 & 0 \\
        IMPALA~\cite{espeholt:impala2018} & 2,643.5 & -1.2 \\
        Ape-X~\cite{horgan:apexdqn2018} & 2,500 & -1 \\
        DeepCS~\cite{Stanton2018DeepCS} & 3,500 & -186 \\
        RND~\cite{burda:rnd2018} & 11,347 & -3 \\
        PPO+CoEX~\cite{Choi2018ContingencyAwareEI} & 11,540 & \na \\
        \textbf{Go-Explore} & \textbf{43,763} & \na \\
        \makecell{\textbf{Go-Explore}\\(domain knowledge)} & \textbf{666,474} & \textbf{59,494} \\
        \textbf{Go-Explore} (best) & \textbf{18,003,200} & \textbf{107,363} \\
        \hline
        DQfD~\cite{hester2017deep} & 4,638 & 57 \\
        TDC+CMC~\cite{aytar2018playing} & 41,098 & 76,813 \\
        Ape-X DQfD~\cite{pohlen2018observe} & 29,384 & 3,997 \\
        LfSD (best)~\cite{salimans2018learning} & 74,500 & \na \\
        \hline
        Average Human~\cite{pohlen2018observe} & 4,753 & 6,464 \\
        Human Expert~\cite{pohlen2018observe} & 34,900 & 47,821 \\
        Human World Record~\cite{atari_scoreboard} & 1,219,200 & 114,000 \\

        \end{tabular}
        \vspace{4mm}
        \caption{\textbf{Comparison to baselines for Montezuma's Revenge and Pitfall.} The second section gives the performance of imitation learning algorithms, while the third gives human performance. Results are given in order of first public release (including preprints). Many historical papers did not consider Pitfall, in which case the score is displayed as ``\na''. Two references are given in cases where the score from a given method does not appear in its original paper, but appears in another paper (usually in a comparison section).}
        \label{tab:scores} 
    \end{center}
\end{table}

\FloatBarrier

\subsection{Fine-grained cell representation for Pitfall without domain knowledge}
\label{sec:pitfall_cell_representation}

\begin{figure}[tbh]
    \centering
    \includegraphics[width=0.33\linewidth]{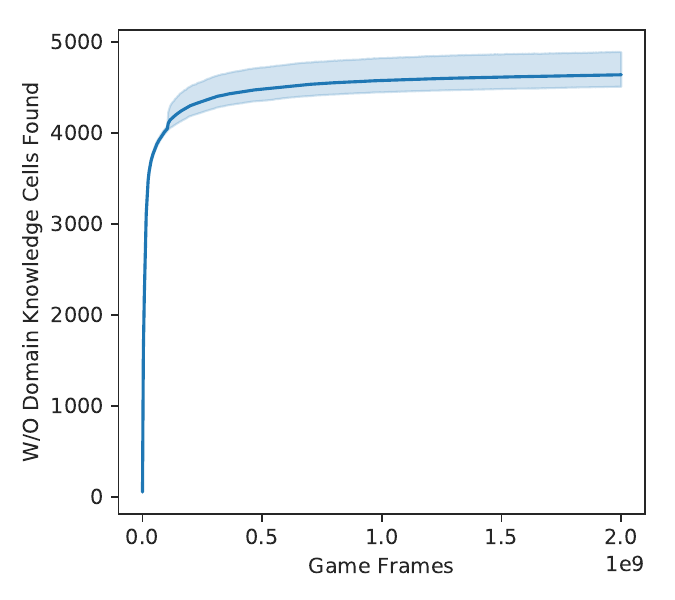}
    \caption{\textbf{Number of without-domain-knowledge cells found during Phase 1 on Pitfall without domain knowledge.} Most cells are found within the first 500M game frames, after which very few new cells are found. This observation suggests that Pitfall without domain knowledge fails because there are too many different states that are mapped to the same Go-Explore cell.}
    \label{fig:pitfall_cells_found_stagnation}
\end{figure}

As mentioned in Section~\ref{sec:pitfall_results}, we attempted an experiment on Pitfall without domain knowledge using the same parameters as with Montezuma's Revenge. This approach did not succeed, as Go-Explore quickly stopped finding new rooms and failed to find any rewards (Fig.~\ref{fig:pitfall_explore}). One potential reason for this failure is that the downscaled cell representation optimized for Montezuma's Revenge conflates too many states into the same cell in Pitfall. This hypothesis is supported by the fact that Go-Explore stops discovering new cells, both when measured post-hoc as domain knowledge cells (Fig.~\ref{fig:pitfall_explore}) and in the downscaled representation of cells in which it is actually searching (Fig.~\ref{fig:pitfall_cells_found_stagnation}).

\begin{figure}[tbh]
    \begin{subfigure}[t]{.33\textwidth}
        \centering
        \includegraphics[width=\linewidth]{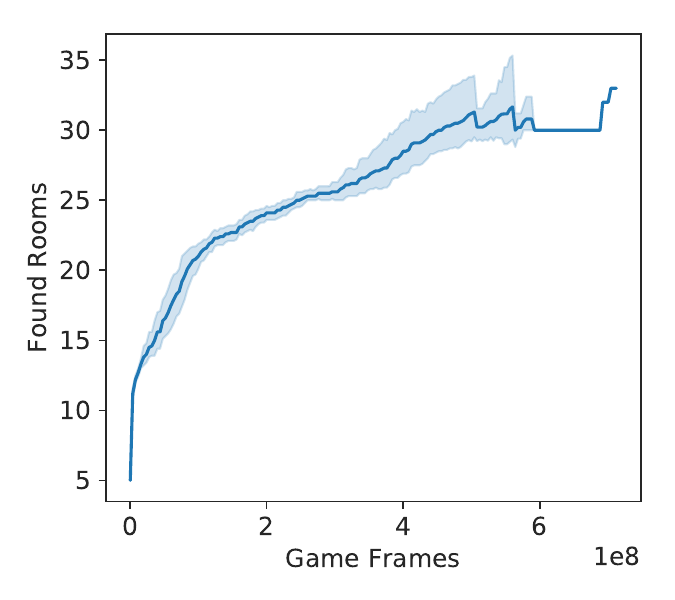}
        \caption{Number of rooms found}
        \label{fig:pitfall_fine_grained_room}
    \end{subfigure}
    \begin{subfigure}[t]{.33\textwidth}
        \centering
        \includegraphics[width=\linewidth]{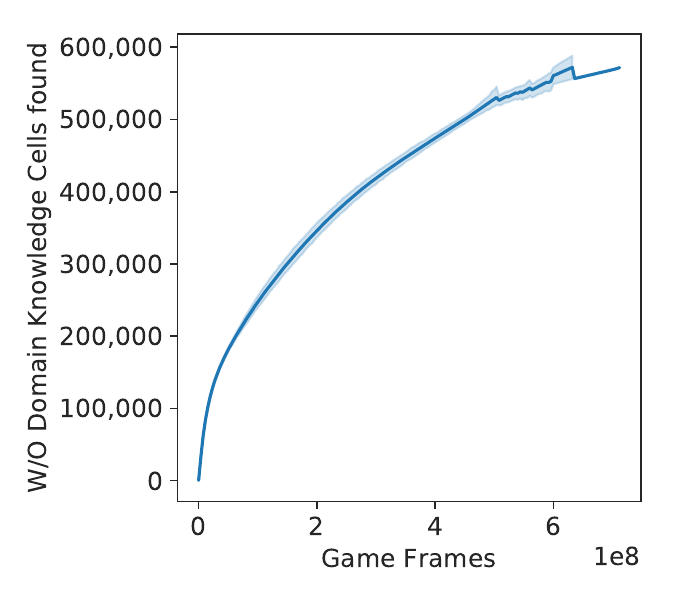}
        \caption{Number of cells found}
        \label{fig:pitfall_fine_grained_cell}
    \end{subfigure}
    \begin{subfigure}[t]{.33\textwidth}
        \centering
        \includegraphics[width=\linewidth]{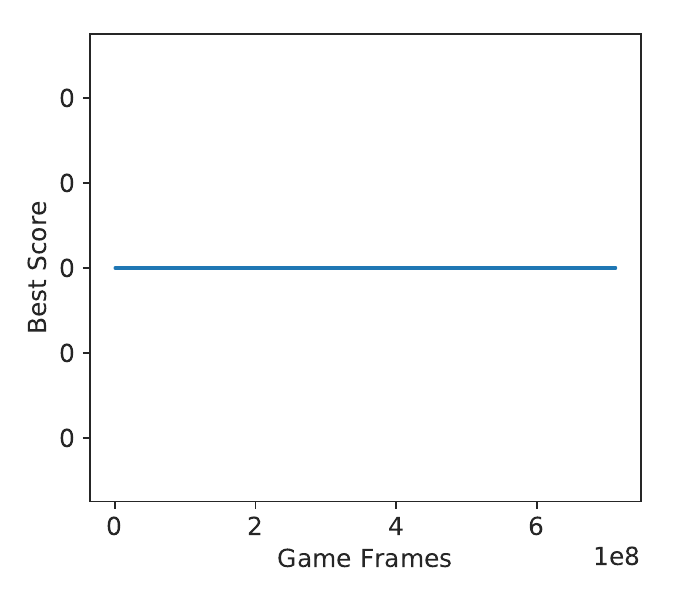}
        \caption{Maximum score in archive}
        \label{fig:pitfall_fine_grained_score}
    \end{subfigure}
    \caption{\textbf{Go-Explore Phase 1 on Pitfall without domain knowledge with a more fine-grained (16 different pixel values instead of 8) cell representation.} While the number of rooms (\subref{fig:pitfall_fine_grained_room}) and the number of cells (\subref{fig:pitfall_fine_grained_cell}) found continues to increase, even after 600M game frames, the runs do not continue beyond this point because they run out of memory. Despite visiting more rooms, Go-Explore still does not find any rewards, although it may have were it able to continue for longer (\subref{fig:pitfall_fine_grained_score}). The noise at the end of sub-figure (\subref{fig:pitfall_fine_grained_room}) is caused by different runs crashing at different times. The plot shows the mean and $95\%$ bootstrapped confidence interval over 20 runs initially, but the number of runs declines over time. The first run crashes around 500M game frames.}
    \label{fig:pitfall_explore_fine_grained}
\end{figure}

To resolve this issue, we looked at different cell representations that would be able to distinguish a larger number of states. A particularly promising cell representation assigns 16, rather than 8, different pixel values to each pixel in the $11 \times 8$ downscaled representation. While this cell representation does result in a larger number of rooms visited and the number of downscaled cells found did not stagnate, the runs terminated prematurely due to exhausting the 50GB of memory available on the virtual machine (Fig.~\ref{fig:pitfall_explore_fine_grained}). Better hardware, distributed computation, or algorithmic improvements are all potential methods to resolve this issue, but we leave their implementation to future work.

\FloatBarrier

\subsection{Failure robustifying long trajectories in Pitfall}
\label{sec:pitfall_long_robustification_failure}

\begin{figure}[tbh]
    \centering
    \includegraphics[width=0.5\linewidth]{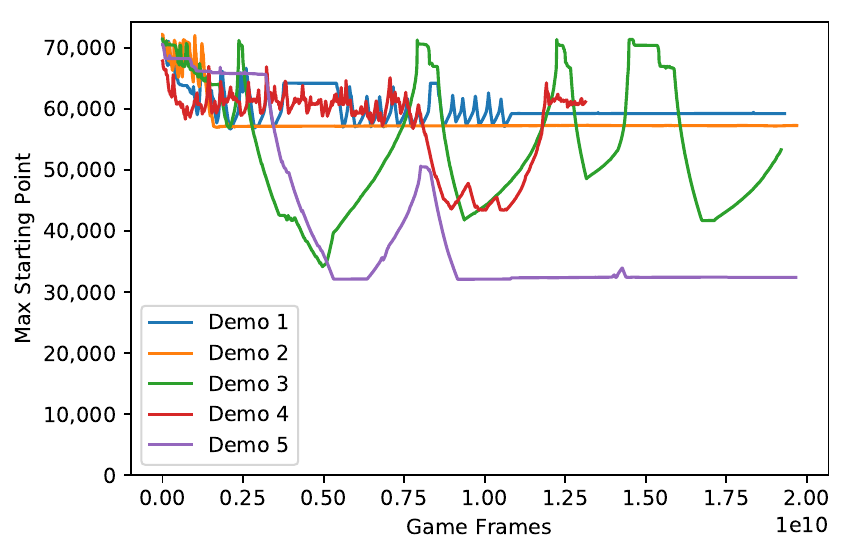}
    \caption{\textbf{Maximum starting point over training when robustifying the full-length trajectories produced by Go-Explore in Phase 1 on Pitfall with domain knowledge.} Unlike in Fig.~\ref{fig:robust_multi}, the lines in this figure represent \emph{separate} robustification attempts, each of which was applied to a single demonstration taken from different runs of Go-Explore Phase 1. None of the 5 robustification attempts reaches a starting point near 0, meaning that robustification failed on these demonstrations. We did not try to robustify from multiple demonstrations for want of time, although doing so may have worked better.}
    \label{fig:pitfall_long_robustification_failure}
\end{figure}

While Go-Explore on Pitfall with domain knowledge is able to find trajectories that score over 70,000 points (Fig.~\ref{fig:pitfall_explore}), the Backward Algorithm was unable to robustify these trajectories (Fig.~\ref{fig:pitfall_long_robustification_failure}).

\FloatBarrier

\subsection{Nearly identical states in Pitfall}
\label{sec:pitfall_nearly_identical_states}

\begin{figure}[tbh]
    \centering
    \includegraphics[width=0.45\linewidth]{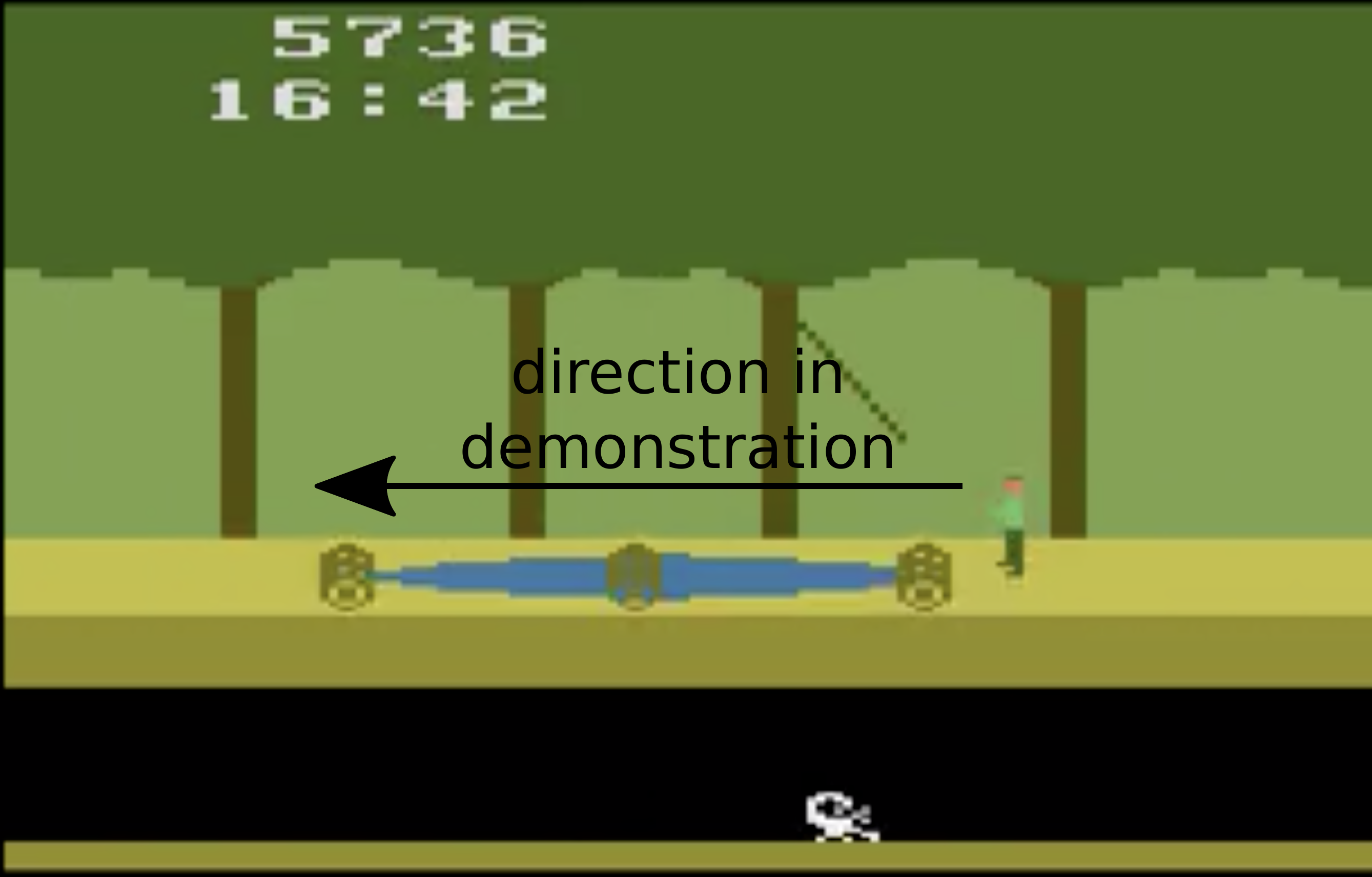}
    \includegraphics[width=0.45\linewidth]{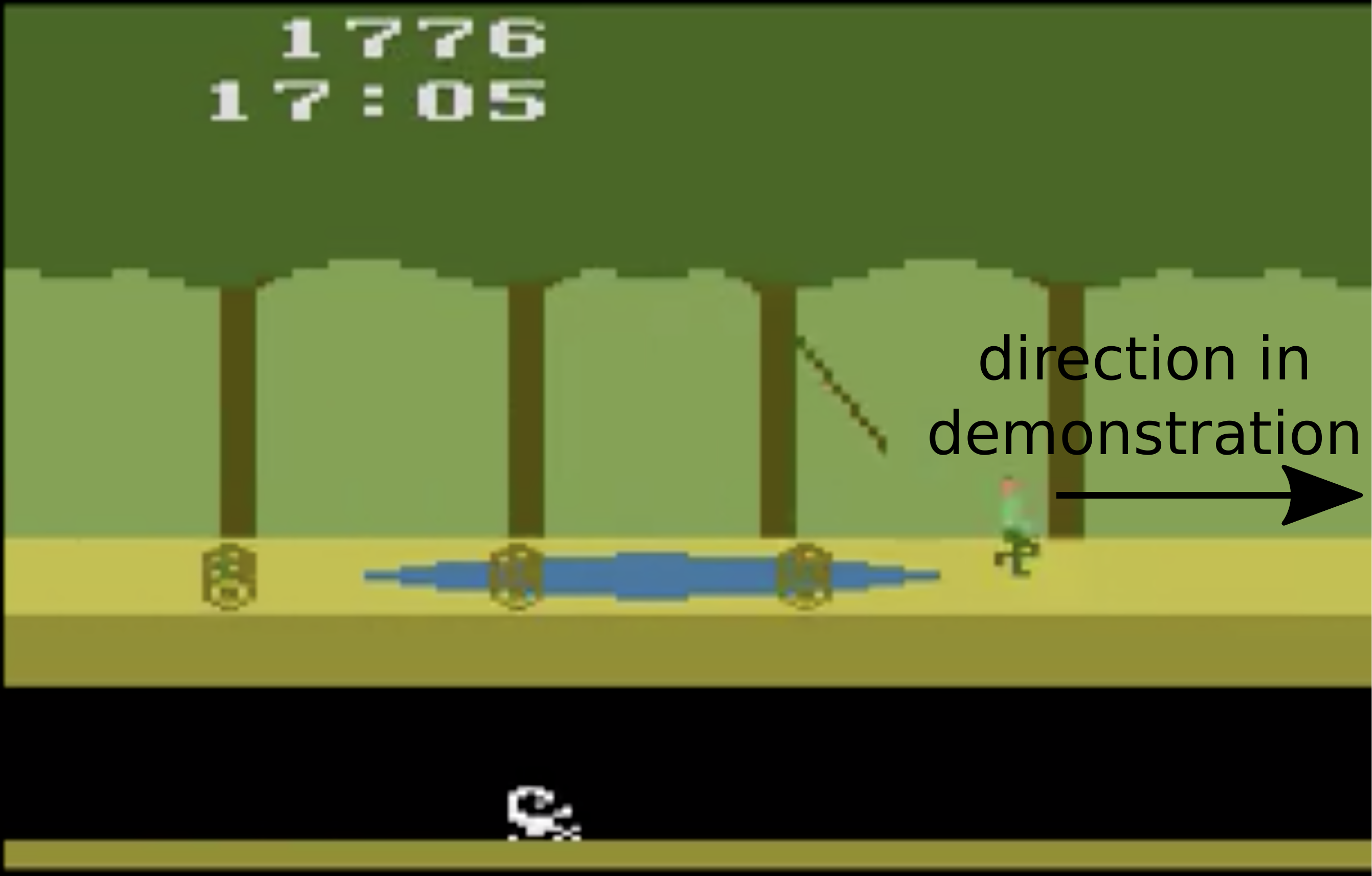}
    \caption{\textbf{Two nearly identical looking states that require different actions to be navigated optimally.} The two screenshots are taken from the same Go-Explore demonstration, but at different times. The rooms are conceptually identical: they both contain a blue pool, a swinging vine, three rolling logs, and a scorpion. However, because the two rooms are located in different areas of the game, the correct actions for navigating the two rooms can be different. In this case, the Go-Explore demonstration navigates the left room right to left, whereas it navigates the right room from left to right. When training a policy in this situation, there will be many similar looking states that require opposite actions. While the moving aspects of the room (i.e. the vine, the logs, and the scorpion) are likely to be in different locations in the two rooms of the demonstration, the fact that they will also be in different locations when entering the rooms at different times makes them poor features for differentiation. Probably the most informative features that can be used to determine in which direction to move are the score counter and the clock (the white numbers in the top left of each image), though, in practice, these small, frequently-changing features seem insufficient to provide the necessary guidance.}

    \label{fig:pitfall_nearly_identical_states}
\end{figure}

Pitfall contains many rooms located in different parts of the game that contain the exact same objects and hazards. These identical rooms can result in nearly identical-looking states that require different actions to be navigated optimally (Fig.~\ref{fig:pitfall_nearly_identical_states}), and they indicate that Pitfall is a Partially Observable Markov Decision Process (POMDP). These nearly identical looking states can pose a problem, both when robustifying trajectories that visit some of these states, and when designing a domain-agnostic cell representation that should, ideally, treat these states as being in different cells.

The general method for handling POMDPs is to condition the current action on all previously observed states, for example by training a recurrent, rather than feed-forward, neural network. For the robustification phase our method already implements a recurrent layer in the neural network, but, possibly due to the way the network is trained with the Backward Algorithm (i.e.\ whenever the agent is started from a particular state, it is not presented with all states that would have come before), this recurrent layer does not appear to completely solve the issue (see also section~\ref{sec:pitfall_results}). A similar approach could be applied for obtaining cell representations (e.g.\ a cell representation could be conditioned on all observations of the trajectory to a particular state, rather than just the observation at a particular state), but care would have to be taken to ensure that actually identical (or nearly identical) states are recognized as such. 

\FloatBarrier

\end{document}